%% file: main.tex
\documentclass[11pt]{article}
\usepackage[preprint]{acl}

\usepackage{times}
\usepackage{latexsym}
\usepackage[T1]{fontenc}
\usepackage[utf8]{inputenc}
\usepackage{microtype}
\usepackage{inconsolata}
\usepackage{graphicx}

\usepackage{xcolor,colortbl}
\usepackage{amsmath,amsfonts,amssymb,mathtools}
\graphicspath{{figures/}}
\usepackage{subcaption}
\usepackage{adjustbox}
\usepackage[breakable]{tcolorbox}
\usepackage{bbding}
\usepackage{booktabs,array,multirow,threeparttable,arydshln}
\usepackage{algorithm,algorithmic}
\usepackage{pifont,textcomp,url,enumitem}
\usepackage[misc]{ifsym}
\usepackage{twemojis}
\usepackage{fontawesome5}

\definecolor{myorange}{HTML}{F2AA84}
\definecolor{mygray}{HTML}{C0C0C0}
\definecolor{mygreen}{HTML}{8ED973}
\definecolor{pink}{HTML}{747199}
\newcommand{\reduce}[1]{\textcolor{pink}{#1}}
\newcommand{\scalea}[1]{\scalebox{0.7}{#1}}
\newcommand{\scaleb}[1]{\scalebox{0.9}{#1}}

\newcommand{\cross}{\textcolor{lightgray}{\XSolidBrush}}
\newcommand{\cbox}[1]{\raisebox{1mm}{\colorbox{#1}{\quad}}}
\newcommand{\upa}{\rlap{$\uparrow$}}
\newcommand{\dna}{\rlap{$\downarrow$}}
\newcommand{\mysquare}{\ensuremath{\square}}
\newcommand{\bst}[1]{\textbf{\textcolor{red}{#1}}}
\newcommand{\subbst}[1]{\textcolor{blue}{\underline{#1}}}
\title{LIVE: An \underline{L}LM-assisted Multi-\underline{I}mage
\underline{V}isual Hallucination \underline{E}valuation Benchmark}
\author{
\textbf{Yitong Qiao}\textsuperscript{1,*}\quad
\textbf{Licheng Pan}\textsuperscript{1,*}\quad
\textbf{Yu Mi}\textsuperscript{1}\quad
\textbf{Bingyang Wang}\textsuperscript{2}\\
\textbf{Chih-Ting Liao}\textsuperscript{3}\quad
\textbf{Chaoyue Yang}\textsuperscript{1}\quad
\textbf{Anda Cheng}\textsuperscript{4}\quad
\textbf{Yufei Lu}\textsuperscript{4}\\
\textbf{Fei Sun}\textsuperscript{5}\quad
\textbf{Zhixuan Chu}\textsuperscript{1,\Letter}\quad
\textbf{Kui Ren}\textsuperscript{1}\\
{\normalfont\normalsize\textsuperscript{1}The State Key Laboratory of Blockchain and Data Security,}\\
{\normalfont\normalsize Zhejiang University}\\
{\normalfont\normalsize\textsuperscript{2}Emory University\quad
\textsuperscript{3}University of New South Wales\quad
\textsuperscript{4}Ant Group}\\
{\normalfont\normalsize\textsuperscript{5}Institute of Computing Technology, Chinese Academy of Sciences}\\
{\normalfont\small\textsuperscript{*}Equal contribution.\quad
\textsuperscript{\Letter}Correspondence: \texttt{zhixuanchu@zju.edu.cn}}\\
{\normalfont\normalsize
\href{https://github.com/Master-PLC/LIVE}{\faGithub\enspace GitHub}\qquad
\href{https://huggingface.co/datasets/Tong613/LIVE-multi-image-bench}{\twemoji[height=0.95em]{1f917}\enspace Hugging Face}}
}

\begin{document}
\maketitle

\begin{abstract}
Despite remarkable progress in Large Vision-Language Models (LVLMs), visual hallucination evaluation remains predominantly focused on single-image scenarios, leaving multi-image contexts largely underexplored. In this work, we identify two key challenges in multi-image hallucination evaluation: \ding{182} \textbf{context-dependent hallucination patterns}, where LVLMs are prone to content confusion across similar images in Uniform Image Contexts (UIC) but suffer from content interference across heterogeneous visual cues in Diverse Image Contexts (DIC); and \ding{183} \textbf{incomplete evaluation coverage}, where traditional single-query methods struggle to evaluate multi-image understanding systematically across different subset sizes. To address these issues, we introduce \textbf{LIVE}, an \textbf{L}LM-assisted multi-\textbf{I}mage \textbf{V}isual hallucination \textbf{E}valuation benchmark. For challenge \ding{182}, LIVE curates over 42k yes/no questions covering both UIC and DIC scenarios across six visual recognition tasks, ranging from low-level perception to high-level cognition. For challenge \ding{183}, we propose a \textbf{Multi-granularity Assessment Protocol (MAP)} to measure hallucination rates across varying numbers of target images. Experiments on various LVLMs demonstrate distinct performance variations between UIC and DIC scenarios. Moreover, MAP reveals that hallucination rates amplify with increasing image quantities, exposing failure modes overlooked by single-query evaluations. This provides a practical foundation for understanding and diagnosing multi-image hallucinations.
\end{abstract}

\section{Introduction}

 Large Vision-Language Models (LVLMs) have demonstrated impressive capabilities across diverse multimodal tasks~\citep{song2024efficiently,lu2025benchmarking,kuang2025natural,sima2024drivelm,huang2024smartedit}. Despite this progress, they still suffer from a critical limitation: \textit{visual hallucination}, where models generate descriptions or answers that are inconsistent with the actual visual content~\citep{guan2024hallusionbench}. This phenomenon significantly undermines model reliability in applications that depend on visually grounded outputs.
 To evaluate this issue, early research predominantly focused on developing single-image benchmarks~\citep{chen2024multi,huang2024visual}. However, practical scenarios increasingly require models to process and reason over multiple images, such as surveillance analysis~\citep{chen2025benchmark}, medical imaging sequences~\citep{yang2025medical}, and general multi-image understanding~\citep{Ye2025M4BenchAB}. Consequently, there is an urgent need to extend hallucination evaluation from single-image scenarios to multi-image application demands.

\begin{table*}[t]
\centering
\caption{Overview of related visual-hallucination and multi-image understanding benchmarks.}
\label{tab:benchmark_comparison}
\small
\setlength{\tabcolsep}{6.3pt}
\renewcommand{\arraystretch}{1}
\begin{adjustbox}{max width=\linewidth}
\begin{threeparttable}
\begin{tabular}{lcccccccccc}
\toprule
\textbf{Benchmark} & \textbf{Multi-Image} & \textbf{UIC} & \textbf{DIC} & \textbf{Object} & \textbf{Material} & \textbf{Color} & \textbf{Sentiment} & \textbf{Action} & \textbf{Position} & \textbf{\# Samples} \\
\midrule
POPE~\citep{li2023evaluating} & \cross & -- & -- & \Checkmark & \cross & \cross & \cross & \cross & \cross & 3000 \\
M-HalDetect~\citep{gunjal2024detecting} & \cross & -- & -- & \Checkmark & \cross & \cross & \cross & \cross & \Checkmark & 16000 \\
AutoHallusion~\citep{wu2024autohallusion} & \cross & -- & -- & \Checkmark & \cross & \cross & \cross & \cross & \Checkmark & 3129 \\
PhD~\citep{liu2025phd} & \cross & -- & -- & \Checkmark & \Checkmark & \Checkmark & \Checkmark & \cross & \Checkmark & 102564 \\
HallusionBench~\citep{guan2024hallusionbench} & \Checkmark$^\dagger$ & -- & -- & \Checkmark & \cross & \cross & \cross & \Checkmark & \Checkmark & 1129 \\
M4Bench~\citep{Ye2025M4BenchAB} & \Checkmark & -- & -- & \Checkmark & \cross & \cross & \cross & \cross & \Checkmark & 5700 \\
MIHBench~\citep{li2025mihbench} & \Checkmark & -- & -- & \Checkmark & \cross & \cross & \cross & \cross & \cross & 4000 \\
LLSAVisionQA~\citep{sun2025explore} & \Checkmark & -- & -- & \cross & \cross & \Checkmark & \cross & \cross & \cross & 4989 \\
\midrule
\textbf{LIVE (Ours)} & \Checkmark & \Checkmark & \Checkmark & \Checkmark & \Checkmark & \Checkmark & \Checkmark & \Checkmark & \Checkmark & \textbf{42038} \\
\bottomrule
\end{tabular}
\begin{tablenotes}
    \scriptsize
    \item \textit{Note}: ``UIC'' denotes Uniform Image Context, and ``DIC'' denotes Diverse Image Context. A dash indicates that the benchmark does not explicitly use this UIC/DIC partition.
    \item M-HalDetect contains 16,000 image--description examples from 4,000 images. POPE reports 3,000 questions per sampling setting. LLSAVisionQA includes 2,990 single-image questions and 1,999 image-pair questions.
    \item AutoHallusion's coverage marks describe its generated examples. Its released sample count also includes HallusionBench.
    \item $^\dagger$HallusionBench includes temporal questions over image sequences.
\end{tablenotes}
\end{threeparttable}
\end{adjustbox}
\end{table*}

While early efforts have begun to explore multi-image hallucination, systematically evaluating models in these complex scenarios presents two fundamental challenges: \ding{182} \textbf{Distinct hallucination patterns under uniform versus diverse image contexts.} In \textit{Uniform Image Contexts} (UIC), where images share similar content, LVLMs may confuse image-specific details and transfer attributes across similar images. For example, in multiple similar street scenes, a model may claim that a red car appearing in only one image is present in all of them. In \textit{Diverse Image Contexts} (DIC), where images differ substantially, models may instead suffer from interference across heterogeneous visual cues, leading to factual errors during cross-image reasoning. These differences suggest that multi-image hallucination should be evaluated under both UIC and DIC rather than treated as a single setting. \ding{183} \textbf{Traditional single-query evaluation provides limited coverage of multi-image understanding across different granularities.} Existing evaluations often ask one question about the entire image set, which gives only a coarse view of model behavior. In practice, however, a model may answer correctly for one image or a pair of images but fail when the query involves three or four images. A more complete evaluation should therefore examine hallucination across different subset sizes to reveal how performance changes with reasoning scope.

To address these challenges, we introduce \textbf{LIVE}, a benchmark for \textbf{L}LM-assisted multi-\textbf{I}mage \textbf{V}isual hallucination \textbf{E}valuation. For challenge \ding{182}, LIVE covers UIC and DIC scenarios across six visual recognition tasks, from low-level perception (object, material, and color) to higher-level cognition (sentiment, action, and position). For challenge \ding{183}, we propose the \textbf{Multi-granularity Assessment Protocol (MAP)}, which measures hallucination across different image subset sizes instead of a single query per image set.
Specifically, LIVE is built through a three-stage pipeline.
(1) \textbf{Scenario-Level Image Curation} uses CLIP-based clustering and sampling to construct diverse and representative UIC and DIC scenarios.
(2) \textbf{Task-Specific Multi-image Hallucination Item (MHI) Mining} extracts factual content from image captions and uses LLMs to generate semantically plausible counterfactuals absent from the images.
(3) \textbf{MHI item-embedded Question Generation} forms balanced yes/no questions targeting different image subset granularities, where factual items correspond to ``yes'' and counterfactual items correspond to ``no''.

LIVE contains 1,871 daily-life images organized into 488 multi-image scenarios and 42,038 binary visual question-answer triplets. Extensive experiments on eleven state-of-the-art LVLMs validate our analysis of the two identified challenges. First, models exhibit distinct context-dependent vulnerabilities, generally suffering higher hallucination rates in DIC compared to UIC scenarios. Second, our MAP evaluation reveals that model performance degrades as the number of queried images increases. This finding highlights how traditional single-query evaluations obscure actual multi-image reasoning limitations. Overall, LIVE provides a concrete foundation for diagnosing multi-image hallucinations. Our main contributions are summarized as follows:

\begin{itemize}[leftmargin=1em]
\item We identify and formalize two key challenges in multi-image visual hallucination evaluation: context-dependent hallucination patterns between UIC and DIC scenarios, and incomplete evaluation coverage in single-query approaches.

\item We propose LIVE, a benchmark covering both UIC and DIC scenarios across six visual recognition tasks, built with a systematic three-stage pipeline for scalable and high-quality construction.

\item We introduce the Multi-granularity Assessment Protocol, evaluating LVLM hallucinations across different image granularities.

\item We conduct extensive experiments on state-of-the-art LVLMs, showing clear context-dependent performance differences and the value of multi-granularity evaluation in multi-image settings.
\end{itemize}

\begin{figure*}[t]
\centering
\includegraphics[width=1.0\linewidth]{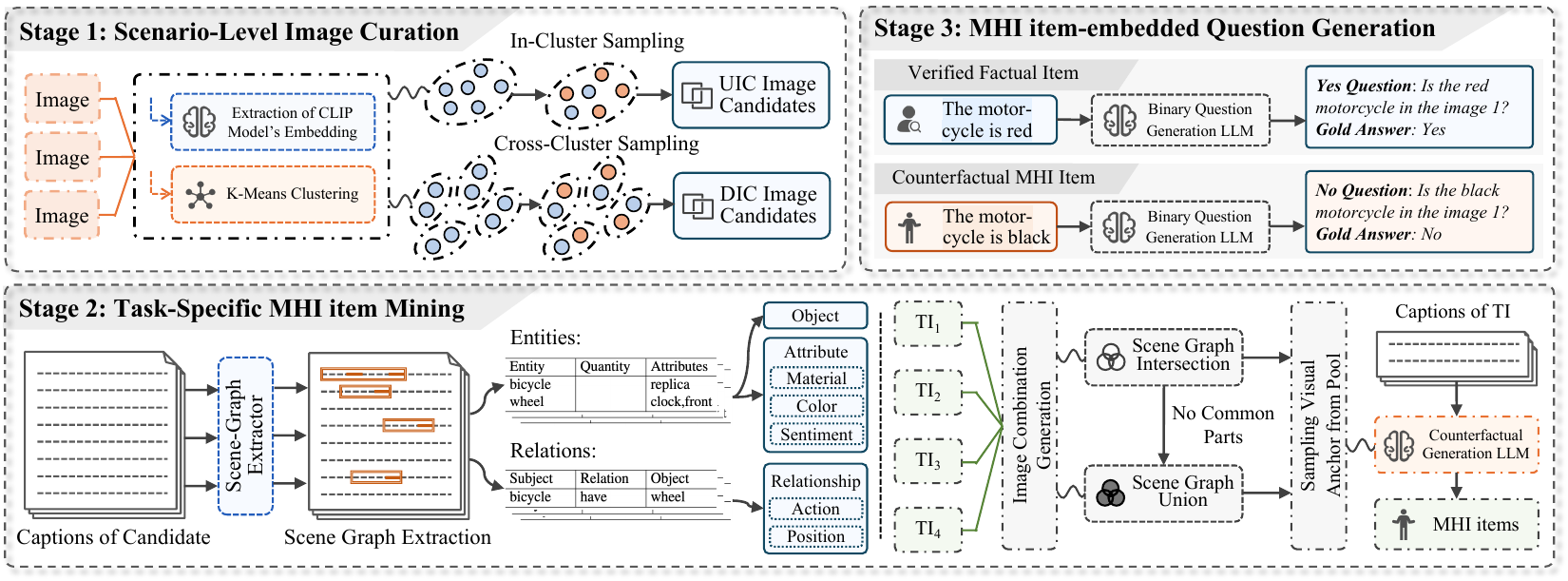}
\caption{Construction pipeline of our LIVE benchmark.}
\label{fig:pipeline}
\end{figure*}

\section{Related Work}
\subsection{Large Vision-Language Models}

Recent advances in multimodal architectures have led to the rapid development of Large Vision-Language Models (LVLMs), which integrate visual and textual understanding and achieve strong performance on a wide range of tasks~\citep{hurst2024gpt,bai2025qwen2,lu2024deepseek,chen2025prioritizing,li2025breen}. Alongside these advances, visual hallucination has become a central concern, referring to cases where model outputs are inconsistent with the visual input~\citep{guan2024hallusionbench,li2025mihbench}. This phenomenon spans object recognition, attribute prediction, and spatial reasoning, limiting the reliability of LVLMs in real-world applications that require factual accuracy~\citep{chen2025benchmark,yang2025medical,xu2024vlm}. This has motivated evaluation methods and benchmarks for hallucination analysis.

\subsection{Single-Image Hallucination Benchmarks}
Early hallucination benchmarks focus on single-image scenarios. POPE~\citep{li2023evaluating} pioneered controlled assessment through binary questioning protocols, establishing methodological foundations for hallucination measurement. Subsequent work, including AutoHallusion~\citep{wu2024autohallusion} and M-HalDetect~\citep{gunjal2024detecting}, expanded evaluation to spatial reasoning, attribute recognition, and complex visual relationships. These benchmarks help identify LVLM limitations~\citep{li2025survey}. Recent work also uses LLMs to generate counterfactual queries across cognitive tasks (e.g., sentiment and position) for single-image hallucination evaluation, such as PhD~\citep{liu2025phd}. Their single-image focus limits evaluation of factual consistency and confusion across multiple images.

\subsection{Multi-Image Hallucination Benchmarks}
Recent multi-image evaluations include M4Bench~\citep{Ye2025M4BenchAB} for general multi-image understanding and MIHBench~\citep{li2025mihbench} for object-related hallucination. They examine different aspects of multi-image understanding. \ding{182} \textbf{Context-dependent hallucination patterns need task-specific evaluation.} M4Bench contrasts single- and multi-domain images, whereas MIHBench examines object existence, count, and identity consistency. LIVE defines UIC and DIC through a shared CLIP-based construction and evaluates both settings across six recognition tasks. \ding{183} \textbf{Target-image subset size provides another evaluation dimension.} M4Bench evaluates object- and part-level granularity. LIVE instead varies the number of target images within a fixed four-image input to reveal errors across target subsets. Full-set evaluation alone can miss these failures.

Table~\ref{tab:benchmark_comparison} positions LIVE against single-image hallucination and multi-image understanding benchmarks. LIVE combines UIC and DIC across six recognition tasks with evaluation over every target-image subset size. Binary questioning and LLM-assisted generation support this evaluation.

\section{LIVE Benchmark Construction}

Figure~\ref{fig:pipeline} shows the three-stage construction pipeline of LIVE. Section~\ref{subsec:image_curation} describes UIC and DIC image curation. Section~\ref{subsec:mihitem_mining} presents Multi-image Hallucination Item (MHI) mining across six visual tasks, and Section~\ref{subsec:question_generation} introduces question generation for the Multi-granularity Assessment Protocol (MAP). Dataset statistics and validation results are provided in Sections~\ref{subsec:dataset_overview} and~\ref{subsec:construction_validation}.

\subsection{Stage 1: Scenario-Level Image Curation}
\label{subsec:image_curation}

To evaluate context-dependent hallucination patterns (challenge \ding{182}), we design a clustering-based curation strategy that constructs two multi-image scenarios: \textbf{UIC} featuring four visually and semantically similar images, and \textbf{DIC} featuring four distinct images. We assess scenario quality through manual inspection and within-group CLIP similarity. Across all six tasks, within-group CLIP similarity is 0.09--0.11 higher for UIC than for DIC, with task-wise values provided in Appendix~\ref{sec:clip_sim}.

\paragraph{CLIP-based Embedding and Clustering}
As illustrated in Fig.~\ref{fig:pipeline}, we start from the MS-COCO val2014 split~\citep{lin2014microsoft}, extract CLIP~\citep{radford2021learning} embeddings for all images and apply K-means clustering~\citep{ahmed2020k} to partition the 512-dimensional embedding space into 1,000 clusters using Faiss~\citep{johnson2019billion}. The clusters group semantically similar images for UIC and separate different visual content for DIC.

\begin{figure}[ht]
    \centering
    \includegraphics[width=0.95\linewidth]{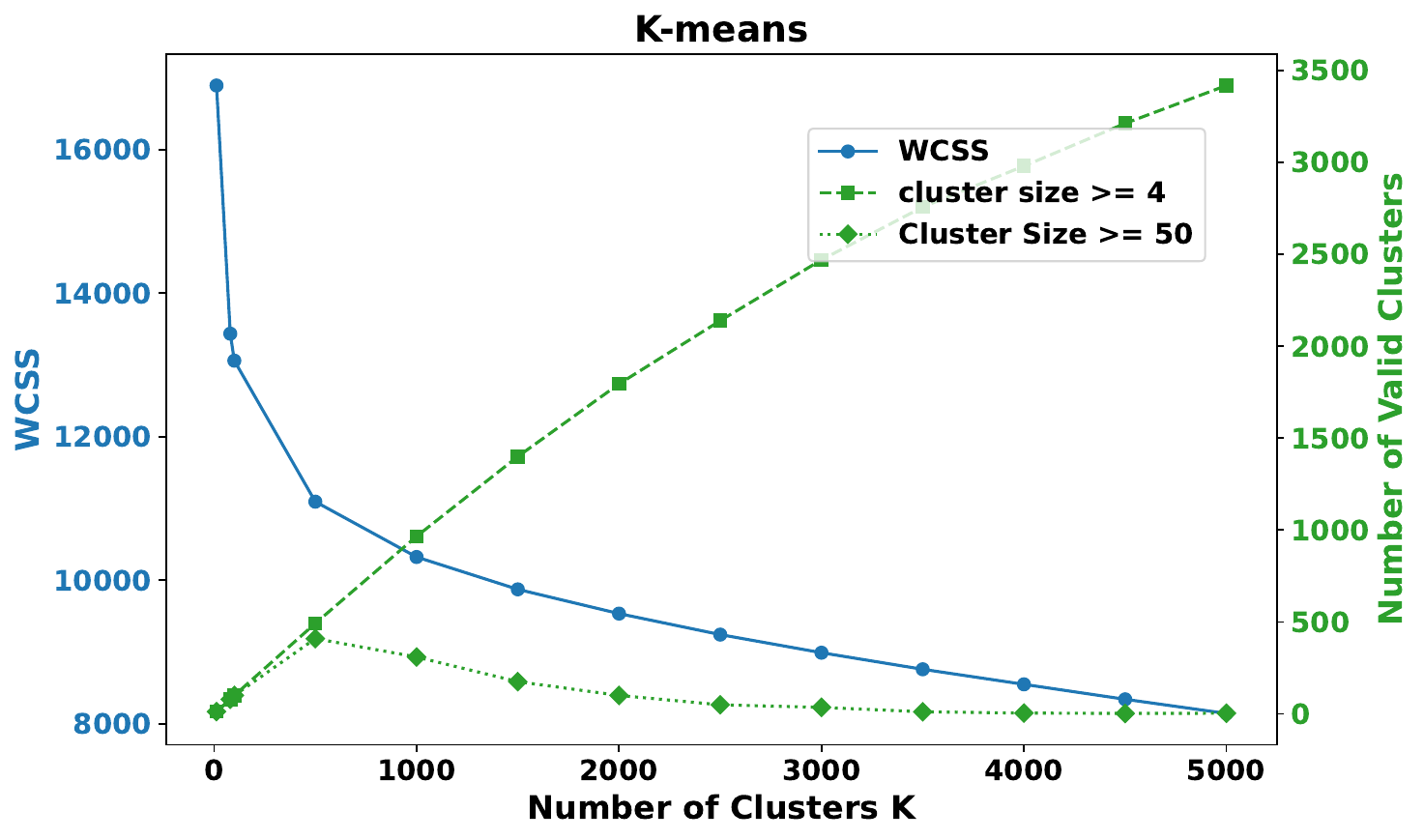}
    \caption{K-means analysis for different numbers of clusters $K$.}
    \label{fig:kmeans_k}
\end{figure}

\paragraph{Empirical Choice of the Cluster Number}
We vary $K$ from 12 to 5,000 and examine the within-cluster sum of squares (WCSS), the number of clusters containing at least four images, and the number containing at least 50 images. Figure~\ref{fig:kmeans_k} shows that the WCSS reduction begins to flatten around $K=1{,}000$, while this setting retains high-density clusters for within-cluster sampling and medium-sized clusters for diverse scenario construction. We choose $K=1{,}000$ to balance within-cluster similarity and the availability of image groups.

\paragraph{UIC Scenario Construction}
For UIC, we sample four images within a cluster to ensure semantic similarity. To avoid trivial cases, we maximize in-cluster dispersion by selecting images with the largest pairwise distances within each cluster. The images share common themes, such as street scenes and indoor environments, while retaining enough variation to reveal cross-image content confusion.

\paragraph{DIC Scenario Construction}
For DIC scenarios, we sample four images from different clusters with large inter-cluster distances in the embedding space. This increases visual and semantic diversity, with varied content (e.g., outdoor scenes, indoor scenes, food, and vehicles). Such diversity enables evaluation of context interference patterns where competing visual information may lead to hallucination.

\begin{figure*}[!t]
\centering
\includegraphics[width=0.95\textwidth]{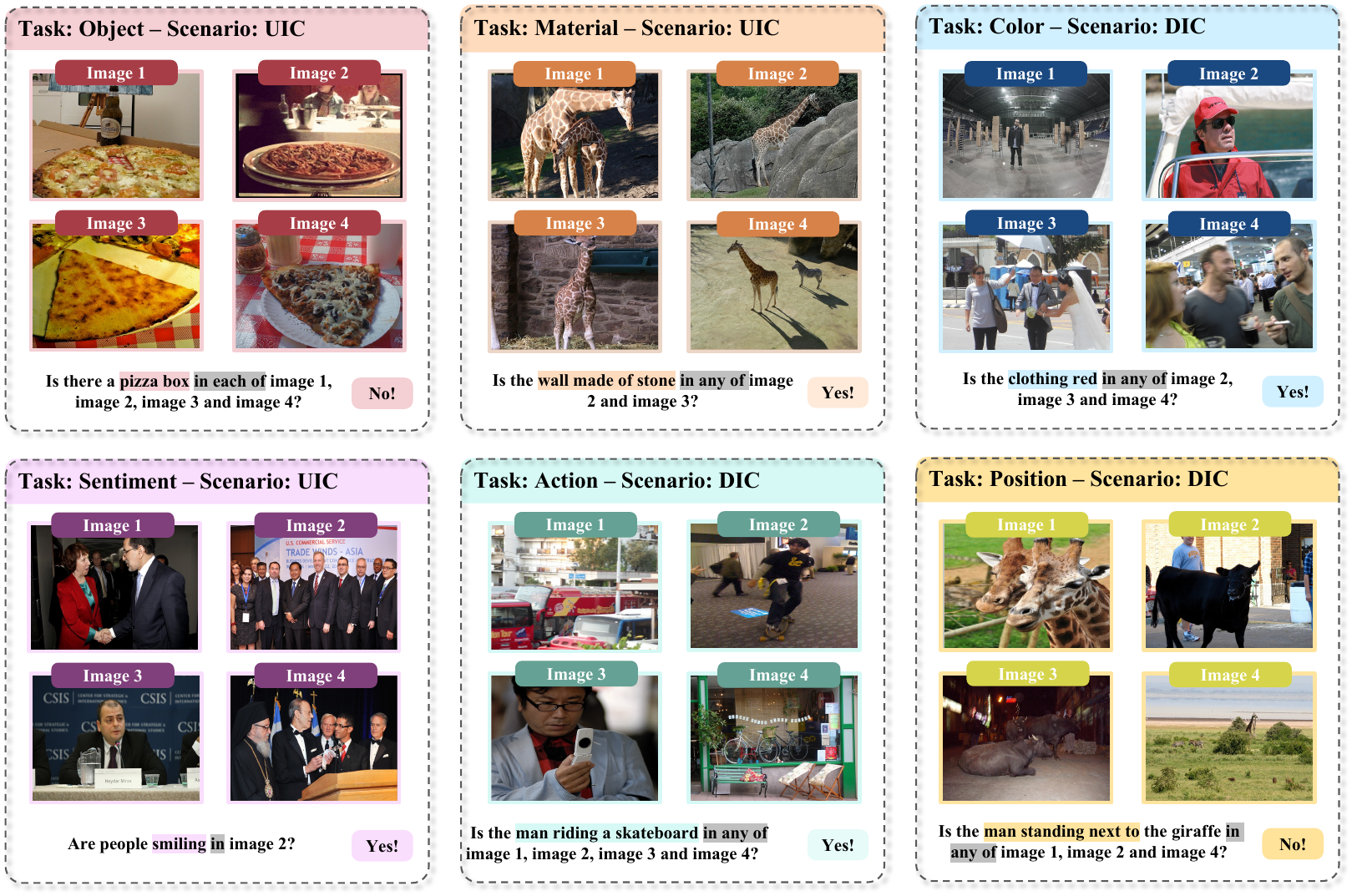}
\caption{Representative LIVE items across all six visual recognition tasks.}
\label{fig:generate_cases}
\end{figure*}

\subsection{Stage 2: Task-Specific Multi-Image Hallucination Item Mining}
\label{subsec:mihitem_mining}
To support hallucination evaluation across cognitive levels, we develop a method to mine MHI items that serve as controlled probes of factual grounding in LVLMs. These items are designed to be semantically plausible and contextually coherent in the given multi-image context while remaining absent from the actual visual content, making them suitable for identifying cross-image content-transfer errors.
We also use human annotation to assess label reliability. Section~\ref{subsec:construction_validation} summarizes the validation evidence. The annotation criteria and validation procedure are described in Appendix~\ref{sec:human_val}.

\paragraph{Multi-Level Visual Content Extraction}
As illustrated in Fig.~\ref{fig:pipeline}, we first apply a scene graph (SG) parser~\citep{li2023factual} to image captions\footnote{MS-COCO dataset~\citep{lin2014microsoft} provides human-annotated captions that describe the content of each image, serving as the input to our scene graph parser.} to extract object-attribute-relationship triplets spanning six visual recognition tasks: \textit{Objects}, \textit{Material}, \textit{Color}, \textit{Sentiment}, \textit{Action}, and \textit{Position}. For each image group, we merge the individual scene graphs into a unified SG and compute either the intersection or the union of visual items, depending on whether common items exist. The intersection captures factual items shared across images, whereas the union preserves items that appear only in subsets of images. Together, these item sets serve as anchors for constructing controlled hallucination probes targeting cross-image confusion.

\paragraph{LLM-Assisted Counterfactual Generation}
Conditioned on the merged scene graph and a chosen visual anchor (object, attribute, or relation), we prompt an LLM~\footnote{We use Qwen2.5-72B-Instruct~\citep{qwen2025qwen25technicalreport}. Prompting templates are provided in Appendix~\ref{app:prompt_templates}.} to generate candidate MHI items that are semantically related to the image group but absent from the target images. This approach leverages the LLM's world knowledge to propose plausible yet incorrect visual content. To improve contextual plausibility, the prompt also incorporates the semantic theme of the image group, encouraging the generated items to remain believable within the corresponding multi-image scenarios. For example, in a street-scene UIC scenario, the model may generate objects such as ``traffic light'' or ``crosswalk,'' which are contextually plausible but not visually present.

\subsection{Stage 3: Multi-Granularity Question Generation}
\label{subsec:question_generation}
To address incomplete evaluation coverage (challenge \ding{183}), MAP uses questions targeting different image subset sizes. In the primary evaluation, all four images are provided at every granularity. Only the number of queried target images varies.

\paragraph{Granularity-Based Question Templates}
For an item (denoted as \mysquare), we use four question template types targeting 1 to 4 images (denoted as \texttt{TI$_1$} to \texttt{TI$_4$}). To maintain linguistic consistency, we employ templates~\footnote{These templates provide few-shot examples for generation. Actual questions incorporate linguistic variations, such as adding ``there'' before \mysquare.}:
\ding{182} \texttt{TI$_1$}: Is \mysquare~in image A?
\ding{183} \texttt{TI$_2$}: Is \mysquare~in each/any of images A and B?
\ding{184} \texttt{TI$_3$}: Is \mysquare~in each/any of images A, B and C?
\ding{185} \texttt{TI$_4$}: Is \mysquare~in each/any of images A, B, C and D?

\paragraph{Balanced Binary Question Construction}
We construct yes/no questions from counterfactual MHI items and verified factual items. An ``any'' question is true if the item is present in at least one target image, whereas an ``each'' question is true only if it is present in every target image. At granularity TI$_g$ where $g \in \{1,2,3,4\}$, there are $\mathrm{C}_{4}^{g}$ possible target-image combinations. The retained questions cover these combinations across the dataset. Fig.~\ref{fig:generate_cases} illustrates all six tasks.

\begin{figure}[!t]
\centering
\subfloat[Distribution of UIC.\label{fig:UIC}]{\includegraphics[width=0.48\columnwidth]{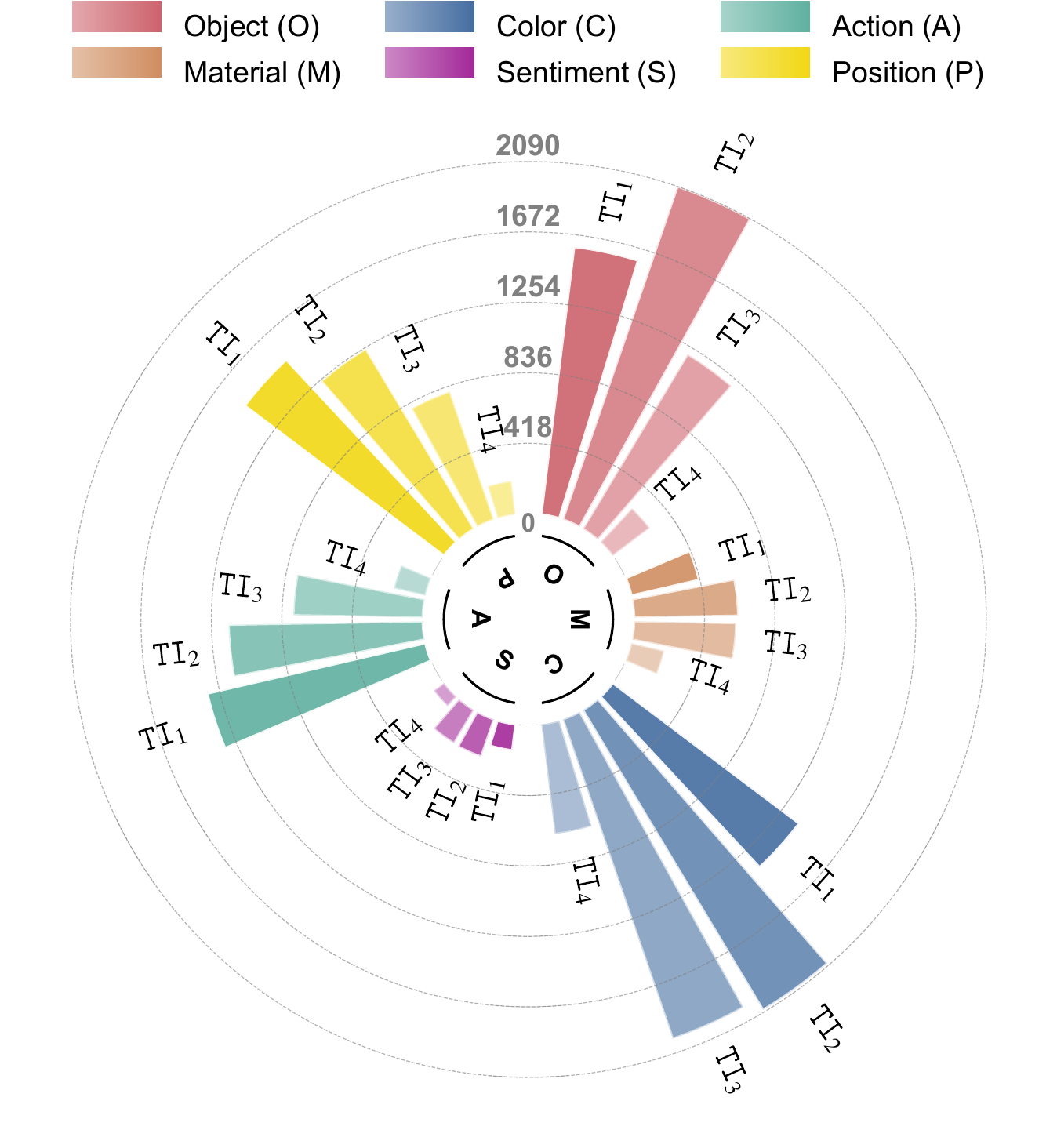}}
\hfill
\subfloat[Distribution of DIC.\label{fig:DIC}]{\includegraphics[width=0.48\columnwidth]{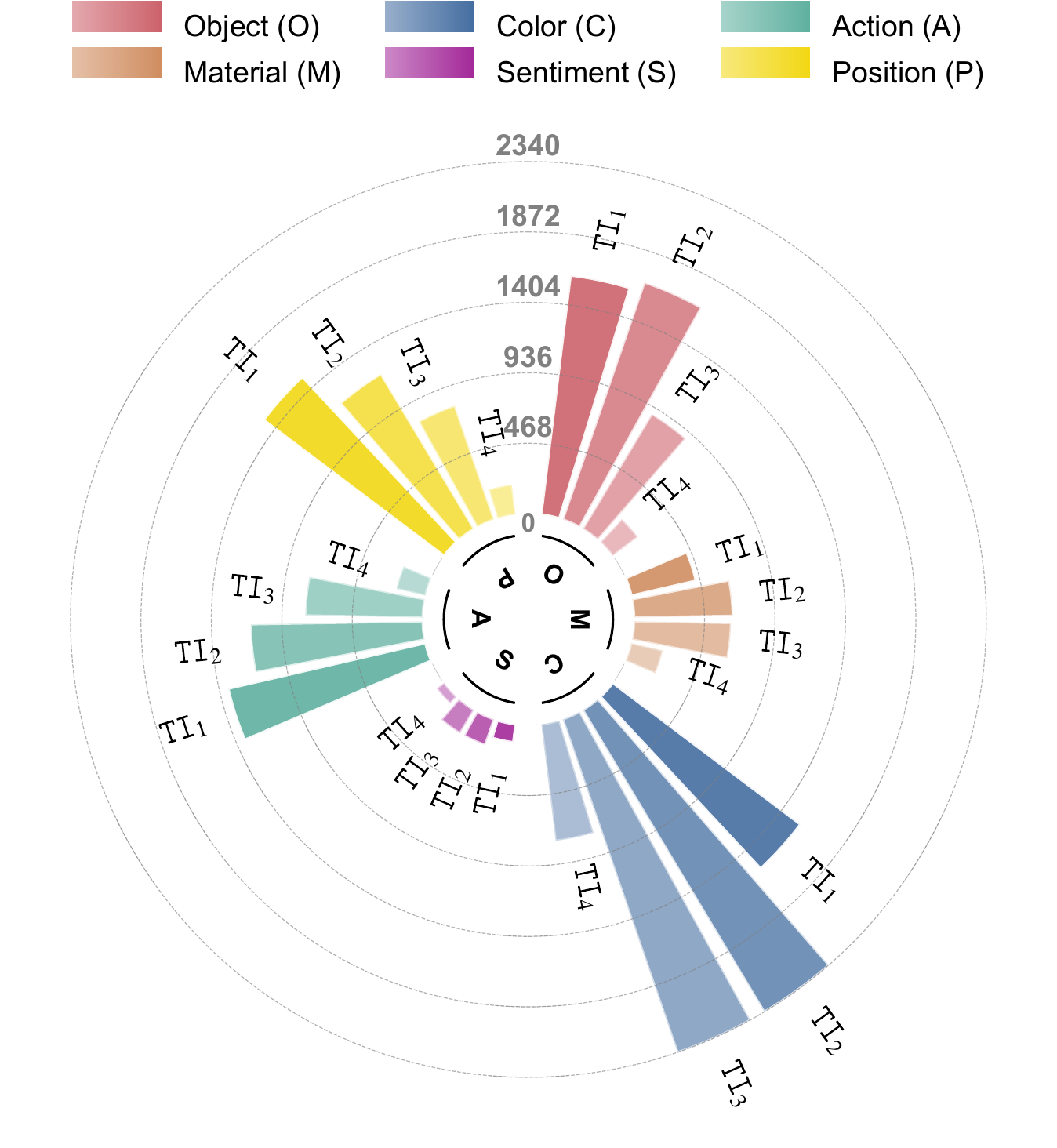}}
\caption{Task distributions in UIC and DIC partitions of LIVE.}
\label{fig:sample_dist}
\end{figure}

\paragraph{Multi-Granularity Assessment Protocol}
Our MAP metric aggregates performance across all granularity levels to evaluate multi-image understanding, and is defined as follows:
\begin{equation}
\text{MAP Score} = \frac{1}{4} \sum_{g=1}^{4} \frac{1}{\mathrm{C}_{4}^{g}} \sum_{i=1}^{\mathrm{C}_{4}^{g}} \text{Metric}_{g,i},
\end{equation}
where $\text{Metric}_{g,i}$ is accuracy or F1 computed over all questions for the $i$-th target-image combination at granularity $g$, within a task and context.

\paragraph{Aggregation Rationale and Reporting}
The inner average assigns equal weight to every target-image combination within a granularity, whereas the outer average assigns equal weight to the four granularities. The numbers of subsets are unequal: a four-image scenario contains four singleton subsets, six pairs, four triples, and one full set. Directly averaging all 15 combinations would emphasize two-image queries simply because they are more numerous. MAP instead asks whether a model remains reliable across reasoning scopes, with each scope contributing one quarter of the final score. We pair positive and negative questions within each task-scenario partition. We report both accuracy and F1, treating `Yes' as the positive class.

MAP should therefore be read together with its granularity profile rather than as an isolated scalar. A high score accompanied by a steep late-stage decline indicates a different failure pattern from a similar score with moderate errors at every granularity. Conversely, a \texttt{TI$_4$}-only score observes just the unique full-set target-image combination and cannot reliably show whether the model is already unreliable on pairs or triples. For this reason, the main table provides task- and context-specific MAP values, while the model-wise curves expose the underlying performance trajectories. Complete values at $T=0$, $T=0.7$, and $T=1.0$ are provided in Appendix~\ref{sec:appendix_results}, with the model-wise granularity profiles in Appendix~\ref{sec:full_multi_granularity}. These results show performance differences across models and granularities.

\subsection{Dataset Overview}
\label{subsec:dataset_overview}

LIVE comprises 488 multi-image scenarios (242 UIC + 246 DIC) drawn from 1,871 daily-life images, paired with 42,038 binary visual question-answer triplets across six visual recognition tasks and four image granularities. Fig.~\ref{fig:sample_dist} presents the sample distribution across tasks, and Table~\ref{tab:benchmark_comparison} compares LIVE with related benchmarks. LIVE evaluates UIC and DIC under a shared construction pipeline across six tasks spanning low-level perception to high-level cognition. MAP evaluates every target-subset size in the four-image setting, extending evaluation beyond the full image set.

\subsection{Validation of Benchmark Construction}
\label{subsec:construction_validation}
For our LLM-assisted pipeline, we validate the caption-derived representations and the counterfactual items before question generation. In a random sample of 100 caption-derived scene graphs, no label-affecting error is observed. Four experts then assess items associated with 500 images for visual absence and contextual plausibility, yielding 98.5\% agreement and Cohen's $\kappa=0.92$; 4.2\% of candidates are rejected or revised before question generation. Validation results, annotation criteria, task-specific checks, and the comparison with image-grounded graphs are provided in Appendix~\ref{sec:human_val}.

\begin{table*}[!t]
\centering
\footnotesize
\renewcommand{\arraystretch}{1.1}
\setlength{\tabcolsep}{2.8pt}
\caption{Overall performance with MAP. Full model-wise granularity results are provided in Appendix~\ref{sec:appendix_results}.}
\label{tab:main_result}
\begin{adjustbox}{max width=\linewidth}
\begin{threeparttable}
\begin{tabular}{lcccccccccccccccccccccc}
    \toprule
    \scalea{\textbf{Task}}
    & \multicolumn{2}{c}{\textbf{\scalea{GPT-4o}}}
    & \multicolumn{2}{c}{\textbf{\scalea{Qwen3-VL-32B}}}
    & \multicolumn{2}{c}{\textbf{\scalea{InternVL-2.5-26B}}}
    & \multicolumn{2}{c}{\textbf{\scalea{Qwen3-VL-8B}}}
    & \multicolumn{2}{c}{\textbf{\scalea{InternVL-2.5-8B}}}
    & \multicolumn{2}{c}{\textbf{\scalea{MiniCPM-V4.5}}}
    & \multicolumn{2}{c}{\textbf{\scalea{MiniCPM-V2.6}}}
    & \multicolumn{2}{c}{\textbf{\scalea{Mantis-8B}}}
    & \multicolumn{2}{c}{\textbf{\scalea{Qwen2.5-VL-7B}}}
    & \multicolumn{2}{c}{\textbf{\scalea{LLaVA-OV}}}
    & \multicolumn{2}{c}{\textbf{\scalea{LLaVA-IL}}} \\
    \cmidrule(lr){2-3} \cmidrule(lr){4-5} \cmidrule(lr){6-7} \cmidrule(lr){8-9} \cmidrule(lr){10-11} \cmidrule(lr){12-13} \cmidrule(lr){14-15} \cmidrule(lr){16-17} \cmidrule(lr){18-19} \cmidrule(lr){20-21} \cmidrule(lr){22-23}
    \scalea{\textbf{MAP}} &  \scalea{F1} & \scalea{Acc} &  \scalea{F1} & \scalea{Acc} &  \scalea{F1} & \scalea{Acc} &  \scalea{F1} & \scalea{Acc} &  \scalea{F1} & \scalea{Acc} &  \scalea{F1} & \scalea{Acc} &  \scalea{F1} & \scalea{Acc} &  \scalea{F1} & \scalea{Acc} &  \scalea{F1} & \scalea{Acc} &  \scalea{F1} & \scalea{Acc} &  \scalea{F1} & \scalea{Acc}  \\
    \midrule
    \rowcolor[HTML]{f0f0f0}
     \multicolumn{23}{c}{\textit{\textbf{Scenario 1: Uniform Image Context}}} \\
    \scalea{Object} & \scaleb{0.974} & \scaleb{0.974} & \scaleb{\bst{0.963}} & \scaleb{\bst{0.964}} & \scaleb{0.912} & \scaleb{0.918} & \scaleb{0.936} & \scaleb{0.941} & \scaleb{0.864} & \scaleb{0.881} & \scaleb{0.942} & \scaleb{\subbst{0.945}} & \scaleb{0.860} & \scaleb{0.879} & \scaleb{0.876} & \scaleb{0.885} & \scaleb{0.792} & \scaleb{0.830} & \scaleb{0.924} & \scaleb{0.928} & \scaleb{\subbst{0.946}} & \scaleb{0.944} \\
    \scalea{Material} & \scaleb{0.918} & \scaleb{0.921} & \scaleb{\bst{0.897}} & \scaleb{\bst{0.906}} & \scaleb{0.854} & \scaleb{0.870} & \scaleb{0.875} & \scaleb{0.886} & \scaleb{0.835} & \scaleb{0.858} & \scaleb{0.868} & \scaleb{0.880} & \scaleb{0.729} & \scaleb{0.791} & \scaleb{0.881} & \scaleb{0.882} & \scaleb{0.447} & \scaleb{0.663} & \scaleb{\subbst{0.896}} & \scaleb{\subbst{0.903}} & \scaleb{0.886} & \scaleb{0.875} \\
    \scalea{Color} & \scaleb{0.934} & \scaleb{0.937} & \scaleb{\bst{0.925}} & \scaleb{\bst{0.931}} & \scaleb{0.865} & \scaleb{0.879} & \scaleb{0.891} & \scaleb{\subbst{0.904}} & \scaleb{0.859} & \scaleb{0.875} & \scaleb{0.881} & \scaleb{0.893} & \scaleb{0.706} & \scaleb{0.784} & \scaleb{0.870} & \scaleb{0.868} & \scaleb{0.503} & \scaleb{0.686} & \scaleb{0.872} & \scaleb{0.885} & \scaleb{\subbst{0.899}} & \scaleb{0.890} \\
    \scalea{Sentiment} & \scaleb{0.844} & \scaleb{0.866} & \scaleb{\bst{0.942}} & \scaleb{\bst{0.946}} & \scaleb{0.901} & \scaleb{0.907} & \scaleb{0.889} & \scaleb{0.908} & \scaleb{0.899} & \scaleb{0.909} & \scaleb{0.892} & \scaleb{0.899} & \scaleb{0.795} & \scaleb{0.830} & \scaleb{0.865} & \scaleb{0.853} & \scaleb{0.693} & \scaleb{0.774} & \scaleb{\subbst{0.913}} & \scaleb{\subbst{0.918}} & \scaleb{0.896} & \scaleb{0.886} \\
    \scalea{Action} & \scaleb{0.833} & \scaleb{0.859} & \scaleb{0.825} & \scaleb{0.852} & \scaleb{0.843} & \scaleb{0.861} & \scaleb{0.759} & \scaleb{0.809} & \scaleb{0.761} & \scaleb{0.807} & \scaleb{\subbst{0.848}} & \scaleb{\subbst{0.868}} & \scaleb{0.603} & \scaleb{0.728} & \scaleb{0.801} & \scaleb{0.817} & \scaleb{0.424} & \scaleb{0.647} & \scaleb{0.846} & \scaleb{0.864} & \scaleb{\bst{0.887}} & \scaleb{\bst{0.886}} \\
    \scalea{Position} & \scaleb{0.771} & \scaleb{0.818} & \scaleb{\subbst{0.821}} & \scaleb{\subbst{0.851}} & \scaleb{0.818} & \scaleb{0.842} & \scaleb{0.773} & \scaleb{0.820} & \scaleb{0.750} & \scaleb{0.801} & \scaleb{0.811} & \scaleb{0.842} & \scaleb{0.547} & \scaleb{0.700} & \scaleb{0.793} & \scaleb{0.803} & \scaleb{0.414} & \scaleb{0.642} & \scaleb{0.789} & \scaleb{0.821} & \scaleb{\bst{0.867}} & \scaleb{\bst{0.857}} \\
    \midrule
    \rowcolor[HTML]{f0f0f0}
    \multicolumn{23}{c}{\textit{\textbf{Scenario 2: Diverse Image Context}}} \\
    \scalea{Object} & \scaleb{0.971} & \scaleb{0.971} & \scaleb{\bst{0.963}} & \scaleb{\bst{0.963}} & \scaleb{0.815} & \scaleb{0.843} & \scaleb{0.904} & \scaleb{0.914} & \scaleb{0.760} & \scaleb{0.808} & \scaleb{0.893} & \scaleb{0.904} & \scaleb{0.712} & \scaleb{0.790} & \scaleb{0.809} & \scaleb{0.831} & \scaleb{0.567} & \scaleb{0.716} & \scaleb{0.837} & \scaleb{0.859} & \scaleb{\subbst{0.921}} & \scaleb{\subbst{0.919}} \\
    \scalea{Material} & \scaleb{0.930} & \scaleb{0.934} & \scaleb{\bst{0.879}} & \scaleb{\bst{0.895}} & \scaleb{0.834} & \scaleb{0.853} & \scaleb{0.838} & \scaleb{0.864} & \scaleb{0.741} & \scaleb{0.796} & \scaleb{0.831} & \scaleb{0.860} & \scaleb{0.588} & \scaleb{0.723} & \scaleb{0.837} & \scaleb{0.840} & \scaleb{0.400} & \scaleb{0.650} & \scaleb{\bst{0.879}} & \scaleb{\subbst{0.887}} & \scaleb{\subbst{0.855}} & \scaleb{0.849} \\
    \scalea{Color} & \scaleb{0.930} & \scaleb{0.934} & \scaleb{\bst{0.919}} & \scaleb{\bst{0.925}} & \scaleb{0.790} & \scaleb{0.830} & \scaleb{0.823} & \scaleb{0.855} & \scaleb{0.841} & \scaleb{0.863} & \scaleb{0.814} & \scaleb{0.846} & \scaleb{0.562} & \scaleb{0.717} & \scaleb{0.846} & \scaleb{0.850} & \scaleb{0.339} & \scaleb{0.635} & \scaleb{0.893} & \scaleb{0.904} & \scaleb{\subbst{0.912}} & \scaleb{\subbst{0.906}} \\
    \scalea{Sentiment} & \scaleb{0.737} & \scaleb{0.798} & \scaleb{\subbst{0.854}} & \scaleb{\bst{0.864}} & \scaleb{0.807} & \scaleb{0.828} & \scaleb{0.758} & \scaleb{0.801} & \scaleb{0.784} & \scaleb{0.816} & \scaleb{0.785} & \scaleb{0.802} & \scaleb{0.693} & \scaleb{0.776} & \scaleb{0.815} & \scaleb{0.824} & \scaleb{0.424} & \scaleb{0.644} & \scaleb{0.832} & \scaleb{0.840} & \scaleb{\bst{0.862}} & \scaleb{\subbst{0.856}} \\
    \scalea{Action} & \scaleb{0.786} & \scaleb{0.831} & \scaleb{\subbst{0.805}} & \scaleb{\subbst{0.843}} & \scaleb{0.740} & \scaleb{0.798} & \scaleb{0.669} & \scaleb{0.765} & \scaleb{0.679} & \scaleb{0.760} & \scaleb{0.804} & \scaleb{0.841} & \scaleb{0.404} & \scaleb{0.653} & \scaleb{0.723} & \scaleb{0.764} & \scaleb{0.258} & \scaleb{0.600} & \scaleb{0.734} & \scaleb{0.793} & \scaleb{\bst{0.874}} & \scaleb{\bst{0.879}} \\
    \scalea{Position} & \scaleb{0.728} & \scaleb{0.793} & \scaleb{\subbst{0.743}} & \scaleb{\subbst{0.804}} & \scaleb{0.713} & \scaleb{0.775} & \scaleb{0.643} & \scaleb{0.749} & \scaleb{0.622} & \scaleb{0.726} & \scaleb{0.716} & \scaleb{0.786} & \scaleb{0.370} & \scaleb{0.636} & \scaleb{0.722} & \scaleb{0.755} & \scaleb{0.237} & \scaleb{0.588} & \scaleb{0.686} & \scaleb{0.761} & \scaleb{\bst{0.835}} & \scaleb{\bst{0.832}} \\
    \bottomrule
\end{tabular}
\begin{tablenotes}
    \item  \scriptsize \textit{Note}:  \bst{Bold} and \subbst{underlined} denote best and second-best results (excluding GPT-4o), respectively.
\end{tablenotes}
\end{threeparttable}
\end{adjustbox}
\end{table*}

\begin{figure}[!t]
\centering

\subfloat[
    UIC example.
    \label{fig:case_uic}
]{\includegraphics[width=0.95\linewidth]{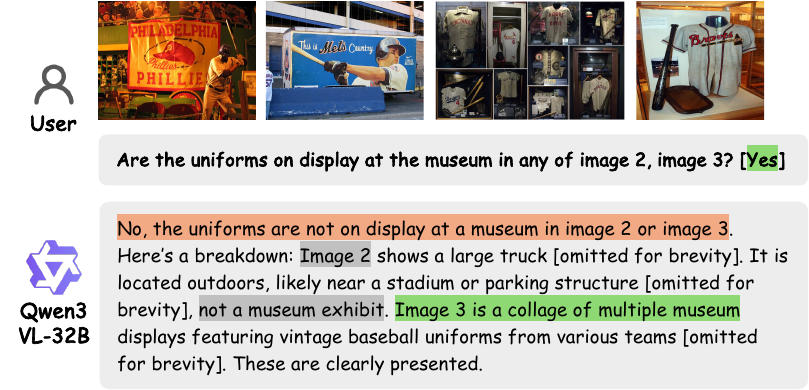}
}

\subfloat[
    DIC example.
    \label{fig:case_dic}
]{\includegraphics[width=0.95\linewidth]{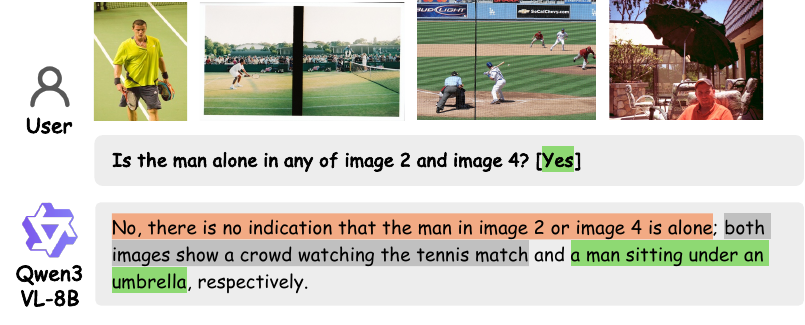}
}

\caption{Examples of incorrect binary answers under UIC and DIC.
\cbox{myorange} denotes factual errors,
\cbox{mygray} denotes conflicts with the target image,
and \cbox{mygreen} denotes correct target-image details.}
\label{fig:case_study}
\end{figure}

\section{Experiments}
We evaluate context-dependent hallucination patterns and the value of MAP across 11 LVLMs.

\subsection{Experimental Setup}

\paragraph{Model Selection}
Following OpenCompass Multi-modal Academic Leaderboard~\citep{duan2024vlmevalkit}, we select ten open-source LVLMs: Qwen3-VL-32B/8B~\citep{Qwen3VL}, MiniCPM-V4.5/2.6~\citep{yu2025minicpm,yao2024minicpm}, Qwen2.5-VL-7B~\citep{bai2025qwen2}, InternVL-2.5-26B/8B~\citep{chen2024expanding}, LLaVA-OV~\citep{lillava2}, LLaVA-IL~\citep{lillava1}, and Mantis-8B~\citep{jiang2024MANTISIM}.
GPT-4o~\citep{hurst2024gpt} provides a closed-source baseline.

\paragraph{Evaluation Protocol}
In the primary binary evaluation, multi-image questions are formatted using the template: \texttt{"<|images|> \{question\}. Answer with `Yes' or `No' first, and then provide your reasoning."}, where images are presented in sequence followed by the question. Open-source models are deployed using vLLM~\citep{kwon2023efficient} with temperature $T=0$ for deterministic outputs. Sensitivity to higher temperatures is analyzed in Section~\ref{sec:robustness}. All models are evaluated on the complete LIVE dataset. Model responses are parsed to extract binary answers.

\subsection{Overall Evaluation on LIVE}
\label{subsec:overall_evaluation}
\begin{figure}[!t]
\centering
\subfloat[
    Accuracy of InternVL-2.5-26B under UIC and DIC.
    \label{fig:InternVL2.5_26B_AAAA_acc}
]{\includegraphics[width=0.48\linewidth]{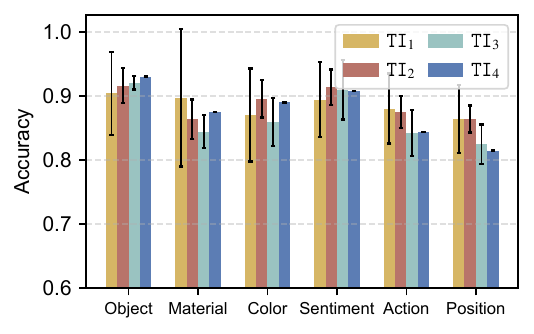}\hspace{3pt}
    \includegraphics[width=0.48\linewidth]{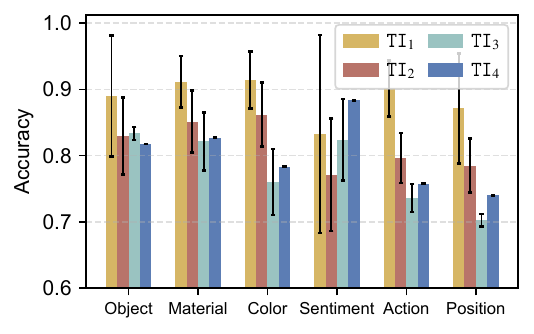}
}

\subfloat[
    Accuracy of MiniCPM-V2.6 under UIC and DIC.
    \label{fig:MiniCPM_2_6_AAAA_acc}
]{\includegraphics[width=0.48\linewidth]{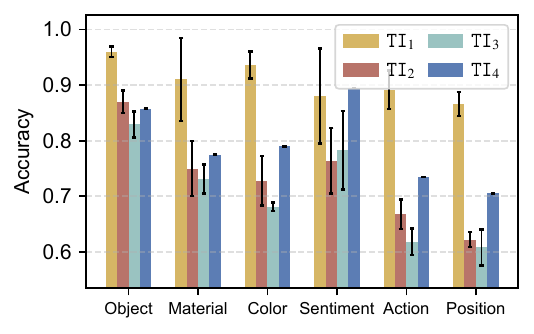}\hspace{3pt}
    \includegraphics[width=0.48\linewidth]{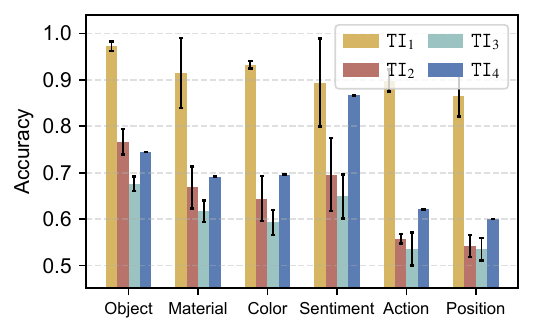}
}
\caption{Multi-granularity evaluation results of
InternVL-2.5-26B and MiniCPM-V2.6.
Additional profiles at all three temperatures are provided in Appendix~\ref{sec:full_multi_granularity}.}
\label{fig:multi-granularity}
\end{figure}

Table~\ref{tab:main_result} summarizes the MAP performance with respect to F1-score and accuracy of 11 LVLMs across different task-scenario combinations. The results confirm our identified challenges regarding context-dependent hallucination and incomplete evaluation coverage, yielding four key observations:
\ding{182} \textbf{Context diversity generally increases hallucination rates.}
Across most open-source models and tasks, DIC is more error-prone than UIC, showing that hallucination behavior varies with visual context. For instance, on the object task, InternVL-2.5-26B's F1 drops from 0.912 (UIC) to 0.815 (DIC), while Qwen2.5-VL-7B experiences a more severe drop from 0.792 to 0.567. Similar degradations occur across material, color, sentiment, and position tasks for mid-tier models, whereas stronger models such as Qwen3-VL-32B and LLaVA-IL demonstrate greater robustness to context variations.
\ding{183} \textbf{LIVE reveals distinct vulnerability patterns across visual recognition tasks.}
Object recognition emerges as the most accessible capability under UIC, with strong models achieving F1 scores above 0.90. In contrast, material and color recognition exhibit much larger variance: weaker models only reach F1 scores around 0.33--0.73, showing uneven factual grounding across architectures. Sentiment, action, and position recognition are often more challenging, with several models falling below 0.50 F1 in DIC. Notably, GPT-4o still exhibits a noticeable performance decay on the sentiment task, dropping from an F1 of 0.844 in UIC to 0.737 in DIC. This indicates that multi-image contextual and relational reasoning remains challenging even for state-of-the-art LVLMs.
\ding{184} \textbf{Multi-granularity evaluation exposes hidden failure modes overlooked by single-query assessment.}
As shown in Fig.~\ref{fig:multi-granularity}, accuracy varies across target-image subset sizes and often recovers at \texttt{TI$_4$} after declining at intermediate granularities. These profiles expose failures that full-set evaluation alone can overlook. Different models also exhibit distinct profiles: some maintain relatively stable performance, whereas others show sharp declines at \texttt{TI$_2$} or \texttt{TI$_3$}.
\ding{185} \textbf{MAP provides a more faithful characterization than conventional metrics.}
Evaluating models solely on \texttt{TI$_4$} resembles traditional single-query protocols and often obscures performance differences. By capturing performance across all subset sizes, MAP reveals that models with similar \texttt{TI$_4$} accuracy can yield different scores due to varying reliability on smaller image combinations. MAP F1 rankings change 75 of 132 model-rank entries across the 12 task-scenario combinations compared with \texttt{TI$_4$}-only evaluation, demonstrating that full-set querying alone can provide an incomplete characterization of model robustness.

\paragraph{Showcase}
Fig.~\ref{fig:case_study} presents representative cases showing how Qwen3-VL models exhibit distinct hallucination patterns under UIC and DIC. In UIC, Qwen3-VL-32B describes the museum display in Image~3 but answers ``No'' to a question asking whether the uniforms are displayed in a museum in any target image. Its answer conflicts with its explanation. In DIC, Qwen3-VL-8B answers ``No'' when asked whether a man is alone in either target image, despite the reference answer ``Yes''. These cases illustrate content confusion in UIC and context interference in DIC.

\subsection{Ablation Studies}
To understand hallucination, we conduct ablation studies on image quantity and presentation order. For image quantity, we focus on \texttt{TI$_1$} and \texttt{TI$_2$}, where @$k$ indicates that the total number of input images is reduced to $k$ by removing $(4-k)$ non-target images while preserving all target images. For image order, we examine \texttt{TI$_3$} and \texttt{TI$_4$} by randomly shuffling the target images while retaining the images and questions in UIC and DIC.

\begin{table}[t]
\centering
\scriptsize
\caption{Ablation study results.}
\label{tab:ablation_result}
\setlength\tabcolsep{1.4pt}
\renewcommand{\arraystretch}{1.08}
\begin{adjustbox}{max width=\linewidth}
\begin{threeparttable}
\begin{tabular}{lcccccccccc}
    \toprule
    \textbf{Setting}
    & \multicolumn{2}{c}{\textbf{\shortstack{InternVL-2.5\\26B}}}
    & \multicolumn{2}{c}{\textbf{\shortstack{InternVL-2.5\\8B}}}
    & \multicolumn{2}{c}{\textbf{Mantis-8B}}
    & \multicolumn{2}{c}{\textbf{LLaVA-OV}}
    & \multicolumn{2}{c}{\textbf{LLaVA-IL}} \\
    \cmidrule(lr){2-3} \cmidrule(lr){4-5} \cmidrule(lr){6-7} \cmidrule(lr){8-9} \cmidrule(lr){10-11}
    \textbf{Metrics} &  F1 & Acc &  F1 & Acc &  F1 & Acc &  F1 & Acc &  F1 & Acc  \\
    \midrule
    \rowcolor[HTML]{f0f0f0}
    \multicolumn{11}{c}{\textit{\textbf{Scenario 1: Uniform Image Context}}} \\
    \multicolumn{11}{l}{\textit{\textbf{Ablation of Image Quantity Effects}}} \\
    \texttt{TI$_1$@4} & 0.870 & 0.885 & 0.801 & 0.838 & 0.859 & 0.845 & 0.884 & 0.893 & 0.884 & 0.877 \\
    \texttt{TI$_1$@3} & \reduce{0.920\upa} & \reduce{0.925\upa} & \reduce{0.844\upa} & \reduce{0.866\upa} & \reduce{0.876\upa} & \reduce{0.864\upa} & \reduce{0.908\upa} & \reduce{0.913\upa} & \reduce{0.893\upa} & \reduce{0.888\upa} \\
    \texttt{TI$_1$@2} & \reduce{0.937\upa} & \reduce{0.941\upa} & \reduce{0.897\upa} & \reduce{0.906\upa} & \reduce{0.890\upa} & \reduce{0.881\upa} & \reduce{0.934\upa} & \reduce{0.936\upa} & \reduce{0.902\upa} & \reduce{0.896\upa} \\
    \texttt{TI$_2$@4} & 0.881 & 0.888 & 0.777 & 0.817 & 0.823 & 0.833 & 0.841 & 0.861 & 0.883 & 0.871 \\
    \texttt{TI$_2$@3} & \reduce{0.890\upa} & \reduce{0.897\upa} & \reduce{0.796\upa} & \reduce{0.831\upa} & \reduce{0.845\upa} & \reduce{0.852\upa} & \reduce{0.858\upa} & \reduce{0.874\upa} & \reduce{0.892\upa} & \reduce{0.885\upa} \\
    \hdashline
    \multicolumn{11}{l}{\textit{\textbf{Ablation of Image Order Effects}}} \\
    \texttt{TI$_3$} & 0.848 & 0.867 & 0.852 & 0.872 & 0.842 & 0.852 & 0.875 & 0.889 & 0.904 & 0.898 \\
    \texttt{TI$_3^\dagger$} & \reduce{0.850\upa} & \reduce{0.869\upa} & \reduce{0.856\upa} & \reduce{0.875\upa} & \reduce{0.843\upa} & \reduce{0.853\upa} & \reduce{0.883\upa} & \reduce{0.895\upa} & \reduce{0.904} & \reduce{0.899\upa} \\
    \texttt{TI$_4$} & 0.863 & 0.877 & 0.881 & 0.893 & 0.867 & 0.875 & 0.894 & 0.903 & 0.915 & 0.912 \\
    \texttt{TI$_4^\dagger$} & \reduce{0.866\upa} & \reduce{0.881\upa} & \reduce{0.883\upa} & \reduce{0.895\upa} & \reduce{0.873\upa} & \reduce{0.881\upa} & \reduce{0.903\upa} & \reduce{0.911\upa} & \reduce{0.910\dna} & \reduce{0.906\dna} \\
    \midrule
    \rowcolor[HTML]{f0f0f0}
    \multicolumn{11}{c}{\textit{\textbf{Scenario 2: Diverse Image Context}}} \\
    \multicolumn{11}{l}{\textit{\textbf{Ablation of Image Quantity Effects}}} \\
    \texttt{TI$_1$@4} & 0.871 & 0.887 & 0.800 & 0.835 & 0.853 & 0.842 & 0.893 & 0.896 & 0.877 & 0.870 \\
    \texttt{TI$_1$@3} & \reduce{0.926\upa} & \reduce{0.929\upa} & \reduce{0.846\upa} & \reduce{0.868\upa} & \reduce{0.876\upa} & \reduce{0.869\upa} & \reduce{0.921\upa} & \reduce{0.923\upa} & \reduce{0.886\upa} & \reduce{0.882\upa} \\
    \texttt{TI$_1$@2} & \reduce{0.937\upa} & \reduce{0.939\upa} & \reduce{0.893\upa} & \reduce{0.906\upa} & \reduce{0.898\upa} & \reduce{0.890\upa} & \reduce{0.933\upa} & \reduce{0.933\upa} & \reduce{0.907\upa} & \reduce{0.901\upa} \\
    \texttt{TI$_2$@4} & 0.784 & 0.816 & 0.659 & 0.750 & 0.752 & 0.787 & 0.737 & 0.792 & 0.874 & 0.869 \\
    \texttt{TI$_2$@3} & \reduce{0.801\upa} & \reduce{0.832\upa} & \reduce{0.647\dna} & \reduce{0.744\dna} & \reduce{0.779\upa} & \reduce{0.806\upa} & \reduce{0.748\upa} & \reduce{0.800\upa} & \reduce{0.881\upa} & \reduce{0.879\upa} \\
    \hdashline
    \multicolumn{11}{l}{\textit{\textbf{Ablation of Image Order Effects}}} \\
    \texttt{TI$_3$} & 0.720 & 0.780 & 0.717 & 0.779 & 0.776 & 0.803 & 0.774 & 0.816 & 0.874 & 0.875 \\
    \texttt{TI$_3^\dagger$} & \reduce{0.708\dna} & \reduce{0.773\dna} & \reduce{0.721\upa} & \reduce{0.782\upa} & \reduce{0.773\dna} & \reduce{0.801\dna} & \reduce{0.774} & \reduce{0.814\dna} & \reduce{0.875\upa} & \reduce{0.876\upa} \\
    \texttt{TI$_4$} & 0.756 & 0.801 & 0.776 & 0.814 & 0.788 & 0.810 & 0.838 & 0.859 & 0.881 & 0.880 \\
    \texttt{TI$_4^\dagger$} & \reduce{0.740\dna} & \reduce{0.793\dna} & \reduce{0.783\upa} & \reduce{0.821\upa} & \reduce{0.772\dna} & \reduce{0.800\dna} & \reduce{0.838} & \reduce{0.857\dna} & \reduce{0.882\upa} & \reduce{0.881\upa} \\
    \bottomrule
\end{tabular}
\begin{tablenotes}
    \scriptsize
    \item  \textit{Note}: $^\dagger$ denotes random shuffling of targeted images.
\end{tablenotes}
\end{threeparttable}
\end{adjustbox}
\end{table}

\paragraph{Ablation of Image Quantity Effects}
Table~\ref{tab:ablation_result} presents how image quantity affects model performance. Key findings include:
\ding{182} \textbf{Reducing redundant visual context generally improves model performance.} Under UIC, InternVL-2.5-26B improves from an F1 of 0.870 in \texttt{TI$_1$@4} to 0.937 in \texttt{TI$_1$@2}. Under DIC, LLaVA-OV improves from 0.893 to 0.933. These gains show that performance is sensitive to additional non-target images.
\ding{183} \textbf{Multi-target settings exhibit similar sensitivity to image quantity.} When reducing the total input from four to three images in \texttt{TI$_2$}, most models also improve under DIC. For example, the average F1 score increases from 0.737 to 0.748 for LLaVA-OV and from 0.874 to 0.881 for LLaVA-IL.

\paragraph{Ablation of Image Order Effects}
Table~\ref{tab:ablation_result} reports how presentation order affects hallucination behaviors. Key findings include:
\ding{182} \textbf{Image presentation order has a marginal impact.} Comparing fixed and randomly shuffled variants (\texttt{TI$_3$} vs. \texttt{TI$_3^\dagger$} and \texttt{TI$_4$} vs. \texttt{TI$_4^\dagger$}), performance differences are typically within 1--2 F1 points and exhibit no systematic or consistent trend. For example, InternVL-2.5-26B obtains 0.848 versus 0.850 on \texttt{TI$_3$} under UIC, while LLaVA-OV remains at 0.774 on both variants under DIC.
\ding{183} \textbf{Order-independence supports the robustness of the evaluation design.} The negligible ordering effect suggests that the observed errors are more sensitive to image quantity and semantic relationships (UIC vs. DIC) than to presentation order. Together with the quantity ablation, these results show that MAP captures changes in contextual load and multi-image reasoning.

\paragraph{Cross-Model Ablation Summary}
Table~\ref{tab:ablation_result} and Appendix~\ref{app:full_ablation} together cover all 11 models. For single-target queries, reducing the input from four images to two improves F1 for most models (10 of 11 in UIC and 9 of 11 in DIC). The average F1 gain is $+0.027$ in each context, with maximum gains of $+0.096$ and $+0.093$, respectively. GPT-4o is one exception, decreasing by 0.005 and 0.018. Most models are sensitive to non-target visual context. For two-target queries, the changes are smaller and less consistent across contexts. This contrast suggests that two-target queries leave less removable context and already require cross-image aggregation, leaving less room for improvement.

Across the order ablation, the mean absolute F1 change under shuffling is 0.008, with a maximum absolute change of 0.030. The larger gains from removing non-target images show that image quantity has a stronger effect than presentation order.

\section{Robustness and Scope Analyses}
\label{sec:robustness}

The primary LIVE protocol uses balanced binary questions over four-image contexts. We further examine model-wise granularity profiles, decoding temperature, logical form, larger image sets, and non-binary response formats. Results and profiles are provided in Appendices~\ref{sec:appendix_results} and~\ref{sec:full_multi_granularity}. Results for logical forms, larger contexts, and non-binary formats are provided in Appendices~\ref{sec:appendix_any_each}--\ref{sec:extended_formats}.

\paragraph{Model-wise granularity profiles}
Fig.~\ref{fig:multi-granularity} shows two representative models. Additional $T=0$ profiles for the remaining nine models, including GPT-4o, are provided in Appendix~\ref{sec:full_multi_granularity}. Some profiles degrade early, whereas others remain stable until substantially larger target subsets. This variation strongly motivates MAP across all subset sizes rather than evaluation at a single granularity.

\paragraph{Decoding temperature}
The evaluation uses temperature $T=0$ for reproducibility. Averaged equally over the 11 models and 12 task--context combinations, MAP accuracy decreases from 0.834 at $T=0$ to 0.814 at $T=0.7$ and 0.798 at $T=1.0$. The mean across the 12 task--context combinations varies by at most 0.003 for GPT-4o and both Qwen3-VL checkpoints, whereas Mantis-8B, LLaVA-OV, and LLaVA-IL lose 0.098, 0.097, and 0.091, respectively, from $T=0$ to $T=1.0$.

\paragraph{Sensitivity to logical form}
Existential (``any'') and universal (``each'') queries require different aggregation. Table~\ref{tab:any_each_main} shows lower accuracy on ``any'' questions for every model at deterministic decoding, with gaps from $-0.052$ to $-0.332$. Results at $T=0.7$ and $T=1.0$ are provided in Appendix~\ref{sec:appendix_any_each}. The same pattern holds at both temperatures.

\begin{table}[!t]
\centering
\footnotesize
\caption{Accuracy under ``any'' and ``each'' logical forms at $T=0$. Difference denotes Any$-$Each.}
\label{tab:any_each_main}
\renewcommand{\arraystretch}{1.1}
\setlength{\tabcolsep}{11pt}
\begin{adjustbox}{max width=\linewidth}
\begin{tabular}{lccc}
\toprule
\textbf{Model} & \textbf{Existential} & \textbf{Universal} & \textbf{Difference} \\
\midrule
GPT-4o & \subbst{0.873} & \bst{0.983} & -0.110 \\
Qwen3-VL-32B & 0.871 & \subbst{0.981} & -0.110 \\
InternVL-2.5-26B & 0.833 & 0.933 & -0.100 \\
Qwen3-VL-8B & 0.808 & 0.957 & -0.149 \\
InternVL-2.5-8B & 0.808 & 0.912 & -0.104 \\
MiniCPM-V4.5 & 0.827 & 0.968 & -0.142 \\
MiniCPM-V2.6 & 0.662 & 0.947 & -0.285 \\
Mantis-8B & 0.819 & 0.871 & \bst{-0.052} \\
Qwen2.5-VL-7B & 0.588 & 0.921 & -0.332 \\
LLaVA-OV & 0.845 & 0.906 & \subbst{-0.061} \\
LLaVA-IL & \bst{0.882} & 0.975 & -0.093 \\
\bottomrule
\end{tabular}
\end{adjustbox}
\begin{tablenotes}
    \item  \scriptsize \textit{Note}:  \bst{Bold} and \subbst{underlined} denote best and second-best results, respectively.
\end{tablenotes}
\end{table}

\paragraph{Scaling to larger image contexts}
We extend LIVE to $k=5$--10 images by sampling target subsets because exhaustive enumeration is combinatorial. Each setting contains 4,250--7,062 binary questions covering all six visual tasks. From $k=5$ to $k=10$, accuracy decreases for nine of eleven models under UIC and all eleven under DIC. DIC is also harder for eight models at $k=5$ and nine at $k=10$. Despite local fluctuations, longer and more diverse contexts generally increase difficulty. The model-wise F1 and accuracy for all six context sizes in both contexts are provided in Appendix~\ref{sec:extended_k}.

\paragraph{Beyond binary questions}
Balanced yes/no questions permit exact scoring, but models may guess or favor one answer. We therefore derive Image Identification and Multiple-Choice QA with 3,000 samples each, and Open-Ended QA with 2,909, covering all six tasks. Construction and scoring details are provided in Appendix~\ref{sec:extended_formats}, with case studies in Appendix~\ref{sec:more_case_studies}. Fig.~\ref{fig:new_tasks_acc} shows that context-dependent differences persist beyond the binary format, although model behavior varies by task.

\begin{figure}[!t]
\centering
\includegraphics[width=0.94\columnwidth]{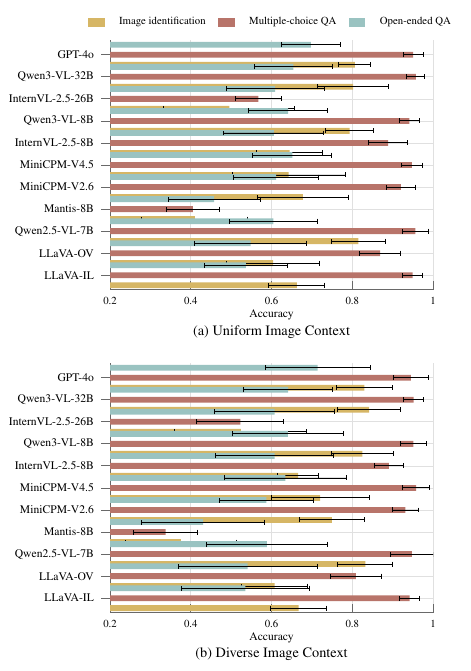}
\caption{Accuracy under Image Identification, Multiple-Choice QA, and Open-Ended QA for (a) UIC and (b) DIC. Bars show mean accuracy across the six visual tasks, and error bars show the standard deviation across tasks.}
\label{fig:new_tasks_acc}
\end{figure}

\section{Conclusions}
In this work, we tackle two fundamental challenges in multi-image visual hallucination evaluation for LVLMs: \ding{182} context-dependent hallucination patterns (UIC vs. DIC) and \ding{183} incomplete coverage of traditional single-query evaluation. To address these, we propose LIVE, an LLM-assisted benchmark providing balanced UIC/DIC coverage across six visual tasks, alongside a Multi-granularity Assessment Protocol that systematically probes hallucination behaviors across varying image subset sizes. Our extensive experiments on state-of-the-art LVLMs reveal significant context-dependent performance variations, demonstrating MAP's necessity for comprehensively evaluating multi-image understanding capabilities.

\section*{Limitations \& Future Work}
LIVE is constructed from MS-COCO daily-life scenes. While LIVE provides a robust evaluation framework—including our extensions to larger image sets and open-ended queries in the Appendix—our current scope primarily focuses on diagnosis rather than cure. Future work will explore hallucination mitigation strategies guided by LIVE's findings, such as developing context-aware decoding or cross-image attention regularization. Additionally, extending this multi-granularity evaluation paradigm to ultra-long visual contexts (e.g., continuous video frames or massive image collections) remains a valuable direction for advancing reliable multimodal architectures.

\bibliography{sample-base}

\appendix
\input{supplemental}

\end{document}

%% file: supplemental.tex
\section{Prompting Templates for LLM-Assisted Counterfactual Generation}
\label{app:prompt_templates}

We provide the prompting templates used in the LLM-assisted counterfactual generation stage. Placeholders such as \{captions\_str\}, \{seed\_object\}, \{class\_type\}, and \{seed\_relationship\} are instantiated with the scene-graph-derived content for each multi-image scenario.

\paragraph{Object-level counterfactual generation}
\begin{tcolorbox}[colback=gray!20, sharp corners, boxrule=0pt, breakable]
You are a creative assistant tasked with generating conflicting information. \\
Background Context (from image captions): \{captions\_str\} \\
Seed Object: \{seed\_object\} \\
Your task is to generate a \emph{new object} that must satisfy the following strict rules:  \\
1. The new object must be conceptually related to the Seed Object, or be its direct antonym / conflicting item (e.g., if the seed is ``horse'', the new object could be ``motorcycle'' or ``zebra'').  \\
2. The new object \textbf{must} conflict with the Background Context. For example, if the context is about a ``horse'' pulling a ``carriage'', generating ``motorcycle'' is a good conflict.  \\
3. The new object \textbf{must not} appear in any of the Background Context descriptions. \\
4. The new object \textbf{must not} be similar to anything in the Background Context descriptions.  \\
5. Provide a brief reason explaining why this new object is conflicting. \\
Output Format (strictly follow):  \\
Object: \texttt{<your new generated object>}  \\
Reason: \texttt{<your reason>} \\
\end{tcolorbox}

\paragraph{Attribute-level counterfactual generation}
\begin{tcolorbox}[colback=gray!20, sharp corners, boxrule=0pt, breakable]
You are a creative assistant tasked with generating conflicting information. \\
Background Context (from image captions): \{captions\_str\} \\
Seed Attribute Class: \{class\_type\}   \\
Seed Attribute: \{seed\_attribute\} \\
Your task is to generate a \emph{new attribute} for the same object, following these strict rules:  \\
1. The new attribute must be from the same class (\{class\_type\}). For example, \{instruction\}.  \\
2. The new attribute \textbf{must} conflict with the Background Context. For example, if the context describes ``black horses'' and the seed is ``horses are black'', a good conflict would be ``horses are white''.  \\
3. The new attribute \textbf{must not} be semantically similar to, or mentioned in, the Background Context.  \\
4. Provide a brief reason explaining why this new attribute is conflicting with the context. \\
Output Format (strictly follow):  \\
Attribute: \texttt{<your new generated attribute>}  \\
Reason: \texttt{<your reason>} \\
\end{tcolorbox}

\paragraph{Relationship-level counterfactual generation}
\begin{tcolorbox}[colback=gray!20, sharp corners, boxrule=0pt, breakable]
You are a creative assistant tasked with generating conflicting information. \\
Background Context (from image captions):  \{captions\_str\} \\
Seed Relationship Class: \{relationship\_class\} \\
Seed Relationship: \{seed\_relationship\}  \\
Your task is to generate a \emph{new relationship} that must satisfy the following strict rules:  \\
1. The new relationship must be a conflicting \{relationship\_class\}.  \\
2. The new relationship \textbf{must} conflict with and be semantically dissimilar to the Background Context.  \\
3. The new relationship \textbf{must not} be semantically similar to, or mentioned in, the Background Context.  \\
4. Provide a brief reason explaining why this new relationship is conflicting with the context. \\
Output Format (strictly follow):  \\
Relationship: \texttt{<your new generated relationship>}  \\
Reason: \texttt{<your reason>} \\
\end{tcolorbox}

\section{Detailed Ablation Study Results}
\label{app:full_ablation}
Table~\ref{tab:appendix_ablation_result} reports ablation results for the remaining six LVLMs, complementing the five models in Table~\ref{tab:ablation_result}. The image-quantity and order settings follow the main text. Arrows indicate score changes relative to the corresponding baseline row.

\begin{table*}[!t]
\centering
\footnotesize
\caption{Additional ablation results for the six models not shown in Table~\ref{tab:ablation_result}.}
\label{tab:appendix_ablation_result}
\setlength\tabcolsep{6pt}
\renewcommand{\arraystretch}{1}
\begin{adjustbox}{max width=\linewidth}
\begin{threeparttable}
\begin{tabular}{lcccccccccccc}
    \toprule
    \scalea{\textbf{Setting}}
    & \multicolumn{2}{c}{\textbf{\scalea{GPT-4o}}}
    & \multicolumn{2}{c}{\textbf{\scalea{Qwen3-VL-32B}}}
    & \multicolumn{2}{c}{\textbf{\scalea{Qwen3-VL-8B}}}
    & \multicolumn{2}{c}{\textbf{\scalea{MiniCPM-V4.5}}}
    & \multicolumn{2}{c}{\textbf{\scalea{MiniCPM-V2.6}}}
    & \multicolumn{2}{c}{\textbf{\scalea{Qwen2.5-VL-7B}}} \\
    \cmidrule(lr){2-3} \cmidrule(lr){4-5} \cmidrule(lr){6-7} \cmidrule(lr){8-9} \cmidrule(lr){10-11} \cmidrule(lr){12-13}
    \scalea{\textbf{Metrics}} &  \scalea{F1} & \scalea{Acc} &  \scalea{F1} & \scalea{Acc} &  \scalea{F1} & \scalea{Acc} &  \scalea{F1} & \scalea{Acc} &  \scalea{F1} & \scalea{Acc} &  \scalea{F1} & \scalea{Acc} \\
    \midrule
    \rowcolor[HTML]{f0f0f0}
    \multicolumn{13}{c}{\textit{\textbf{Scenario 1: Uniform Image Context}}} \\
    \multicolumn{13}{l}{\textit{\textbf{Ablation of Image Quantity Effects}}} \\
    \scalea{\texttt{TI$_1$@4}} & \scaleb{0.945} & \scaleb{0.946} & \scaleb{0.935} & \scaleb{0.939} & \scaleb{0.909} & \scaleb{0.919} & \scaleb{0.944} & \scaleb{0.946} & \scaleb{0.901} & \scaleb{0.907} & \scaleb{0.830} & \scaleb{0.858} \\
    \scalea{\texttt{TI$_1$@3}} & \scaleb{\reduce{0.949\upa}} & \scaleb{\reduce{0.951\upa}} & \scaleb{\reduce{0.938\upa}} & \scaleb{\reduce{0.943\upa}} & \scaleb{\reduce{0.912\upa}} & \scaleb{\reduce{0.921\upa}} & \scaleb{\reduce{0.947\upa}} & \scaleb{\reduce{0.949\upa}} & \scaleb{\reduce{0.912\upa}} & \scaleb{\reduce{0.917\upa}} & \scaleb{\reduce{0.832\upa}} & \scaleb{\reduce{0.858}} \\
    \scalea{\texttt{TI$_1$@2}} & \scaleb{\reduce{0.940\dna}} & \scaleb{\reduce{0.943\dna}} & \scaleb{\reduce{0.938\upa}} & \scaleb{\reduce{0.942\upa}} & \scaleb{\reduce{0.911\upa}} & \scaleb{\reduce{0.921\upa}} & \scaleb{\reduce{0.956\upa}} & \scaleb{\reduce{0.957\upa}} & \scaleb{\reduce{0.912\upa}} & \scaleb{\reduce{0.916\upa}} & \scaleb{\reduce{0.841\upa}} & \scaleb{\reduce{0.866\upa}} \\
    \scalea{\texttt{TI$_2$@4}} & \scaleb{0.839} & \scaleb{0.866} & \scaleb{0.842} & \scaleb{0.869} & \scaleb{0.783} & \scaleb{0.828} & \scaleb{0.811} & \scaleb{0.839} & \scaleb{0.624} & \scaleb{0.734} & \scaleb{0.467} & \scaleb{0.661} \\
    \scalea{\texttt{TI$_2$@3}} & \scaleb{\reduce{0.853\upa}} & \scaleb{\reduce{0.877\upa}} & \scaleb{\reduce{0.849\upa}} & \scaleb{\reduce{0.871\upa}} & \scaleb{\reduce{0.783}} & \scaleb{\reduce{0.827\dna}} & \scaleb{\reduce{0.814\upa}} & \scaleb{\reduce{0.841\upa}} & \scaleb{\reduce{0.625\upa}} & \scaleb{\reduce{0.736\upa}} & \scaleb{\reduce{0.433\dna}} & \scaleb{\reduce{0.645\dna}} \\
    \hdashline
    \multicolumn{13}{l}{\textit{\textbf{Ablation of Image Order Effects}}} \\
    \scalea{\texttt{TI$_3$}} & \scaleb{0.843} & \scaleb{0.869} & \scaleb{0.878} & \scaleb{0.894} & \scaleb{0.823} & \scaleb{0.855} & \scaleb{0.860} & \scaleb{0.875} & \scaleb{0.571} & \scaleb{0.708} & \scaleb{0.431} & \scaleb{0.649} \\
    \scalea{\texttt{TI$_3^\dagger$}} & \scaleb{\reduce{0.842\dna}} & \scaleb{\reduce{0.869}} & \scaleb{\reduce{0.858\dna}} & \scaleb{\reduce{0.880\dna}} & \scaleb{\reduce{0.814\dna}} & \scaleb{\reduce{0.848\dna}} & \scaleb{\reduce{0.848\dna}} & \scaleb{\reduce{0.865\dna}} & \scaleb{\reduce{0.541\dna}} & \scaleb{\reduce{0.692\dna}} & \scaleb{\reduce{0.425\dna}} & \scaleb{\reduce{0.644\dna}} \\
    \scalea{\texttt{TI$_4$}} & \scaleb{0.889} & \scaleb{0.902} & \scaleb{0.927} & \scaleb{0.932} & \scaleb{0.900} & \scaleb{0.910} & \scaleb{0.880} & \scaleb{0.891} & \scaleb{0.731} & \scaleb{0.793} & \scaleb{0.455} & \scaleb{0.659} \\
    \scalea{\texttt{TI$_4^\dagger$}} & \scaleb{\reduce{0.885\dna}} & \scaleb{\reduce{0.899\dna}} & \scaleb{\reduce{0.910\dna}} & \scaleb{\reduce{0.917\dna}} & \scaleb{\reduce{0.908\upa}} & \scaleb{\reduce{0.916\upa}} & \scaleb{\reduce{0.867\dna}} & \scaleb{\reduce{0.881\dna}} & \scaleb{\reduce{0.719\dna}} & \scaleb{\reduce{0.786\dna}} & \scaleb{\reduce{0.439\dna}} & \scaleb{\reduce{0.650\dna}} \\
    \midrule
    \rowcolor[HTML]{f0f0f0}
    \multicolumn{13}{c}{\textit{\textbf{Scenario 2: Diverse Image Context}}} \\
    \multicolumn{13}{l}{\textit{\textbf{Ablation of Image Quantity Effects}}} \\
    \scalea{\texttt{TI$_1$@4}} & \scaleb{0.953} & \scaleb{0.954} & \scaleb{0.924} & \scaleb{0.928} & \scaleb{0.907} & \scaleb{0.917} & \scaleb{0.930} & \scaleb{0.933} & \scaleb{0.908} & \scaleb{0.913} & \scaleb{0.836} & \scaleb{0.861} \\
    \scalea{\texttt{TI$_1$@3}} & \scaleb{\reduce{0.936\dna}} & \scaleb{\reduce{0.938\dna}} & \scaleb{\reduce{0.925\upa}} & \scaleb{\reduce{0.929\upa}} & \scaleb{\reduce{0.911\upa}} & \scaleb{\reduce{0.919\upa}} & \scaleb{\reduce{0.936\upa}} & \scaleb{\reduce{0.939\upa}} & \scaleb{\reduce{0.913\upa}} & \scaleb{\reduce{0.917\upa}} & \scaleb{\reduce{0.837\upa}} & \scaleb{\reduce{0.863\upa}} \\
    \scalea{\texttt{TI$_1$@2}} & \scaleb{\reduce{0.935\dna}} & \scaleb{\reduce{0.938\dna}} & \scaleb{\reduce{0.933\upa}} & \scaleb{\reduce{0.936\upa}} & \scaleb{\reduce{0.898\dna}} & \scaleb{\reduce{0.910\dna}} & \scaleb{\reduce{0.944\upa}} & \scaleb{\reduce{0.946\upa}} & \scaleb{\reduce{0.915\upa}} & \scaleb{\reduce{0.919\upa}} & \scaleb{\reduce{0.855\upa}} & \scaleb{\reduce{0.875\upa}} \\
    \scalea{\texttt{TI$_2$@4}} & \scaleb{0.792} & \scaleb{0.838} & \scaleb{0.790} & \scaleb{0.835} & \scaleb{0.643} & \scaleb{0.747} & \scaleb{0.698} & \scaleb{0.768} & \scaleb{0.427} & \scaleb{0.646} & \scaleb{0.242} & \scaleb{0.572} \\
    \scalea{\texttt{TI$_2$@3}} & \scaleb{\reduce{0.809\upa}} & \scaleb{\reduce{0.849\upa}} & \scaleb{\reduce{0.802\upa}} & \scaleb{\reduce{0.841\upa}} & \scaleb{\reduce{0.642\dna}} & \scaleb{\reduce{0.746\dna}} & \scaleb{\reduce{0.689\dna}} & \scaleb{\reduce{0.766\dna}} & \scaleb{\reduce{0.397\dna}} & \scaleb{\reduce{0.633\dna}} & \scaleb{\reduce{0.203\dna}} & \scaleb{\reduce{0.556\dna}} \\
    \hdashline
    \multicolumn{13}{l}{\textit{\textbf{Ablation of Image Order Effects}}} \\
    \scalea{\texttt{TI$_3$}} & \scaleb{0.796} & \scaleb{0.841} & \scaleb{0.833} & \scaleb{0.863} & \scaleb{0.712} & \scaleb{0.782} & \scaleb{0.801} & \scaleb{0.832} & \scaleb{0.324} & \scaleb{0.601} & \scaleb{0.189} & \scaleb{0.556} \\
    \scalea{\texttt{TI$_3^\dagger$}} & \scaleb{\reduce{0.795\dna}} & \scaleb{\reduce{0.843\upa}} & \scaleb{\reduce{0.837\upa}} & \scaleb{\reduce{0.863}} & \scaleb{\reduce{0.721\upa}} & \scaleb{\reduce{0.788\upa}} & \scaleb{\reduce{0.800\dna}} & \scaleb{\reduce{0.832}} & \scaleb{\reduce{0.324}} & \scaleb{\reduce{0.601}} & \scaleb{\reduce{0.161\dna}} & \scaleb{\reduce{0.547\dna}} \\
    \scalea{\texttt{TI$_4$}} & \scaleb{0.847} & \scaleb{0.874} & \scaleb{0.894} & \scaleb{0.903} & \scaleb{0.829} & \scaleb{0.853} & \scaleb{0.799} & \scaleb{0.826} & \scaleb{0.561} & \scaleb{0.704} & \scaleb{0.216} & \scaleb{0.566} \\
    \scalea{\texttt{TI$_4^\dagger$}} & \scaleb{\reduce{0.857\upa}} & \scaleb{\reduce{0.879\upa}} & \scaleb{\reduce{0.878\dna}} & \scaleb{\reduce{0.890\dna}} & \scaleb{\reduce{0.831\upa}} & \scaleb{\reduce{0.856\upa}} & \scaleb{\reduce{0.810\upa}} & \scaleb{\reduce{0.837\upa}} & \scaleb{\reduce{0.556\dna}} & \scaleb{\reduce{0.702\dna}} & \scaleb{\reduce{0.212\dna}} & \scaleb{\reduce{0.563\dna}} \\
    \bottomrule
\end{tabular}
\begin{tablenotes}
    \scriptsize
    \item  \textit{Note}: $^\dagger$ denotes random shuffling of targeted images. $\uparrow/\downarrow$ denote an increase/decrease versus corresponding baseline row.
\end{tablenotes}
\end{threeparttable}
\end{adjustbox}
\end{table*}

\section{Detailed Multi-Granularity Evaluation Results under Different Temperatures}
\label{sec:appendix_results}

In this section, we provide the comprehensive performance breakdowns of all evaluated LVLMs on the LIVE benchmark, supplementing the aggregated MAP results presented in Section~\ref{subsec:overall_evaluation}.
We evaluate three sampling temperatures: 0 (greedy decoding, used in the main experiments), 0.7, and 1.0. While varying the temperature introduces expected fluctuations in absolute metric scores---reflecting the inherent trade-off between response diversity and factual grounding---the core observations remain highly consistent. Detailed results are provided in Tables~\ref{tab:supp_main_result_t0}, \ref{tab:supp_main_result_t07}, and \ref{tab:supp_main_result_t10}, respectively.

\begin{table*}[tp]
\centering
\footnotesize
\setlength{\tabcolsep}{3pt}
\renewcommand{\arraystretch}{1}
\caption{Performance by target-image granularity at $T=0$. The corresponding MAP scores are reported in Table~\ref{tab:main_result}.}
\label{tab:supp_main_result_t0}
\begin{adjustbox}{max width=\linewidth}
\begin{tabular}{lc cccccccccccccccccccccc}
    \toprule
    \multicolumn{2}{c}{\textbf{\scalea{Task}}}
    & \multicolumn{2}{c}{\textbf{\scalea{GPT-4o}}}
    & \multicolumn{2}{c}{\textbf{\scalea{Qwen3-VL-32B}}}
    & \multicolumn{2}{c}{\textbf{\scalea{InternVL-2.5-26B}}}
    & \multicolumn{2}{c}{\textbf{\scalea{Qwen3-VL-8B}}}
    & \multicolumn{2}{c}{\textbf{\scalea{InternVL-2.5-8B}}}
    & \multicolumn{2}{c}{\textbf{\scalea{MiniCPM-V4.5}}}
    & \multicolumn{2}{c}{\textbf{\scalea{MiniCPM-V2.6}}}
    & \multicolumn{2}{c}{\textbf{\scalea{Mantis-8B}}}
    & \multicolumn{2}{c}{\textbf{\scalea{Qwen2.5-VL-7B}}}
    & \multicolumn{2}{c}{\textbf{\scalea{LLaVA-OV}}}
    & \multicolumn{2}{c}{\textbf{\scalea{LLaVA-IL}}} \\

    \cmidrule(lr){3-4} \cmidrule(lr){5-6} \cmidrule(lr){7-8} \cmidrule(lr){9-10} \cmidrule(lr){11-12} \cmidrule(lr){13-14} \cmidrule(lr){15-16} \cmidrule(lr){17-18} \cmidrule(lr){19-20} \cmidrule(lr){21-22} \cmidrule(lr){23-24}

    \scalea{\textbf{Class}} & \scalea{\textbf{TI}} & \scalea{F1} & \scalea{Acc} & \scalea{F1} & \scalea{Acc} & \scalea{F1} & \scalea{Acc} & \scalea{F1} & \scalea{Acc} & \scalea{F1} & \scalea{Acc} & \scalea{F1} & \scalea{Acc} & \scalea{F1} & \scalea{Acc} & \scalea{F1} & \scalea{Acc} & \scalea{F1} & \scalea{Acc} & \scalea{F1} & \scalea{Acc} & \scalea{F1} & \scalea{Acc}  \\
    \midrule
    \rowcolor[HTML]{f0f0f0}
    \multicolumn{24}{c}{\textit{\textbf{Scenario 1: Uniform Image Context}}} \\
    \midrule
    \multirow{4}{*}{\scalea{Object}} & \scalea{\texttt{TI$_1$}} & \scaleb{0.969} & \scaleb{0.969} & \scaleb{\subbst{0.963}} & \scaleb{\subbst{0.964}} & \scaleb{0.896} & \scaleb{0.904} & \scaleb{\bst{0.969}} & \scaleb{\bst{0.969}} & \scaleb{0.815} & \scaleb{0.845} & \scaleb{0.961} & \scaleb{0.962} & \scaleb{0.959} & \scaleb{0.960} & \scaleb{0.894} & \scaleb{0.890} & \scaleb{0.892} & \scaleb{0.903} & \scaleb{0.925} & \scaleb{0.926} & \scaleb{0.901} & \scaleb{0.896} \\
\bottomrule
\end{tabular}
\end{adjustbox}

\begin{tablenotes}
    \item  \scriptsize \textit{Note}:  \bst{Bold} and \subbst{underlined} denote best and second-best results (excluding GPT-4o).
\end{tablenotes}
\end{table*}

\begin{table*}[tp]
\centering
\footnotesize
\setlength{\tabcolsep}{3pt}
\renewcommand{\arraystretch}{1}
\caption{Performance by target-image granularity at $T=0.7$. Avg denotes MAP over \texttt{TI$_1$}--\texttt{TI$_4$}.}
\label{tab:supp_main_result_t07}
\begin{adjustbox}{max width=\linewidth}
\begin{tabular}{lc cccccccccccccccccccccc}
    \toprule
    \multicolumn{2}{c}{\textbf{\scalea{Task}}}
    & \multicolumn{2}{c}{\textbf{\scalea{GPT-4o}}}
    & \multicolumn{2}{c}{\textbf{\scalea{Qwen3-VL-32B}}}
    & \multicolumn{2}{c}{\textbf{\scalea{InternVL-2.5-26B}}}
    & \multicolumn{2}{c}{\textbf{\scalea{Qwen3-VL-8B}}}
    & \multicolumn{2}{c}{\textbf{\scalea{InternVL-2.5-8B}}}
    & \multicolumn{2}{c}{\textbf{\scalea{MiniCPM-V4.5}}}
    & \multicolumn{2}{c}{\textbf{\scalea{MiniCPM-V2.6}}}
    & \multicolumn{2}{c}{\textbf{\scalea{Mantis-8B}}}
    & \multicolumn{2}{c}{\textbf{\scalea{Qwen2.5-VL-7B}}}
    & \multicolumn{2}{c}{\textbf{\scalea{LLaVA-OV}}}
    & \multicolumn{2}{c}{\textbf{\scalea{LLaVA-IL}}} \\

    \cmidrule(lr){3-4} \cmidrule(lr){5-6} \cmidrule(lr){7-8} \cmidrule(lr){9-10} \cmidrule(lr){11-12} \cmidrule(lr){13-14} \cmidrule(lr){15-16} \cmidrule(lr){17-18} \cmidrule(lr){19-20} \cmidrule(lr){21-22} \cmidrule(lr){23-24}

    \scalea{\textbf{Class}} & \scalea{\textbf{TI}} & \scalea{F1} & \scalea{Acc} & \scalea{F1} & \scalea{Acc} & \scalea{F1} & \scalea{Acc} & \scalea{F1} & \scalea{Acc} & \scalea{F1} & \scalea{Acc} & \scalea{F1} & \scalea{Acc} & \scalea{F1} & \scalea{Acc} & \scalea{F1} & \scalea{Acc} & \scalea{F1} & \scalea{Acc} & \scalea{F1} & \scalea{Acc} & \scalea{F1} & \scalea{Acc}  \\
    \midrule
    \rowcolor[HTML]{f0f0f0}
    \multicolumn{24}{c}{\textit{\textbf{Scenario 1: Uniform Image Context}}} \\
    \midrule
    \multirow{5}{*}{\scalea{Object}} & \scalea{\texttt{TI$_1$}} & \scaleb{0.968} & \scaleb{0.968} & \scaleb{\subbst{0.964}} & \scaleb{\subbst{0.964}} & \scaleb{0.866} & \scaleb{0.877} & \scaleb{\bst{0.969}} & \scaleb{\bst{0.969}} & \scaleb{0.786} & \scaleb{0.819} & \scaleb{0.959} & \scaleb{0.960} & \scaleb{0.944} & \scaleb{0.945} & \scaleb{0.859} & \scaleb{0.856} & \scaleb{0.864} & \scaleb{0.879} & \scaleb{0.822} & \scaleb{0.834} & \scaleb{0.861} & \scaleb{0.853} \\
\bottomrule
\end{tabular}
\end{adjustbox}

\begin{tablenotes}
    \item  \scriptsize \textit{Note}:  \bst{Bold} and \subbst{underlined} denote best and second-best results (excluding GPT-4o).
\end{tablenotes}
\end{table*}

\begin{table*}[tp]
\centering
\footnotesize
\setlength{\tabcolsep}{3pt}
\renewcommand{\arraystretch}{1}
\caption{Performance by target-image granularity at $T=1.0$. Avg denotes MAP over \texttt{TI$_1$}--\texttt{TI$_4$}.}
\label{tab:supp_main_result_t10}
\begin{adjustbox}{max width=\linewidth}
\begin{tabular}{lc cccccccccccccccccccccc}
    \toprule
    \multicolumn{2}{c}{\textbf{\scalea{Task}}}
    & \multicolumn{2}{c}{\textbf{\scalea{GPT-4o}}}
    & \multicolumn{2}{c}{\textbf{\scalea{Qwen3-VL-32B}}}
    & \multicolumn{2}{c}{\textbf{\scalea{InternVL-2.5-26B}}}
    & \multicolumn{2}{c}{\textbf{\scalea{Qwen3-VL-8B}}}
    & \multicolumn{2}{c}{\textbf{\scalea{InternVL-2.5-8B}}}
    & \multicolumn{2}{c}{\textbf{\scalea{MiniCPM-V4.5}}}
    & \multicolumn{2}{c}{\textbf{\scalea{MiniCPM-V2.6}}}
    & \multicolumn{2}{c}{\textbf{\scalea{Mantis-8B}}}
    & \multicolumn{2}{c}{\textbf{\scalea{Qwen2.5-VL-7B}}}
    & \multicolumn{2}{c}{\textbf{\scalea{LLaVA-OV}}}
    & \multicolumn{2}{c}{\textbf{\scalea{LLaVA-IL}}} \\

    \cmidrule(lr){3-4} \cmidrule(lr){5-6} \cmidrule(lr){7-8} \cmidrule(lr){9-10} \cmidrule(lr){11-12} \cmidrule(lr){13-14} \cmidrule(lr){15-16} \cmidrule(lr){17-18} \cmidrule(lr){19-20} \cmidrule(lr){21-22} \cmidrule(lr){23-24}

    \scalea{\textbf{Class}} & \scalea{\textbf{TI}} & \scalea{F1} & \scalea{Acc} & \scalea{F1} & \scalea{Acc} & \scalea{F1} & \scalea{Acc} & \scalea{F1} & \scalea{Acc} & \scalea{F1} & \scalea{Acc} & \scalea{F1} & \scalea{Acc} & \scalea{F1} & \scalea{Acc} & \scalea{F1} & \scalea{Acc} & \scalea{F1} & \scalea{Acc} & \scalea{F1} & \scalea{Acc} & \scalea{F1} & \scalea{Acc}  \\
    \midrule
    \rowcolor[HTML]{f0f0f0}
    \multicolumn{24}{c}{\textit{\textbf{Scenario 1: Uniform Image Context}}} \\
    \midrule
    \multirow{5}{*}{\scalea{Object}} & \scalea{\texttt{TI$_1$}} & \scaleb{0.969} & \scaleb{0.969} & \scaleb{\subbst{0.962}} & \scaleb{\subbst{0.963}} & \scaleb{0.853} & \scaleb{0.864} & \scaleb{\bst{0.970}} & \scaleb{\bst{0.970}} & \scaleb{0.763} & \scaleb{0.802} & \scaleb{0.945} & \scaleb{0.946} & \scaleb{0.931} & \scaleb{0.932} & \scaleb{0.797} & \scaleb{0.796} & \scaleb{0.845} & \scaleb{0.863} & \scaleb{0.781} & \scaleb{0.794} & \scaleb{0.831} & \scaleb{0.824} \\
\bottomrule
\end{tabular}
\end{adjustbox}

\begin{tablenotes}
    \item  \scriptsize \textit{Note}:  \bst{Bold} and \subbst{underlined} denote best and second-best results (excluding GPT-4o).
\end{tablenotes}
\end{table*}

\section{Additional Multi-Granularity Evaluation Results}
\label{sec:full_multi_granularity}
\begin{figure*}[tp]
\centering
\begin{subfigure}[b]{0.24\linewidth}
    \centering
    \includegraphics[width=\linewidth]{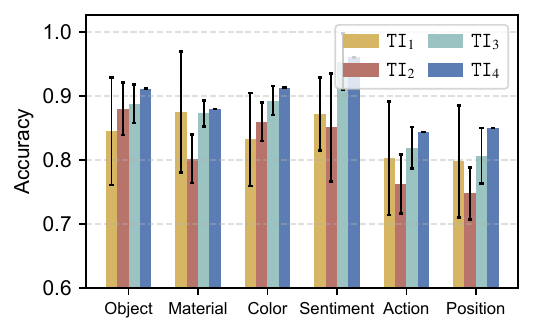}
    \caption{InternVL-2.5-8B (UIC)}
\end{subfigure}
\hfill
\begin{subfigure}[b]{0.24\linewidth}
    \centering
    \includegraphics[width=\linewidth]{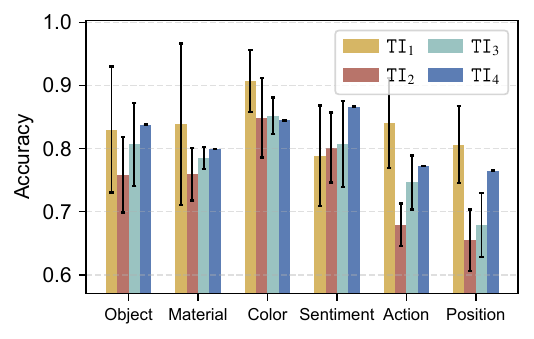}
    \caption{InternVL-2.5-8B (DIC)}
\end{subfigure}
\hfill
\begin{subfigure}[b]{0.24\linewidth}
    \centering
    \includegraphics[width=\linewidth]{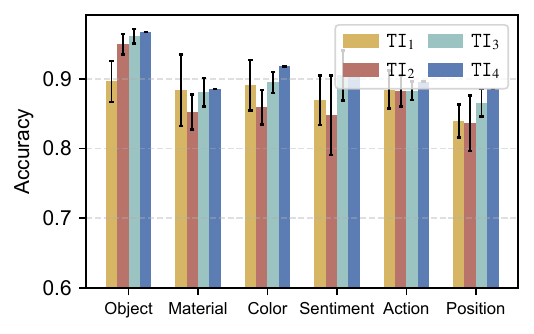}
    \caption{LLaVA-IL (UIC)}
\end{subfigure}
\hfill
\begin{subfigure}[b]{0.24\linewidth}
    \centering
    \includegraphics[width=\linewidth]{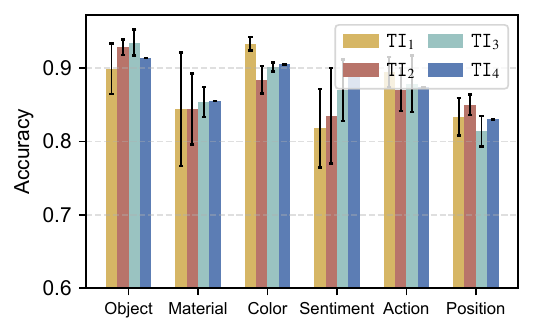}
    \caption{LLaVA-IL (DIC)}
\end{subfigure}

\vspace{0.25cm}

\begin{subfigure}[b]{0.24\linewidth}
    \centering
    \includegraphics[width=\linewidth]{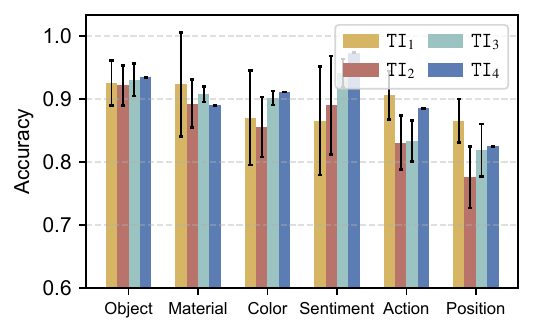}
    \caption{LLaVA-OV (UIC)}
\end{subfigure}
\hfill
\begin{subfigure}[b]{0.24\linewidth}
    \centering
    \includegraphics[width=\linewidth]{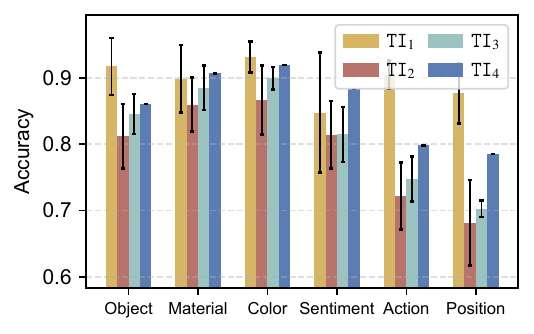}
    \caption{LLaVA-OV (DIC)}
\end{subfigure}
\hfill
\begin{subfigure}[b]{0.24\linewidth}
    \centering
    \includegraphics[width=\linewidth]{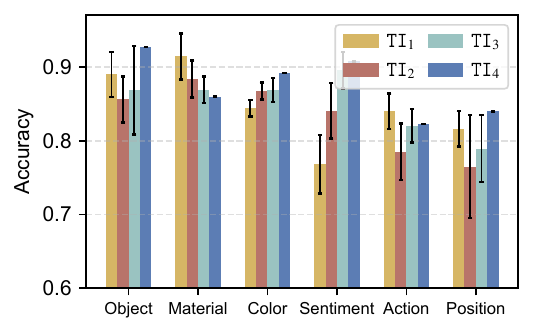}
    \caption{Mantis-8B (UIC)}
\end{subfigure}
\hfill
\begin{subfigure}[b]{0.24\linewidth}
    \centering
    \includegraphics[width=\linewidth]{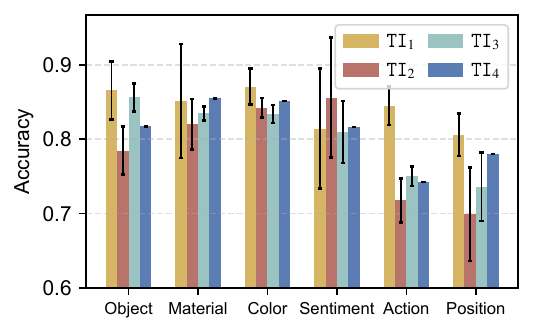}
    \caption{Mantis-8B (DIC)}
\end{subfigure}

\vspace{0.25cm}

\begin{subfigure}[b]{0.24\linewidth}
    \centering
    \includegraphics[width=\linewidth]{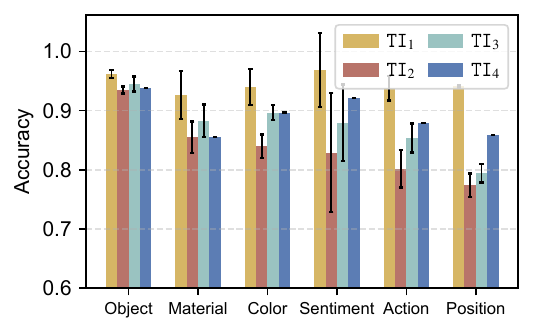}
    \caption{MiniCPM-V4.5 (UIC)}
\end{subfigure}
\hfill
\begin{subfigure}[b]{0.24\linewidth}
    \centering
    \includegraphics[width=\linewidth]{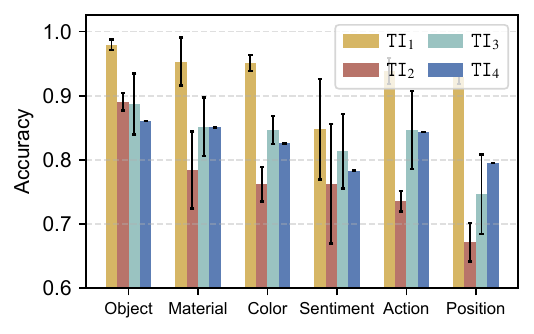}
    \caption{MiniCPM-V4.5 (DIC)}
\end{subfigure}
\hfill
\begin{subfigure}[b]{0.24\linewidth}
    \centering
    \includegraphics[width=\linewidth]{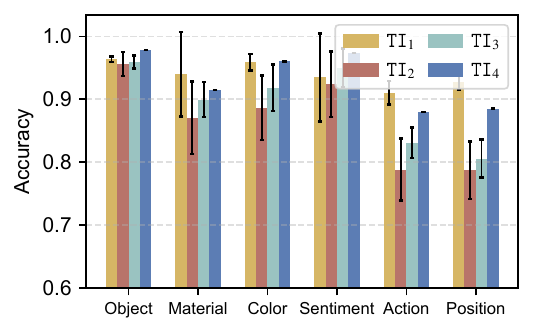}
    \caption{Qwen3-VL-32B (UIC)}
\end{subfigure}
\hfill
\begin{subfigure}[b]{0.24\linewidth}
    \centering
    \includegraphics[width=\linewidth]{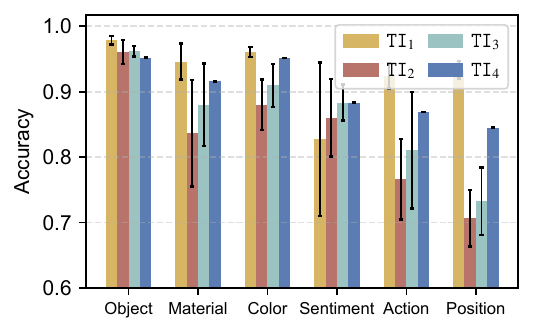}
    \caption{Qwen3-VL-32B (DIC)}
\end{subfigure}

\vspace{0.25cm}

\begin{subfigure}[b]{0.24\linewidth}
    \centering
    \includegraphics[width=\linewidth]{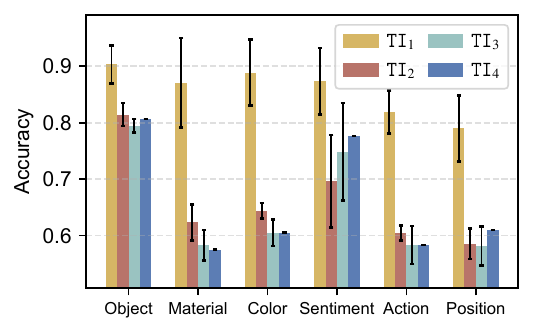}
    \caption{Qwen2.5-VL-7B (UIC)}
\end{subfigure}
\hfill
\begin{subfigure}[b]{0.24\linewidth}
    \centering
    \includegraphics[width=\linewidth]{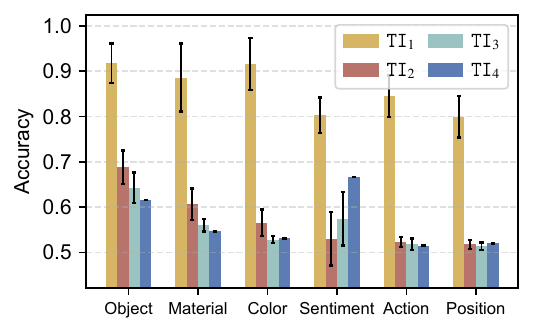}
    \caption{Qwen2.5-VL-7B (DIC)}
\end{subfigure}
\hfill
\begin{subfigure}[b]{0.24\linewidth}
    \centering
    \includegraphics[width=\linewidth]{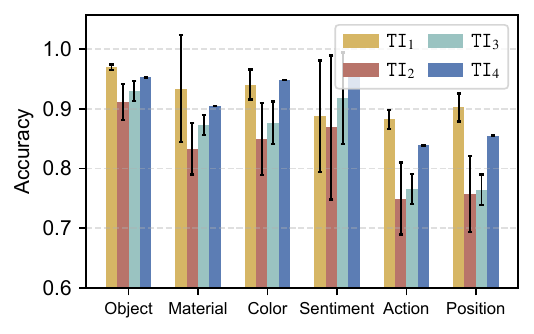}
    \caption{Qwen3-VL-8B (UIC)}
\end{subfigure}
\hfill
\begin{subfigure}[b]{0.24\linewidth}
    \centering
    \includegraphics[width=\linewidth]{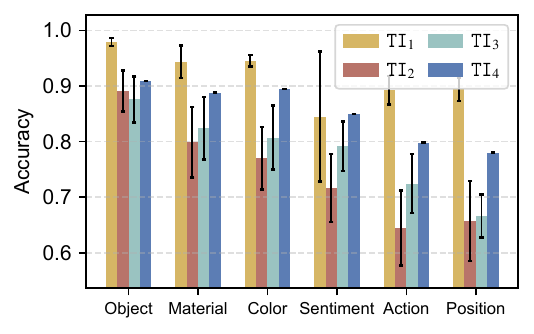}
    \caption{Qwen3-VL-8B (DIC)}
\end{subfigure}

\vspace{0.25cm}

\begin{subfigure}[b]{0.24\linewidth}
    \centering
    \includegraphics[width=\linewidth]{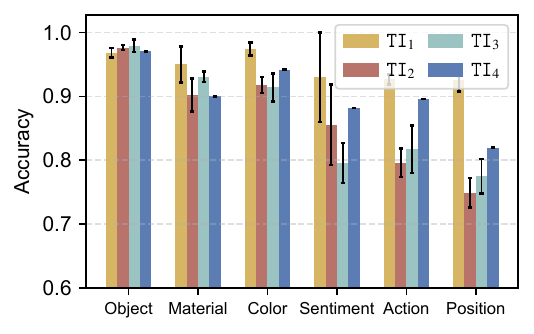}
    \caption{GPT-4o (UIC)}
\end{subfigure}
\hfill
\begin{subfigure}[b]{0.24\linewidth}
    \centering
    \includegraphics[width=\linewidth]{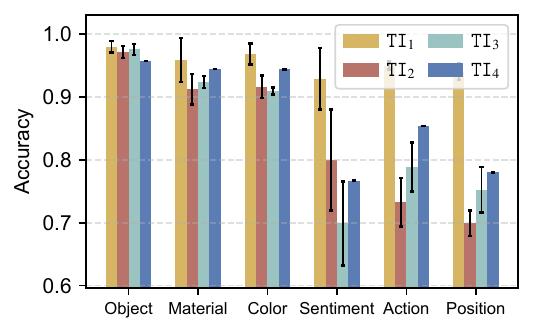}
    \caption{GPT-4o (DIC)}
\end{subfigure}
\hfill
\begin{subfigure}[b]{0.48\linewidth}
    \mbox{}
\end{subfigure}

\caption{Full multi-granularity evaluation results on LIVE for the remaining nine LVLMs at a temperature of 0. For each model, we report accuracy under Uniform Image Context (UIC, left) and Diverse Image Context (DIC, right).}
\label{fig:full_multi_granularity_curves_t0}
\end{figure*}

\begin{figure*}[tp]
\centering
\begin{subfigure}[b]{0.24\linewidth}
    \centering
    \includegraphics[width=\linewidth]{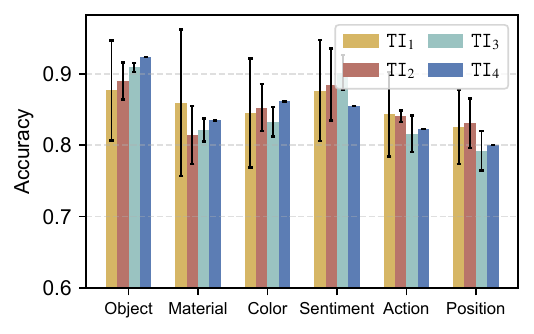}
    \caption{InternVL-2.5-26B (UIC)}
\end{subfigure}
\hfill
\begin{subfigure}[b]{0.24\linewidth}
    \centering
    \includegraphics[width=\linewidth]{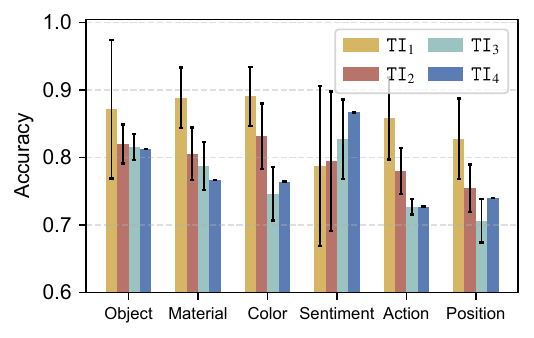}
    \caption{InternVL-2.5-26B (DIC)}
\end{subfigure}
\hfill
\begin{subfigure}[b]{0.24\linewidth}
    \centering
    \includegraphics[width=\linewidth]{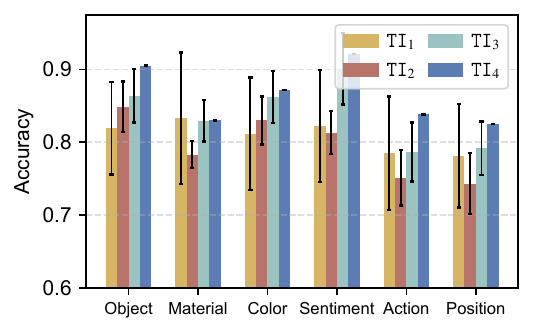}
    \caption{InternVL-2.5-8B (UIC)}
\end{subfigure}
\hfill
\begin{subfigure}[b]{0.24\linewidth}
    \centering
    \includegraphics[width=\linewidth]{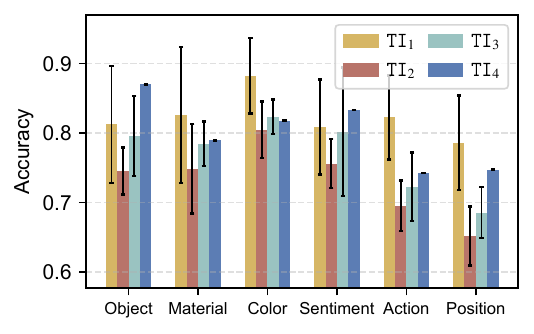}
    \caption{InternVL-2.5-8B (DIC)}
\end{subfigure}

\vspace{0.25cm}

\begin{subfigure}[b]{0.24\linewidth}
    \centering
    \includegraphics[width=\linewidth]{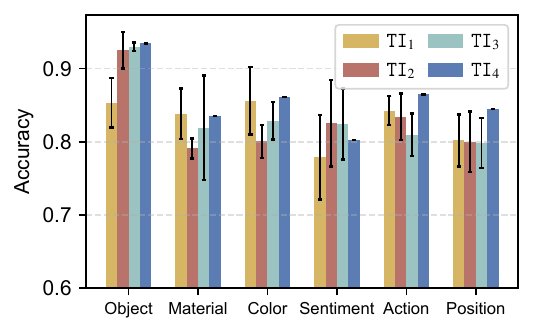}
    \caption{LLaVA-IL (UIC)}
\end{subfigure}
\hfill
\begin{subfigure}[b]{0.24\linewidth}
    \centering
    \includegraphics[width=\linewidth]{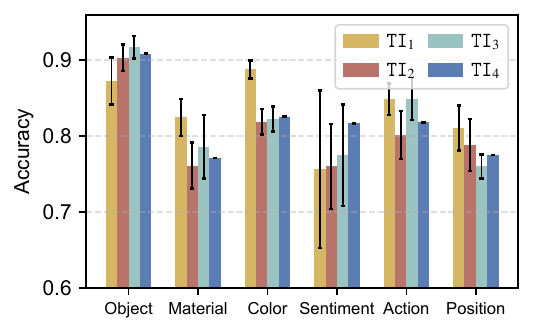}
    \caption{LLaVA-IL (DIC)}
\end{subfigure}
\hfill
\begin{subfigure}[b]{0.24\linewidth}
    \centering
    \includegraphics[width=\linewidth]{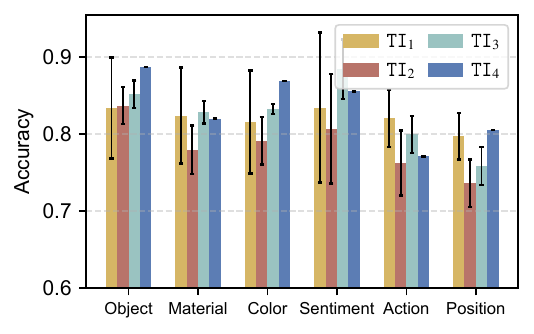}
    \caption{LLaVA-OV (UIC)}
\end{subfigure}
\hfill
\begin{subfigure}[b]{0.24\linewidth}
    \centering
    \includegraphics[width=\linewidth]{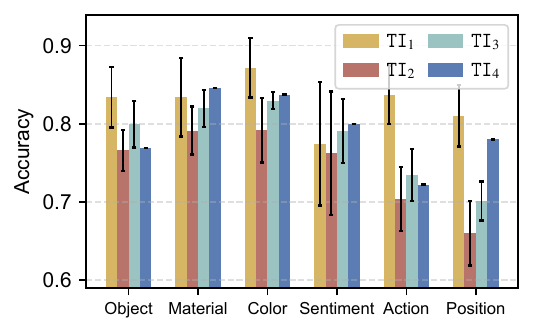}
    \caption{LLaVA-OV (DIC)}
\end{subfigure}

\vspace{0.25cm}

\begin{subfigure}[b]{0.24\linewidth}
    \centering
    \includegraphics[width=\linewidth]{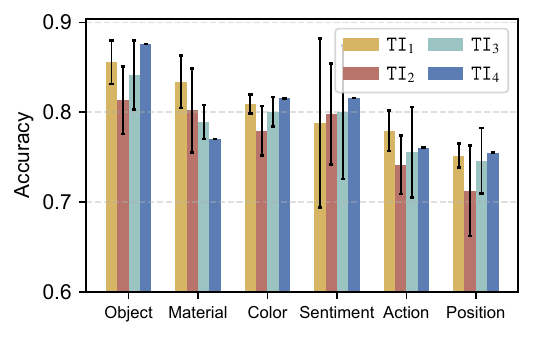}
    \caption{Mantis-8B (UIC)}
\end{subfigure}
\hfill
\begin{subfigure}[b]{0.24\linewidth}
    \centering
    \includegraphics[width=\linewidth]{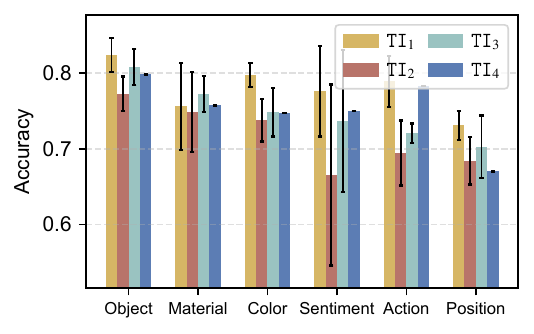}
    \caption{Mantis-8B (DIC)}
\end{subfigure}
\hfill
\begin{subfigure}[b]{0.24\linewidth}
    \centering
    \includegraphics[width=\linewidth]{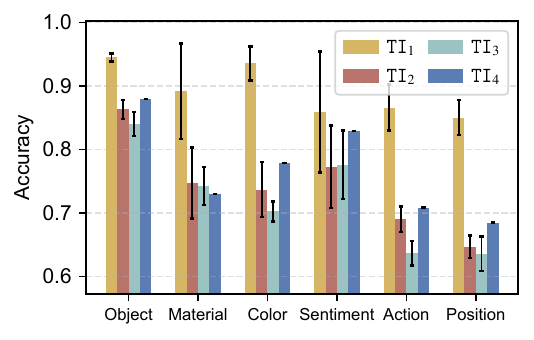}
    \caption{MiniCPM-V2.6 (UIC)}
\end{subfigure}
\hfill
\begin{subfigure}[b]{0.24\linewidth}
    \centering
    \includegraphics[width=\linewidth]{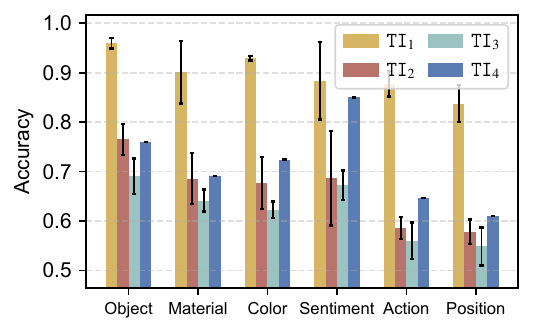}
    \caption{MiniCPM-V2.6 (DIC)}
\end{subfigure}

\vspace{0.25cm}

\begin{subfigure}[b]{0.24\linewidth}
    \centering
    \includegraphics[width=\linewidth]{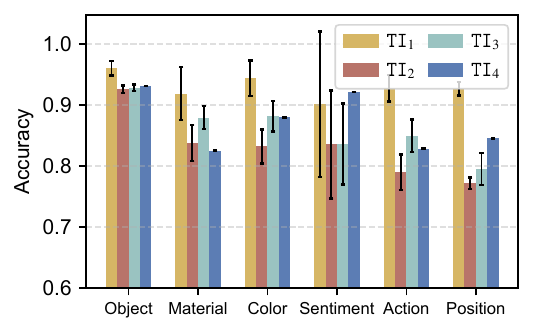}
    \caption{MiniCPM-V4.5 (UIC)}
\end{subfigure}
\hfill
\begin{subfigure}[b]{0.24\linewidth}
    \centering
    \includegraphics[width=\linewidth]{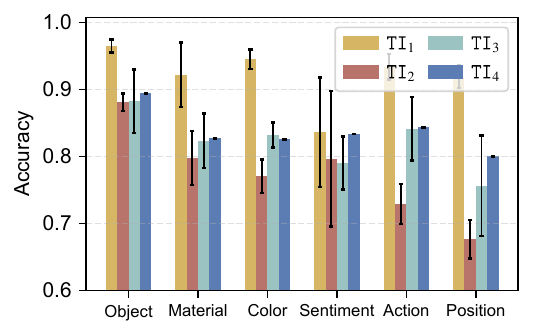}
    \caption{MiniCPM-V4.5 (DIC)}
\end{subfigure}
\hfill
\begin{subfigure}[b]{0.24\linewidth}
    \centering
    \includegraphics[width=\linewidth]{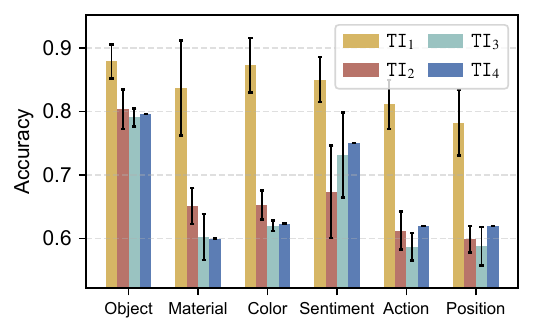}
    \caption{Qwen2.5-VL-7B (UIC)}
\end{subfigure}
\hfill
\begin{subfigure}[b]{0.24\linewidth}
    \centering
    \includegraphics[width=\linewidth]{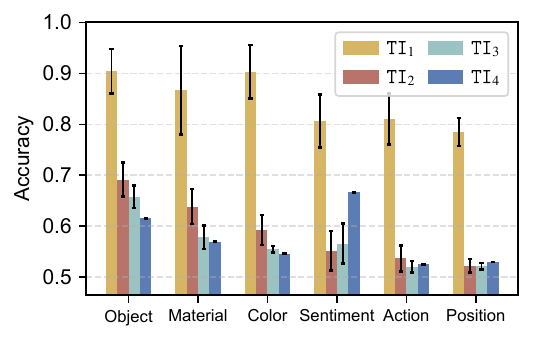}
    \caption{Qwen2.5-VL-7B (DIC)}
\end{subfigure}

\vspace{0.25cm}

\begin{subfigure}[b]{0.24\linewidth}
    \centering
    \includegraphics[width=\linewidth]{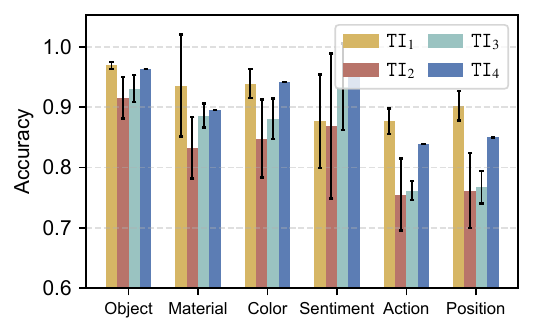}
    \caption{Qwen3-VL-8B (UIC)}
\end{subfigure}
\hfill
\begin{subfigure}[b]{0.24\linewidth}
    \centering
    \includegraphics[width=\linewidth]{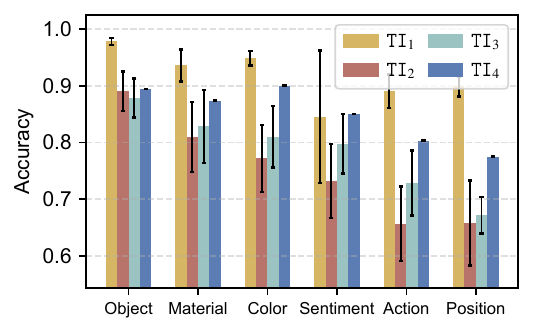}
    \caption{Qwen3-VL-8B (DIC)}
\end{subfigure}
\hfill
\begin{subfigure}[b]{0.24\linewidth}
    \centering
    \includegraphics[width=\linewidth]{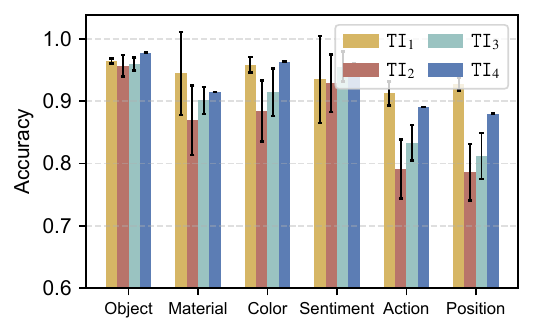}
    \caption{Qwen3-VL-32B (UIC)}
\end{subfigure}
\hfill
\begin{subfigure}[b]{0.24\linewidth}
    \centering
    \includegraphics[width=\linewidth]{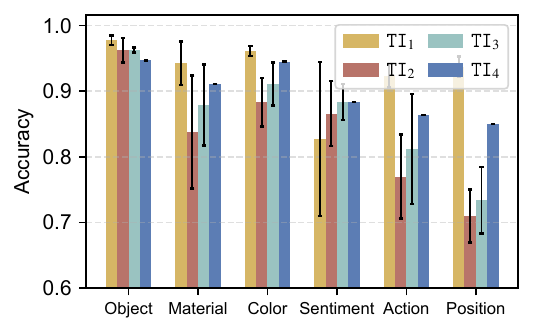}
    \caption{Qwen3-VL-32B (DIC)}
\end{subfigure}

\vspace{0.25cm}

\begin{subfigure}[b]{0.24\linewidth}
    \centering
    \includegraphics[width=\linewidth]{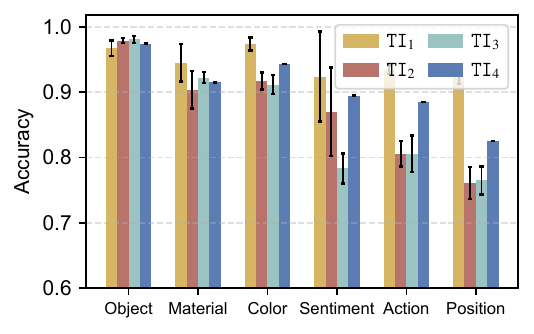}
    \caption{GPT-4o (UIC)}
\end{subfigure}
\hfill
\begin{subfigure}[b]{0.24\linewidth}
    \centering
    \includegraphics[width=\linewidth]{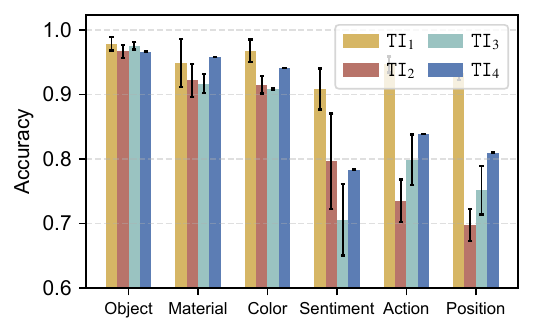}
    \caption{GPT-4o (DIC)}
\end{subfigure}
\hfill
\begin{subfigure}[b]{0.48\linewidth}
    \mbox{}
\end{subfigure}

\caption{Full multi-granularity evaluation results on LIVE for all 11 LVLMs at a temperature of 0.7. For each model, we report accuracy under Uniform Image Context (UIC, left) and Diverse Image Context (DIC, right).}
\label{fig:full_multi_granularity_curves_t07}
\end{figure*}

\begin{figure*}[tp]
\centering
\begin{subfigure}[b]{0.24\linewidth}
    \centering
    \includegraphics[width=\linewidth]{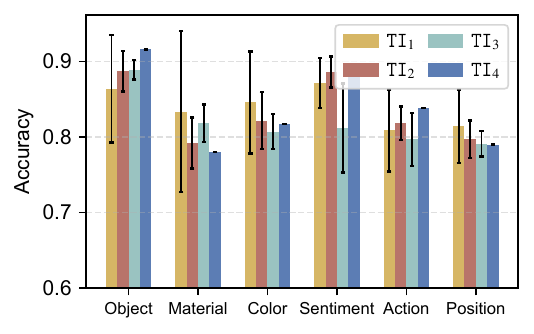}
    \caption{InternVL-2.5-26B (UIC)}
\end{subfigure}
\hfill
\begin{subfigure}[b]{0.24\linewidth}
    \centering
    \includegraphics[width=\linewidth]{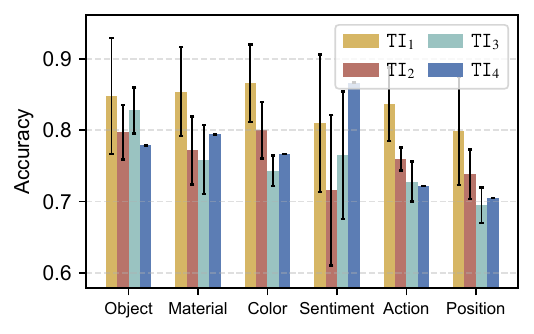}
    \caption{InternVL-2.5-26B (DIC)}
\end{subfigure}
\hfill
\begin{subfigure}[b]{0.24\linewidth}
    \centering
    \includegraphics[width=\linewidth]{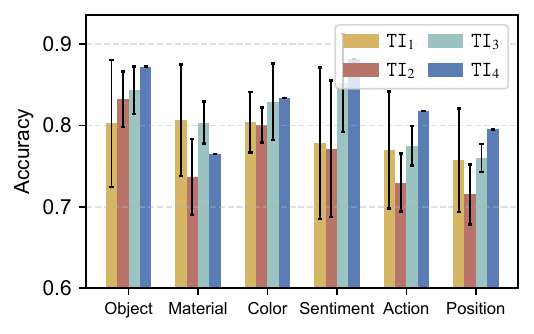}
    \caption{InternVL-2.5-8B (UIC)}
\end{subfigure}
\hfill
\begin{subfigure}[b]{0.24\linewidth}
    \centering
    \includegraphics[width=\linewidth]{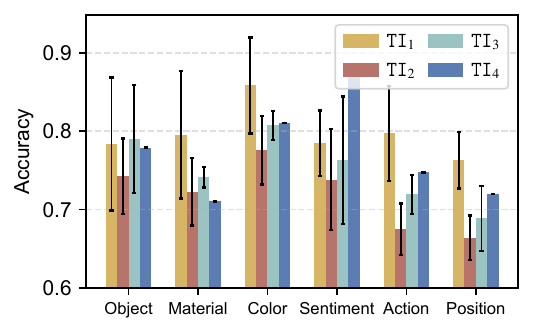}
    \caption{InternVL-2.5-8B (DIC)}
\end{subfigure}

\vspace{0.25cm}

\begin{subfigure}[b]{0.24\linewidth}
    \centering
    \includegraphics[width=\linewidth]{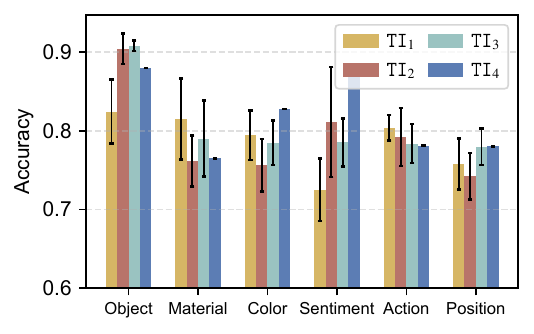}
    \caption{LLaVA-IL (UIC)}
\end{subfigure}
\hfill
\begin{subfigure}[b]{0.24\linewidth}
    \centering
    \includegraphics[width=\linewidth]{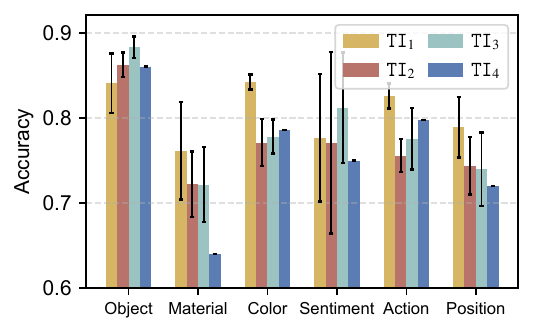}
    \caption{LLaVA-IL (DIC)}
\end{subfigure}
\hfill
\begin{subfigure}[b]{0.24\linewidth}
    \centering
    \includegraphics[width=\linewidth]{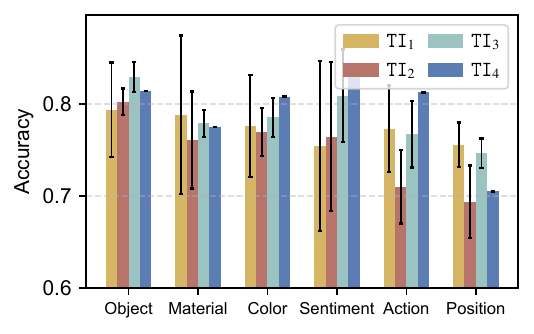}
    \caption{LLaVA-OV (UIC)}
\end{subfigure}
\hfill
\begin{subfigure}[b]{0.24\linewidth}
    \centering
    \includegraphics[width=\linewidth]{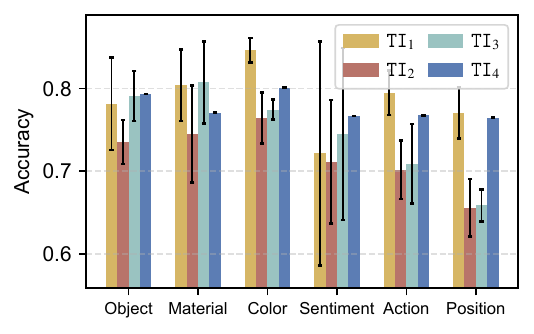}
    \caption{LLaVA-OV (DIC)}
\end{subfigure}

\vspace{0.25cm}

\begin{subfigure}[b]{0.24\linewidth}
    \centering
    \includegraphics[width=\linewidth]{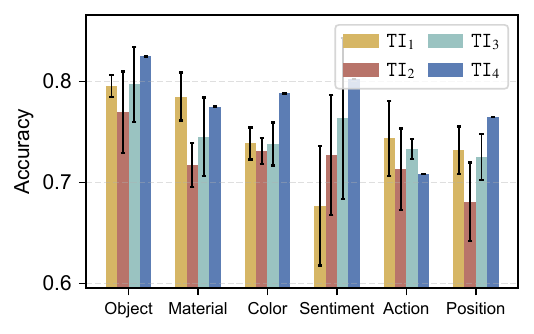}
    \caption{Mantis-8B (UIC)}
\end{subfigure}
\hfill
\begin{subfigure}[b]{0.24\linewidth}
    \centering
    \includegraphics[width=\linewidth]{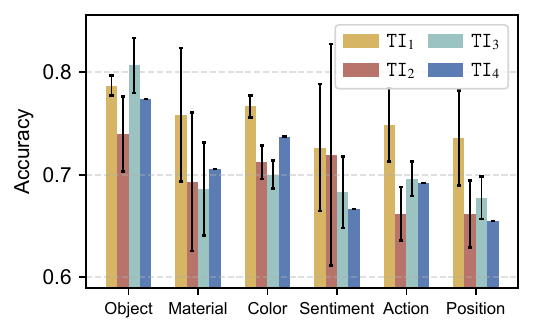}
    \caption{Mantis-8B (DIC)}
\end{subfigure}
\hfill
\begin{subfigure}[b]{0.24\linewidth}
    \centering
    \includegraphics[width=\linewidth]{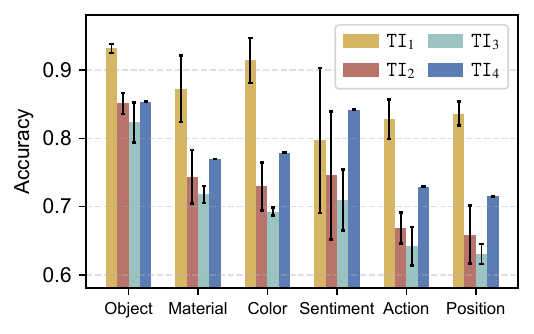}
    \caption{MiniCPM-V2.6 (UIC)}
\end{subfigure}
\hfill
\begin{subfigure}[b]{0.24\linewidth}
    \centering
    \includegraphics[width=\linewidth]{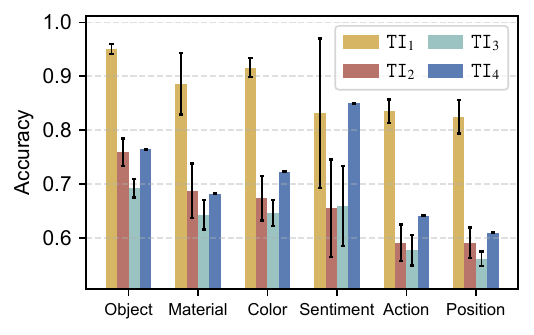}
    \caption{MiniCPM-V2.6 (DIC)}
\end{subfigure}

\vspace{0.25cm}

\begin{subfigure}[b]{0.24\linewidth}
    \centering
    \includegraphics[width=\linewidth]{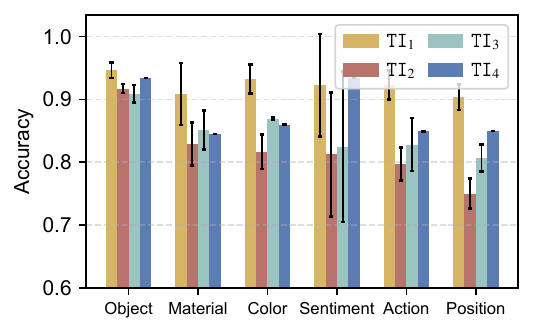}
    \caption{MiniCPM-V4.5 (UIC)}
\end{subfigure}
\hfill
\begin{subfigure}[b]{0.24\linewidth}
    \centering
    \includegraphics[width=\linewidth]{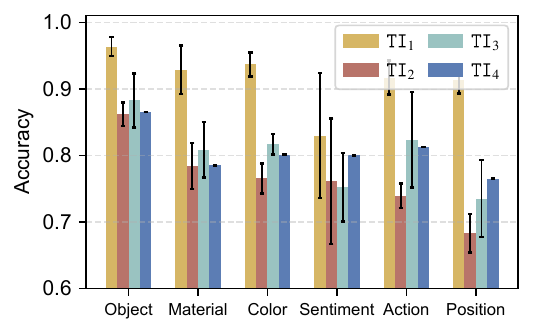}
    \caption{MiniCPM-V4.5 (DIC)}
\end{subfigure}
\hfill
\begin{subfigure}[b]{0.24\linewidth}
    \centering
    \includegraphics[width=\linewidth]{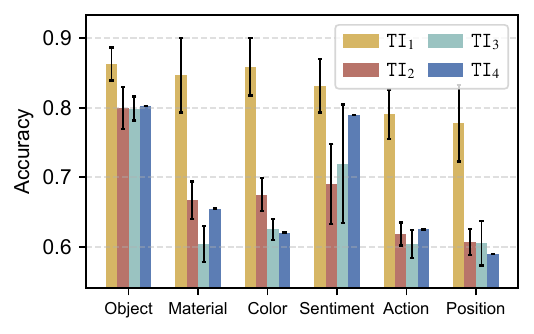}
    \caption{Qwen2.5-VL-7B (UIC)}
\end{subfigure}
\hfill
\begin{subfigure}[b]{0.24\linewidth}
    \centering
    \includegraphics[width=\linewidth]{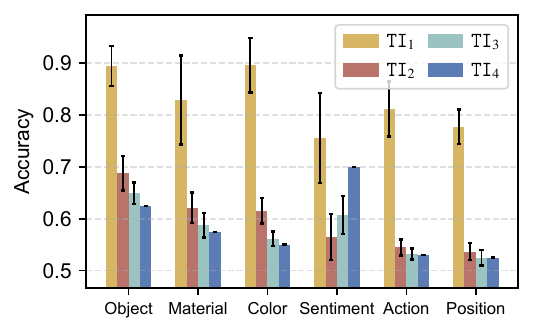}
    \caption{Qwen2.5-VL-7B (DIC)}
\end{subfigure}

\vspace{0.25cm}

\begin{subfigure}[b]{0.24\linewidth}
    \centering
    \includegraphics[width=\linewidth]{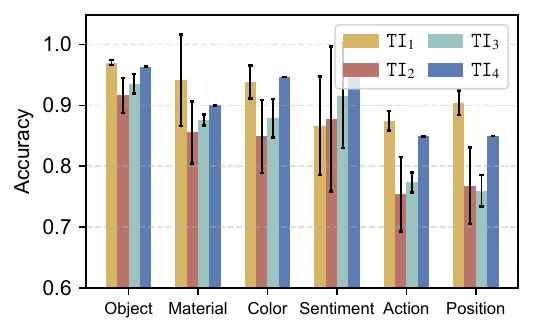}
    \caption{Qwen3-VL-8B (UIC)}
\end{subfigure}
\hfill
\begin{subfigure}[b]{0.24\linewidth}
    \centering
    \includegraphics[width=\linewidth]{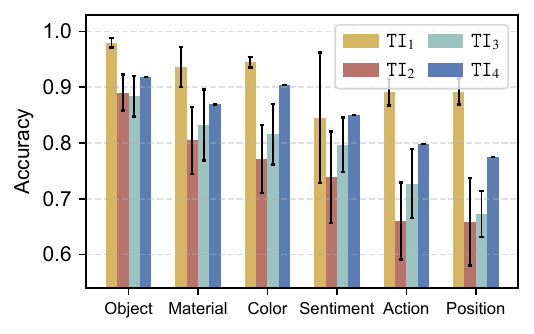}
    \caption{Qwen3-VL-8B (DIC)}
\end{subfigure}
\hfill
\begin{subfigure}[b]{0.24\linewidth}
    \centering
    \includegraphics[width=\linewidth]{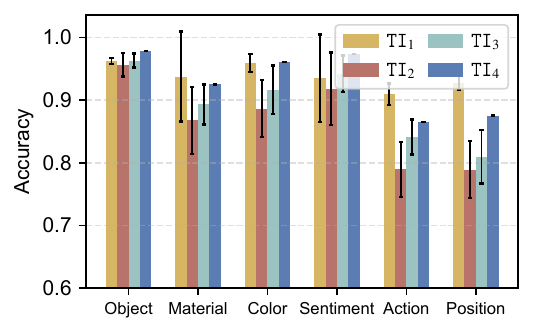}
    \caption{Qwen3-VL-32B (UIC)}
\end{subfigure}
\hfill
\begin{subfigure}[b]{0.24\linewidth}
    \centering
    \includegraphics[width=\linewidth]{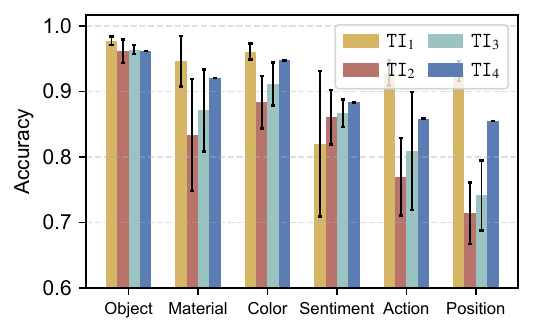}
    \caption{Qwen3-VL-32B (DIC)}
\end{subfigure}

\vspace{0.25cm}

\begin{subfigure}[b]{0.24\linewidth}
    \centering
    \includegraphics[width=\linewidth]{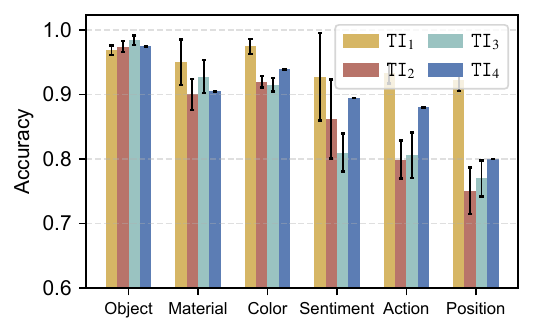}
    \caption{GPT-4o (UIC)}
\end{subfigure}
\hfill
\begin{subfigure}[b]{0.24\linewidth}
    \centering
    \includegraphics[width=\linewidth]{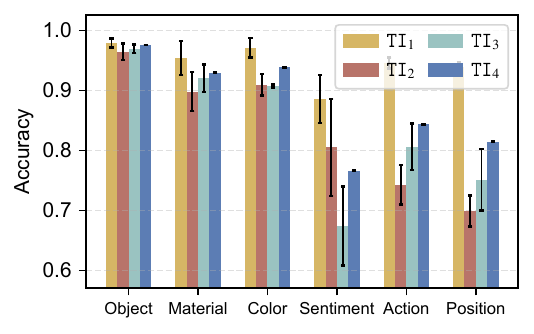}
    \caption{GPT-4o (DIC)}
\end{subfigure}
\hfill
\begin{subfigure}[b]{0.48\linewidth}
    \mbox{}
\end{subfigure}

\caption{Full multi-granularity evaluation results on LIVE for all 11 LVLMs at a temperature of 1.0. For each model, we report accuracy under Uniform Image Context (UIC, left) and Diverse Image Context (DIC, right).}
\label{fig:full_multi_granularity_curves_t10}
\end{figure*}

Figure~\ref{fig:full_multi_granularity_curves_t0} illustrates the complete multi-granularity accuracy curves for the remaining nine LVLMs evaluated on LIVE under Uniform Image Context (UIC) and Diverse Image Context (DIC) settings at $T=0$. Results for all 11 models at $T=0.7$ and $T=1.0$ are provided in Figure \ref{fig:full_multi_granularity_curves_t07} and Figure \ref{fig:full_multi_granularity_curves_t10}, respectively.

\section{Human Validation and Construction Cost}
\label{sec:human_val}

Table~\ref{tab:supp_construction_validation} summarizes the validation of the benchmark construction process. The following paragraphs define the validation unit, exclusion criteria, and task-specific checks underlying these results.

\begin{table*}[!t]
\centering
\footnotesize
\caption{Validation of the LIVE benchmark construction process.}
\label{tab:supp_construction_validation}
\setlength{\tabcolsep}{8pt}
\renewcommand{\arraystretch}{1.02}
\begin{adjustbox}{max width=\linewidth}
\begin{tabular*}{\linewidth}{@{\hspace{\tabcolsep}\extracolsep{\fill}}>{\raggedright\arraybackslash}p{0.20\linewidth}>{\raggedright\arraybackslash}p{0.28\linewidth}>{\raggedright\arraybackslash}p{0.38\linewidth}}
\toprule
\textbf{Component} & \textbf{Validation setting} & \textbf{Result} \\
\midrule
Caption-derived scene graphs & Manual evaluation of 100 images & No label-affecting error observed in the sample \\
Counterfactual items & Four expert annotators; 500 images & 98.5\% agreement; $\kappa=0.92$; 4.2\% rejected or revised \\
Sentiment items & Additional visual evaluation & 95\% judged valid \\
Image-grounded graph pilot & Qwen3-VL-32B; 500 images & Comparable object and relation counts, but less stable relation labels \\
\bottomrule
\end{tabular*}
\end{adjustbox}
\end{table*}

\paragraph{Validation Unit and Criteria}
The validation unit is a candidate MHI item associated with an image group. Four expert annotators, all graduate researchers with computer-vision backgrounds, independently assess items associated with 500 images. For each item, annotators determine whether (1) it is absent from every image for which it serves as a counterfactual and (2) it remains contextually plausible as a hallucination probe. Each validated item defines a family of questions across target-image subsets, so the assessment is performed at the item level rather than on paraphrases sharing the same visual label.

\paragraph{Exclusion Criteria and Resolution}
Annotators reject or revise a candidate when the proposed negative item is visible, the visual evidence is ambiguous, or the generated item is invalid or unrelated to the scenario. These cases account for 4.2\% of the reviewed candidates. Agreement on the independently assigned item labels is 98.5\%, with Cohen's $\kappa=0.92$.

\paragraph{Scene-Graph and Task-Specific Validation}
Before counterfactual generation, we use all available human-written MS-COCO captions for each image, remove caption-sparse samples, and examine the caption-derived graphs of 100 sampled images. The review finds no label-affecting graph error. A 500-image Qwen3-VL-32B pilot produces similar object and relation counts, although its relation labels are less stable. Sentiment items receive an additional visual evaluation of whether each image supports the factual label and excludes its counterfactual opposite. Overall, 95\% of these items are judged valid, reflecting the greater subjectivity of sentiment relative to object or attribute recognition.

\paragraph{Construction Cost}
Scene-graph parsing requires little computation relative to LLM inference. Counterfactual generation uses the open-weight Qwen2.5-72B-Instruct model locally and takes approximately four hours on two H200 GPUs. These figures describe the scale of benchmark construction and may vary with hardware.

Counterfactual generation uses local inference and incurs no direct API charges. The construction costs therefore come from GPU computation and human annotation.

\section{UIC/DIC Scenario Construction}
\label{sec:clip_sim}

UIC and DIC are drawn from the same MS-COCO source pool and follow the same item-construction procedure and task definitions. Their difference lies in image grouping. UIC samples images from the same CLIP cluster, whereas DIC samples images from distant clusters. We calculate the average within-group CLIP ViT-B/32 cosine similarity to characterize this difference.

\begin{table}[htbp]
\centering
\footnotesize
\caption{Average intra-group CLIP ViT-B/32 cosine similarity under uniform and diverse image contexts. Difference is Uniform$-$Diverse.}
\label{tab:supp_clip_similarity}
\renewcommand{\arraystretch}{1.02}
\setlength{\tabcolsep}{6pt}
\begin{adjustbox}{max width=\linewidth}
\begin{tabular}{lccc}
\toprule
\textbf{Recognition task} & \textbf{Uniform context} & \textbf{Diverse context} & \textbf{Difference} \\
\midrule
Object & 0.65 & 0.54 & 0.11 \\
Material & 0.64 & 0.55 & 0.09 \\
Color & 0.65 & 0.55 & 0.10 \\
Sentiment & 0.62 & 0.52 & 0.10 \\
Action & 0.65 & 0.56 & 0.09 \\
Position & 0.65 & 0.56 & 0.09 \\
\bottomrule
\end{tabular}
\end{adjustbox}
\end{table}

Table~\ref{tab:supp_clip_similarity} shows a difference of 0.09--0.11 for every recognition task. The significant gap confirms our scenarios possess distinct semantic distributions: UIC implies high visual cohesion, whereas DIC spans broad, heterogeneous contexts.

\section{Analysis of ``Any'' vs. ``Each'' Logical Forms}
\label{sec:appendix_any_each}

\begin{table}[tbp]
    \centering
    \scriptsize
    \setlength{\tabcolsep}{4pt}
    \renewcommand{\arraystretch}{1.1}
    \caption{Performance comparison between ``Any'' and ``Each'' logical formulations at $T=0.7$ and $T=1.0$. Gap is calculated as ($\text{Acc}_{\text{Any}} - \text{Acc}_{\text{Each}}$).}
    \label{tab:any_vs_each}
    \begin{adjustbox}{max width=\linewidth}
\begin{tabular*}{\linewidth}{@{\hspace{\tabcolsep}\extracolsep{\fill}}l ccc ccc}
        \toprule
        \multirow{2}{*}{\textbf{Model}} & \multicolumn{3}{c}{\textbf{Temperature = 0.7}} & \multicolumn{3}{c}{\textbf{Temperature = 1.0}} \\
        \cmidrule(lr){2-4} \cmidrule(lr){5-7}
        & \textbf{Any} & \textbf{Each} & \textbf{Gap} & \textbf{Any} & \textbf{Each} & \textbf{Gap} \\
        \midrule
GPT-4o & 0.874 & 0.987 & -0.113 & 0.872 & 0.982 & -0.110 \\
Qwen3-VL-32B & 0.871 & 0.981 & -0.110 & 0.872 & 0.980 & -0.108 \\
InternVL-2.5-26B & 0.809 & 0.905 & -0.096 & 0.790 & 0.892 & -0.102 \\
Qwen3-VL-8B & 0.811 & 0.957 & -0.146 & 0.813 & 0.957 & -0.144 \\
InternVL-2.5-8B & 0.790 & 0.853 & -0.063 & 0.767 & 0.846 & -0.079 \\
MiniCPM-V4.5 & 0.820 & 0.957 & -0.137 & 0.809 & 0.952 & -0.143 \\
MiniCPM-V2.6 & 0.680 & 0.927 & -0.247 & 0.680 & 0.911 & -0.231 \\
Mantis-8B & 0.758 & 0.823 & -0.065 & 0.720 & 0.796 & -0.076 \\
Qwen2.5-VL-7B & 0.602 & 0.902 & -0.300 & 0.612 & 0.882 & -0.270 \\
LLaVA-OV & 0.789 & 0.811 & -0.022 & 0.756 & 0.791 & -0.035 \\
LLaVA-IL & 0.825 & 0.952 & -0.127 & 0.785 & 0.919 & -0.134 \\
        \bottomrule
    \end{tabular*}
\end{adjustbox}
\end{table}

As noted in our main paper, the LIVE benchmark includes two distinct logical formulations for multi-image queries: existential quantification (\textit{``Is \mysquare~in \textbf{any} of...''}) and universal quantification (\textit{``Is \mysquare~in \textbf{each} of...''}). These impose fundamentally different reasoning demands. The ``any'' condition requires the model to correctly identify at least one positive occurrence across multiple images (where a single presence yields a \textsc{Yes}), whereas the ``each'' condition requires the model to verify presence in all queried images (where a single absence is sufficient to yield a \textsc{No}).

Table~\ref{tab:any_vs_each} reports accuracy under both formulations at $T=0.7$ and $T=1.0$, complementing the $T=0$ results in Table~\ref{tab:any_each_main}. Accuracy on ``each'' queries remains higher than on ``any'' queries across all tested models and temperatures. The gap varies across models and temperatures, but its direction remains unchanged.

\section{Extended Evaluation on Larger Image Contexts (\texorpdfstring{$k=5$ to $10$}{k=5 to 10})}
\label{sec:extended_k}

\begin{table*}[!tp]
\centering
\footnotesize
\renewcommand{\arraystretch}{1}
\caption{Performance with $k=5$--10 input images. F1 and accuracy are averaged across the six visual recognition tasks.}
\label{tab:k_images_results}
\begin{adjustbox}{max width=\linewidth}
\begin{tabular}{lcccccccccccccccccccccc}
    \toprule
    \scalea{\textbf{$k$}}
    & \multicolumn{2}{c}{\textbf{\scalea{GPT-4o}}}
    & \multicolumn{2}{c}{\textbf{\scalea{Qwen3-VL-32B}}}
    & \multicolumn{2}{c}{\textbf{\scalea{InternVL-2.5-26B}}}
    & \multicolumn{2}{c}{\textbf{\scalea{Qwen3-VL-8B}}}
    & \multicolumn{2}{c}{\textbf{\scalea{InternVL-2.5-8B}}}
    & \multicolumn{2}{c}{\textbf{\scalea{MiniCPM-V4.5}}}
    & \multicolumn{2}{c}{\textbf{\scalea{MiniCPM-V2.6}}}
    & \multicolumn{2}{c}{\textbf{\scalea{Mantis-8B}}}
    & \multicolumn{2}{c}{\textbf{\scalea{Qwen2.5-VL-7B}}}
    & \multicolumn{2}{c}{\textbf{\scalea{LLaVA-OV}}}
    & \multicolumn{2}{c}{\textbf{\scalea{LLaVA-IL}}} \\
    \cmidrule(lr){2-3} \cmidrule(lr){4-5} \cmidrule(lr){6-7} \cmidrule(lr){8-9} \cmidrule(lr){10-11} \cmidrule(lr){12-13} \cmidrule(lr){14-15} \cmidrule(lr){16-17} \cmidrule(lr){18-19} \cmidrule(lr){20-21} \cmidrule(lr){22-23}
    \scalea{\textbf{Metrics}} &  \scalea{F1} & \scalea{Acc} &  \scalea{F1} & \scalea{Acc} &  \scalea{F1} & \scalea{Acc} &  \scalea{F1} & \scalea{Acc} &  \scalea{F1} & \scalea{Acc} &  \scalea{F1} & \scalea{Acc} &  \scalea{F1} & \scalea{Acc} &  \scalea{F1} & \scalea{Acc} &  \scalea{F1} & \scalea{Acc} &  \scalea{F1} & \scalea{Acc} &  \scalea{F1} & \scalea{Acc}  \\
    \midrule
    \rowcolor[HTML]{f0f0f0}
     \multicolumn{23}{c}{\textit{\textbf{Scenario 1: Uniform Image Context (UIC)}}} \\
    \scalea{k5} & \scaleb{0.872} & \scaleb{0.873} & \scaleb{\bst{0.914}} & \scaleb{\bst{0.910}} & \scaleb{0.865} & \scaleb{0.864} & \scaleb{\subbst{0.902}} & \scaleb{\subbst{0.902}} & \scaleb{0.856} & \scaleb{0.869} & \scaleb{0.896} & \scaleb{0.892} & \scaleb{0.869} & \scaleb{0.876} & \scaleb{0.814} & \scaleb{0.808} & \scaleb{0.736} & \scaleb{0.800} & \scaleb{0.884} & \scaleb{0.882} & \scaleb{0.849} & \scaleb{0.820} \\
    \scalea{k6} & \scaleb{0.867} & \scaleb{0.871} & \scaleb{\bst{0.915}} & \scaleb{\bst{0.914}} & \scaleb{0.899} & \scaleb{0.897} & \scaleb{\subbst{0.910}} & \scaleb{\subbst{0.908}} & \scaleb{0.860} & \scaleb{0.872} & \scaleb{0.904} & \scaleb{0.904} & \scaleb{0.876} & \scaleb{0.885} & \scaleb{0.818} & \scaleb{0.814} & \scaleb{0.763} & \scaleb{0.813} & \scaleb{0.874} & \scaleb{0.874} & \scaleb{0.834} & \scaleb{0.806} \\
    \scalea{k7} & \scaleb{0.870} & \scaleb{0.875} & \scaleb{\bst{0.899}} & \scaleb{\bst{0.901}} & \scaleb{0.841} & \scaleb{0.843} & \scaleb{\subbst{0.893}} & \scaleb{\subbst{0.896}} & \scaleb{0.835} & \scaleb{0.848} & \scaleb{0.891} & \scaleb{0.891} & \scaleb{0.841} & \scaleb{0.854} & \scaleb{0.760} & \scaleb{0.757} & \scaleb{0.721} & \scaleb{0.791} & \scaleb{0.836} & \scaleb{0.841} & \scaleb{0.830} & \scaleb{0.799} \\
    \scalea{k8} & \scaleb{0.877} & \scaleb{0.881} & \scaleb{\bst{0.913}} & \scaleb{\bst{0.913}} & \scaleb{0.849} & \scaleb{0.849} & \scaleb{\subbst{0.896}} & \scaleb{\subbst{0.896}} & \scaleb{0.834} & \scaleb{0.855} & \scaleb{0.891} & \scaleb{0.891} & \scaleb{0.844} & \scaleb{0.860} & \scaleb{0.752} & \scaleb{0.756} & \scaleb{0.713} & \scaleb{0.792} & \scaleb{0.863} & \scaleb{0.865} & \scaleb{0.829} & \scaleb{0.792} \\
    \scalea{k9} & \scaleb{0.867} & \scaleb{0.871} & \scaleb{\subbst{0.873}} & \scaleb{\bst{0.876}} & \scaleb{0.846} & \scaleb{0.847} & \scaleb{0.861} & \scaleb{0.869} & \scaleb{0.833} & \scaleb{0.852} & \scaleb{\bst{0.874}} & \scaleb{\subbst{0.872}} & \scaleb{0.832} & \scaleb{0.849} & \scaleb{0.767} & \scaleb{0.771} & \scaleb{0.677} & \scaleb{0.770} & \scaleb{0.829} & \scaleb{0.838} & \scaleb{0.816} & \scaleb{0.778} \\
    \scalea{k10} & \scaleb{0.876} & \scaleb{0.875} & \scaleb{\subbst{0.894}} & \scaleb{\subbst{0.889}} & \scaleb{0.822} & \scaleb{0.836} & \scaleb{0.883} & \scaleb{0.887} & \scaleb{0.811} & \scaleb{0.841} & \scaleb{\bst{0.897}} & \scaleb{\bst{0.899}} & \scaleb{0.821} & \scaleb{0.846} & \scaleb{0.745} & \scaleb{0.745} & \scaleb{0.710} & \scaleb{0.785} & \scaleb{0.800} & \scaleb{0.817} & \scaleb{0.801} & \scaleb{0.757} \\
    \midrule
    \rowcolor[HTML]{f0f0f0}
    \multicolumn{23}{c}{\textit{\textbf{Scenario 2: Diverse Image Context (DIC)}}} \\
    \scalea{k5} & \scaleb{0.886} & \scaleb{0.889} & \scaleb{\subbst{0.897}} & \scaleb{\subbst{0.891}} & \scaleb{0.875} & \scaleb{0.878} & \scaleb{0.889} & \scaleb{0.885} & \scaleb{0.834} & \scaleb{0.854} & \scaleb{\bst{0.900}} & \scaleb{\bst{0.899}} & \scaleb{0.848} & \scaleb{0.863} & \scaleb{0.756} & \scaleb{0.767} & \scaleb{0.664} & \scaleb{0.765} & \scaleb{0.840} & \scaleb{0.848} & \scaleb{0.816} & \scaleb{0.795} \\
    \scalea{k6} & \scaleb{0.858} & \scaleb{0.866} & \scaleb{0.864} & \scaleb{0.887} & \scaleb{0.858} & \scaleb{0.858} & \scaleb{\subbst{0.865}} & \scaleb{\subbst{0.888}} & \scaleb{0.787} & \scaleb{0.830} & \scaleb{\bst{0.890}} & \scaleb{\bst{0.903}} & \scaleb{0.791} & \scaleb{0.831} & \scaleb{0.716} & \scaleb{0.752} & \scaleb{0.592} & \scaleb{0.737} & \scaleb{0.811} & \scaleb{0.838} & \scaleb{0.826} & \scaleb{0.800} \\
    \scalea{k7} & \scaleb{0.826} & \scaleb{0.855} & \scaleb{\bst{0.900}} & \scaleb{\subbst{0.897}} & \scaleb{0.826} & \scaleb{0.839} & \scaleb{0.863} & \scaleb{0.874} & \scaleb{0.842} & \scaleb{0.857} & \scaleb{\subbst{0.896}} & \scaleb{\bst{0.898}} & \scaleb{0.861} & \scaleb{0.875} & \scaleb{0.737} & \scaleb{0.761} & \scaleb{0.694} & \scaleb{0.777} & \scaleb{0.847} & \scaleb{0.851} & \scaleb{0.821} & \scaleb{0.791} \\
    \scalea{k8} & \scaleb{0.863} & \scaleb{0.873} & \scaleb{\bst{0.907}} & \scaleb{\bst{0.903}} & \scaleb{0.847} & \scaleb{0.852} & \scaleb{0.883} & \scaleb{0.877} & \scaleb{0.820} & \scaleb{0.840} & \scaleb{\subbst{0.902}} & \scaleb{\subbst{0.901}} & \scaleb{0.852} & \scaleb{0.865} & \scaleb{0.692} & \scaleb{0.715} & \scaleb{0.636} & \scaleb{0.743} & \scaleb{0.830} & \scaleb{0.829} & \scaleb{0.794} & \scaleb{0.752} \\
    \scalea{k9} & \scaleb{0.852} & \scaleb{0.864} & \scaleb{0.866} & \scaleb{\subbst{0.879}} & \scaleb{0.828} & \scaleb{0.842} & \scaleb{\subbst{0.867}} & \scaleb{0.873} & \scaleb{0.775} & \scaleb{0.818} & \scaleb{\bst{0.875}} & \scaleb{\bst{0.884}} & \scaleb{0.771} & \scaleb{0.812} & \scaleb{0.701} & \scaleb{0.722} & \scaleb{0.590} & \scaleb{0.726} & \scaleb{0.777} & \scaleb{0.804} & \scaleb{0.792} & \scaleb{0.769} \\
    \scalea{k10} & \scaleb{0.832} & \scaleb{0.856} & \scaleb{\bst{0.886}} & \scaleb{\bst{0.884}} & \scaleb{0.832} & \scaleb{0.842} & \scaleb{0.864} & \scaleb{0.857} & \scaleb{0.802} & \scaleb{0.835} & \scaleb{\subbst{0.867}} & \scaleb{\subbst{0.865}} & \scaleb{0.747} & \scaleb{0.803} & \scaleb{0.660} & \scaleb{0.693} & \scaleb{0.573} & \scaleb{0.721} & \scaleb{0.792} & \scaleb{0.814} & \scaleb{0.816} & \scaleb{0.794} \\
    \bottomrule
\end{tabular}
\end{adjustbox}

\begin{tablenotes}
    \item  \scriptsize \textit{Note}:  \bst{Bold} and \subbst{underlined} denote best and second-best results (excluding GPT-4o), respectively.
\end{tablenotes}
\end{table*}

We extend the primary four-image evaluation to larger input contexts containing $k=5$ to $k=10$ images.

For $k=5$--10, we retain the anchor extraction, counterfactual generation, filtering, and task definitions used for the primary four-image benchmark. Exhaustive enumeration becomes impractical as $k$ grows, so we sample target-image subsets while retaining balanced yes/no labels and coverage across the six recognition tasks. Labels follow from the membership of validated factual or counterfactual items in each sampled target set. These sampled settings extend the primary four-image evaluation but are not included in its MAP score. Table~\ref{tab:k_images_results} reports the model-wise results. Each $k$-image configuration contains 4,250--7,062 balanced binary questions with coverage across the six visual recognition tasks. For conciseness and ease of comparison, we report the averaged F1 scores and accuracy across all six tasks.

Table~\ref{tab:k_images_results} supports two observations:

\ding{182} \textbf{Lower Performance at $k=10$ than at $k=5$.} In these sampled settings, most models obtain lower F1 and accuracy at $k=10$ than at $k=5$. For instance, in UIC scenarios, Qwen3-VL-32B's F1 score decreases from 0.914 at $k=5$ to 0.894 at $k=10$, while Qwen2.5-VL-7B's F1 score drops from 0.736 to 0.710.

\ding{183} \textbf{Persistent Context Interference in DIC Scenarios.} Reaffirming our findings from the main text, Diverse Image Contexts (DIC) generally induce higher hallucination rates than Uniform Image Contexts (UIC) across larger image sets. For example, at $k=6$, Qwen3-VL-8B achieves an F1 of 0.910 in UIC but drops notably to 0.865 in DIC. Similarly, at the extreme $k=10$ setting, Qwen2.5-VL-7B exhibits a severe drop from an F1 of 0.710 in UIC to 0.573 in DIC. These results show that performance differences between UIC and DIC persist in larger image contexts.

\section{Extension Beyond Binary Questioning}
\label{sec:extended_formats}

We extend LIVE to Image Identification, Multiple-Choice QA, and Open-Ended QA, with 3,000, 3,000, and 2,909 samples, respectively, covering the six visual recognition tasks. Table~\ref{tab:format_extensions} summarizes their questions and scoring rules. Figure~\ref{fig:new_tasks_acc} presents the results, and Appendix~\ref{sec:more_case_studies} provides the complete qualitative examples.

\begin{table}[htbp]
\centering
\footnotesize
\caption{Non-binary extensions of LIVE. Each format covers the six visual tasks.}
\label{tab:format_extensions}
\setlength{\tabcolsep}{6.5pt}
\renewcommand{\arraystretch}{1.02}
\begin{adjustbox}{max width=\linewidth}
\begin{tabular}{>{\raggedright\arraybackslash}p{0.25\columnwidth}>{\raggedright\arraybackslash}p{0.29\columnwidth}>{\raggedright\arraybackslash}p{0.30\columnwidth}}
\toprule
\textbf{Format} & \textbf{Primary probe} & \textbf{Scoring} \\
\midrule
Image Identification & Localize every image containing the queried item & Exact set match over image indices \\
\addlinespace[3pt]
Multiple-Choice QA & Select one answer from four options & Exact option match \\
\addlinespace[3pt]
Open-Ended QA & Describe a queried visual property without a fixed answer list & Qwen2.5-72B-Instruct judge against concise references \\
\bottomrule
\end{tabular}
\end{adjustbox}
\end{table}

\begin{figure*}[!t]
    \centering
    \begin{subfigure}[t]{0.47\linewidth}
        \centering
        \vspace{0pt}
        \includegraphics[width=\linewidth]{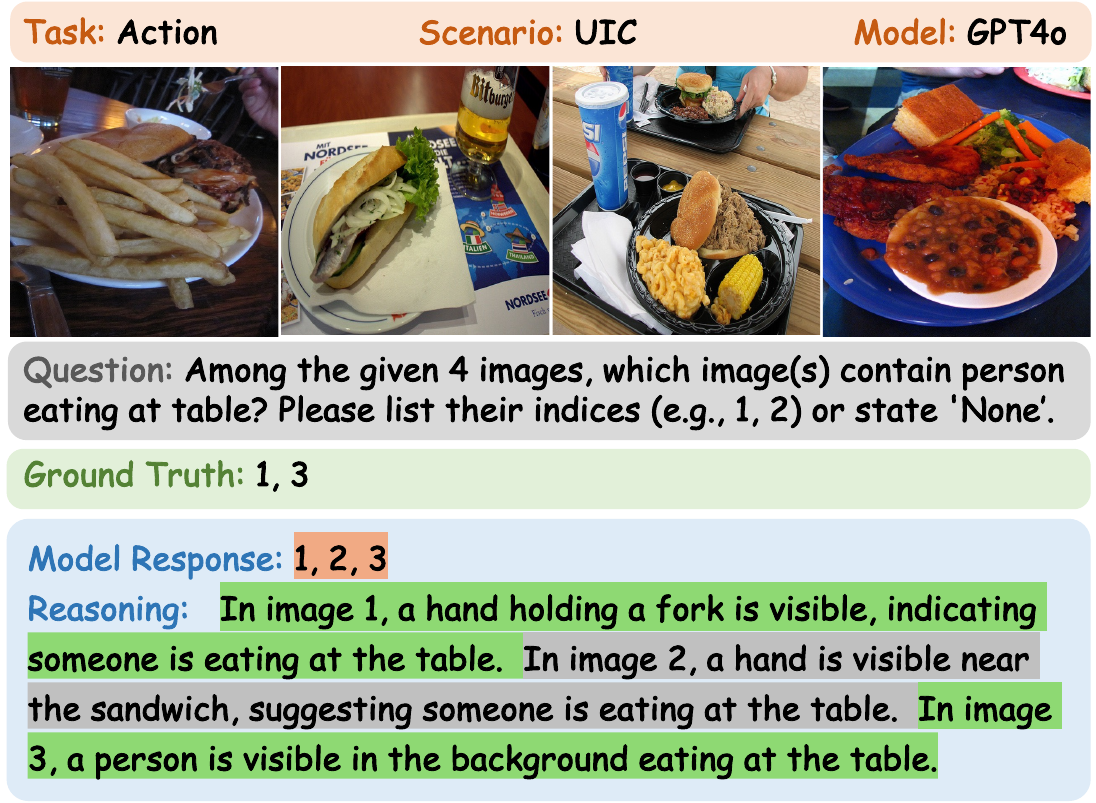}
    \end{subfigure}\hfill
    \begin{subfigure}[t]{0.47\linewidth}
        \centering
        \vspace{0pt}
        \includegraphics[width=\linewidth]{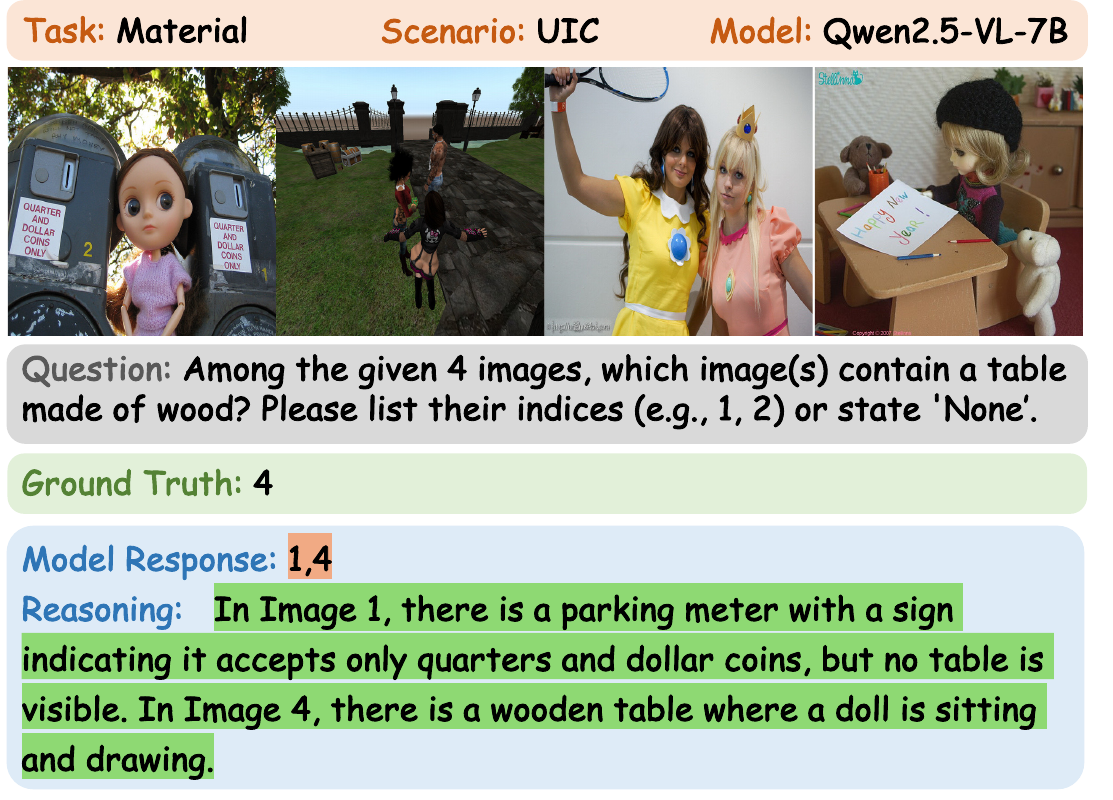}
    \end{subfigure}

    \par\vspace{2mm}
    \begin{subfigure}[t]{0.47\linewidth}
        \centering
        \vspace{0pt}
        \includegraphics[width=\linewidth]{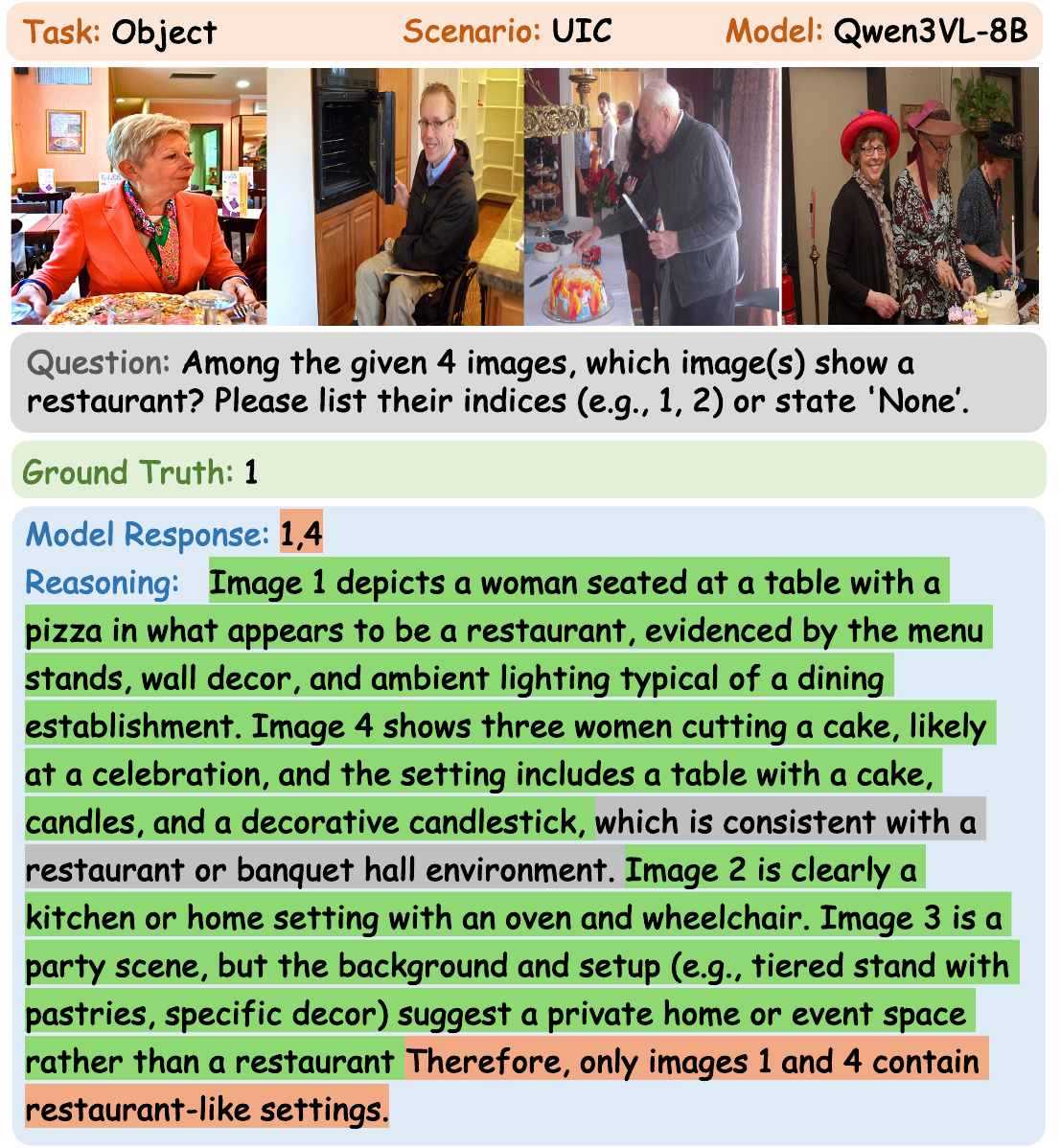}
    \end{subfigure}\hfill
    \begin{subfigure}[t]{0.47\linewidth}
        \centering
        \vspace{0pt}
        \includegraphics[width=\linewidth]{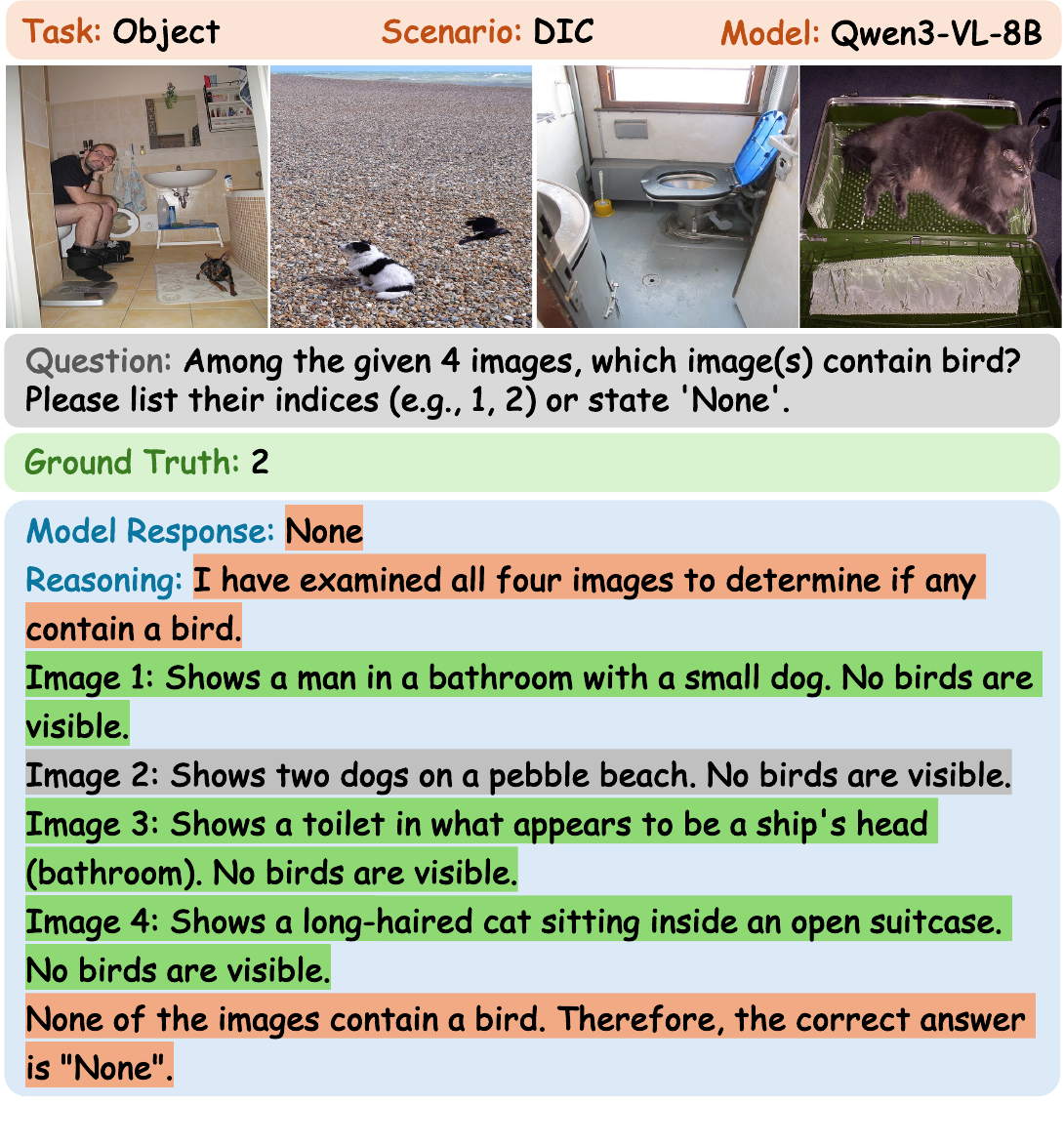}
    \end{subfigure}

    \caption{Image Identification examples under UIC and DIC.}
    \label{fig:case_study_id}
\end{figure*}

\paragraph{Image Identification}
Existence questions are transformed into localization prompts of the form: \textit{``Among the given four images, which image(s) contain [item]? List their indices or state `None'.''} The reference answer is the complete set of matching image indices, and a prediction is correct only when the parsed set exactly matches that reference. This format directly penalizes content transfer: adding an index because an item from one image is hallucinated in a visually similar image makes the complete set incorrect.

\paragraph{Multiple-Choice QA}
We construct four-option factual and exclusion questions, such as \textit{``Which option is not present in any of Images X and Y?''} The options are drawn from the factual and counterfactual item pools produced in Stage~2. We use exact option-letter matching, with a 25\% chance level for uniform random guessing.

\paragraph{Open-Ended QA}
Structured items are also converted into free-form factual prompts, for example, \textit{``What type of road is visible in any of Images 1 and 4?''} Since equivalent answers may use different surface forms, Qwen2.5-72B-Instruct judges the predicted response against a concise reference answer, accepting paraphrases that preserve its essential facts.

\section{More Case Studies}
\label{sec:more_case_studies}

\begin{figure*}[!tbp]
    \centering
    \begin{subfigure}[t]{0.46\linewidth}
        \centering
        \vspace{0pt}
        \includegraphics[width=\linewidth]{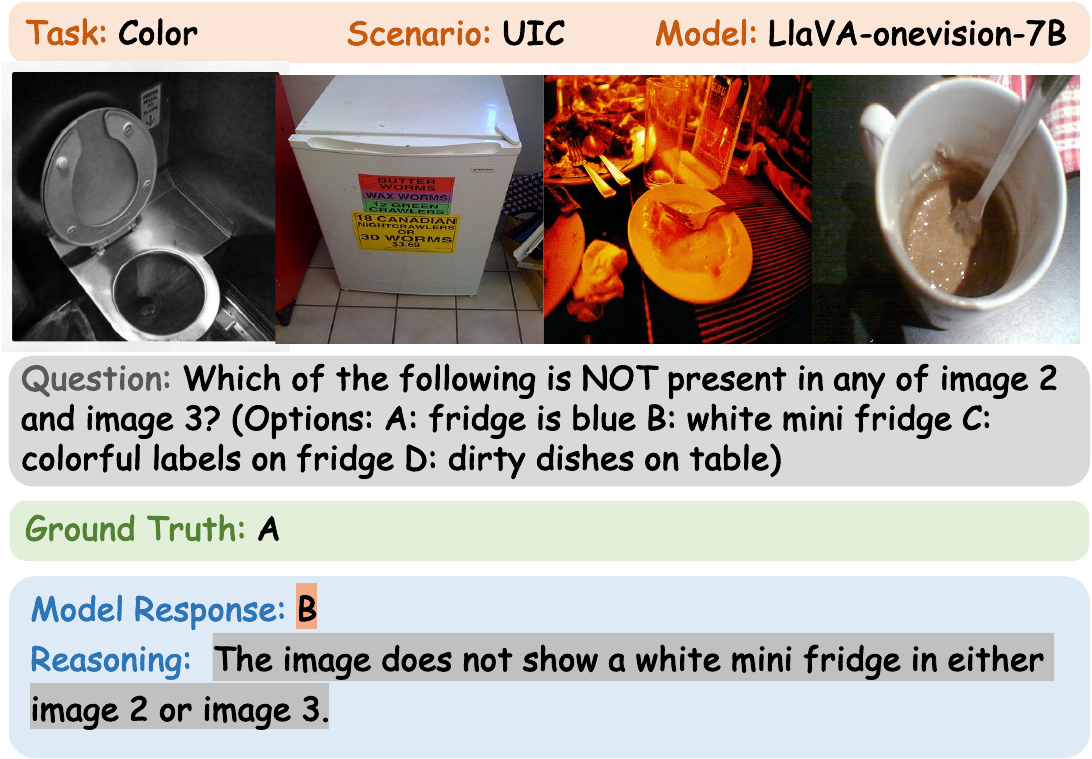}
    \end{subfigure}\hfill
    \begin{subfigure}[t]{0.46\linewidth}
        \centering
        \vspace{0pt}
        \includegraphics[width=\linewidth]{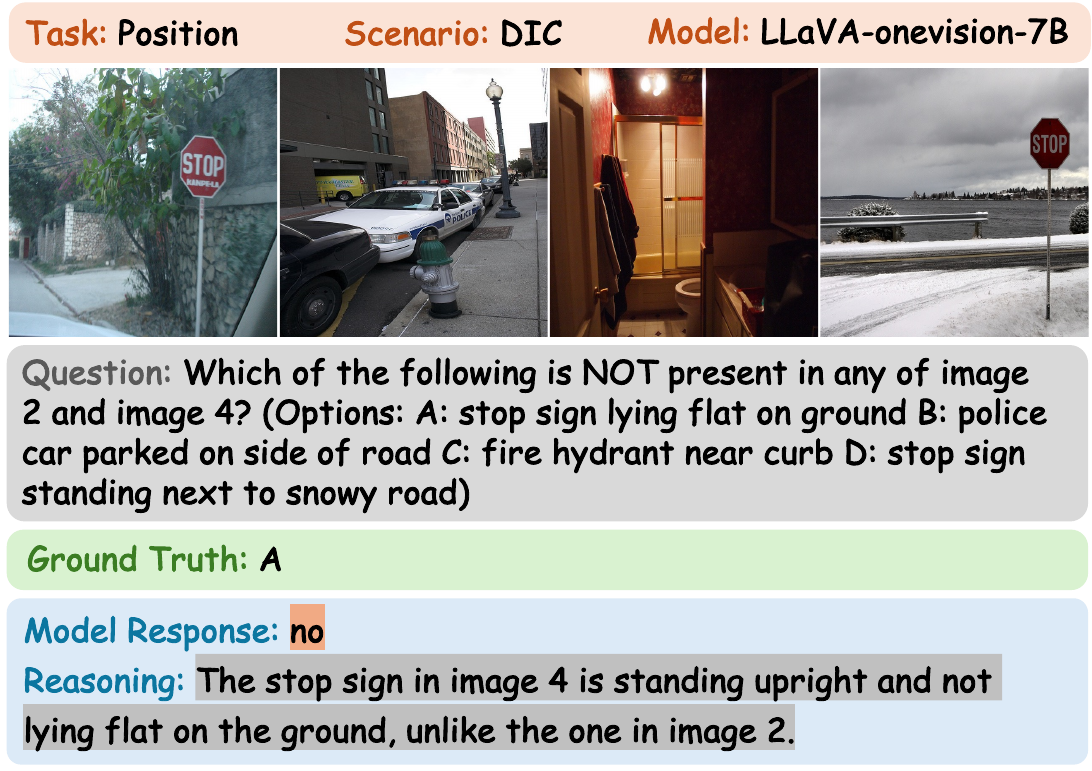}
    \end{subfigure}

    \par\vspace{2mm}
    \begin{subfigure}[t]{0.46\linewidth}
        \centering
        \vspace{0pt}
        \includegraphics[width=\linewidth]{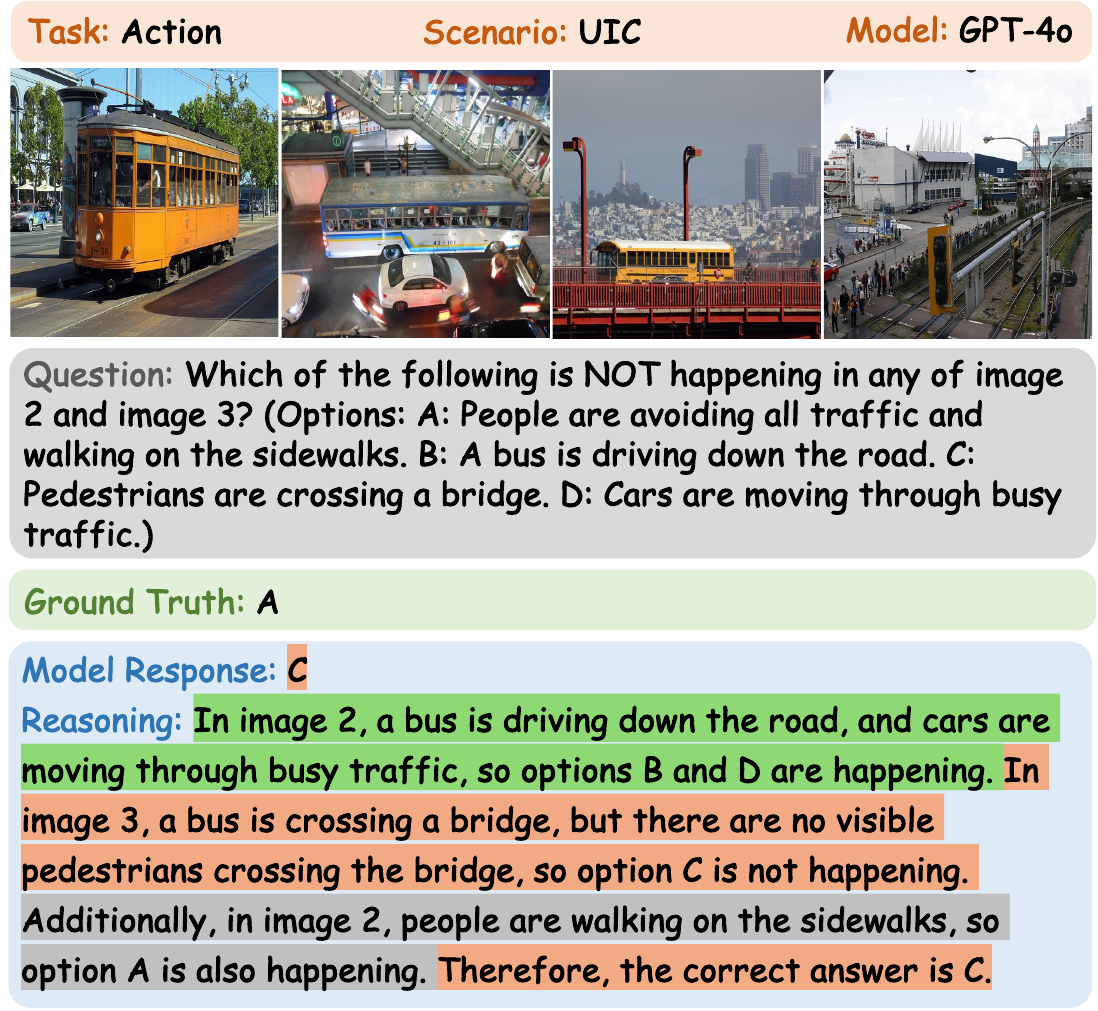}
    \end{subfigure}\hfill
    \begin{subfigure}[t]{0.46\linewidth}
        \centering
        \vspace{0pt}
        \includegraphics[width=\linewidth]{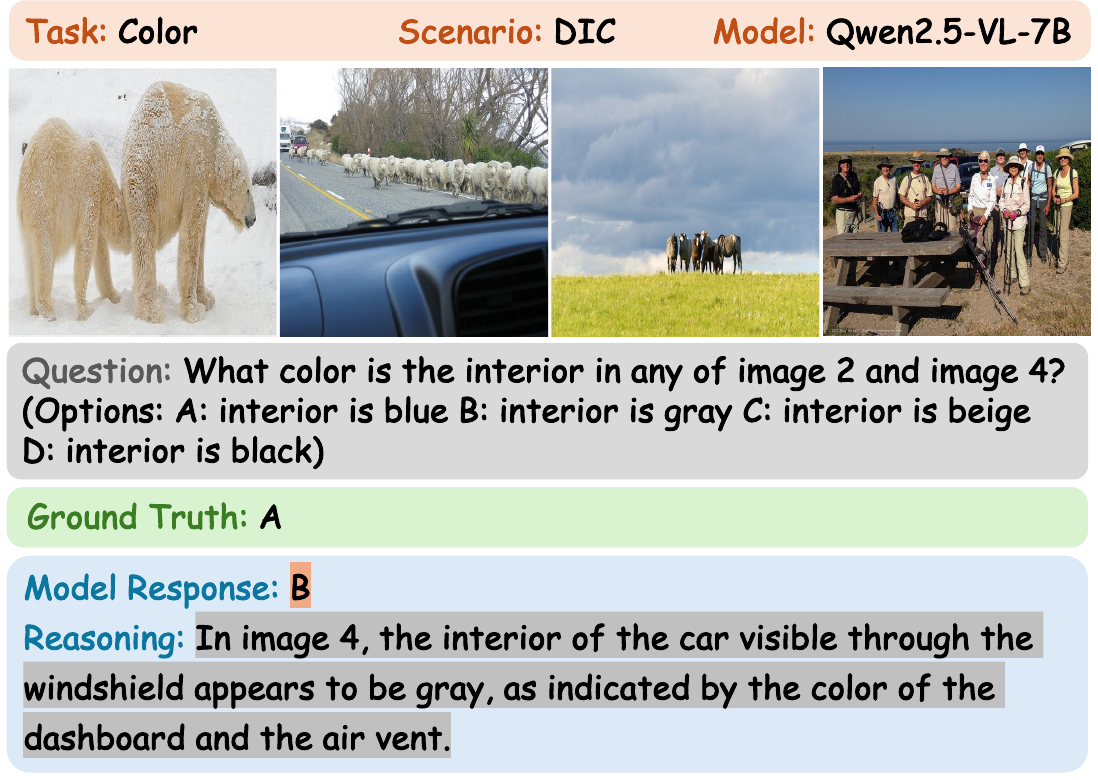}
    \end{subfigure}

    \caption{Multiple-Choice QA examples under UIC (left) and DIC (right).}
    \label{fig:case_study_mcq}
\end{figure*}

\begin{figure*}[!tbp]
    \centering
    \begin{subfigure}[t]{0.46\linewidth}
        \centering
        \vspace{0pt}
        \includegraphics[width=\linewidth]{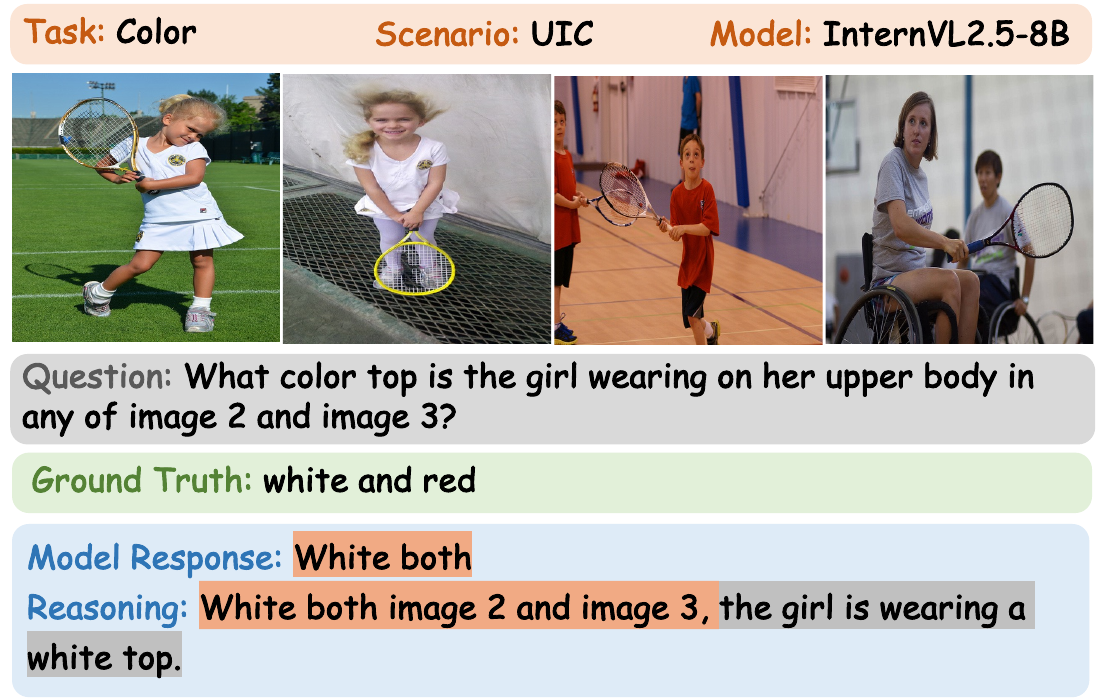}
        \par\vspace{2mm}
        \includegraphics[width=\linewidth]{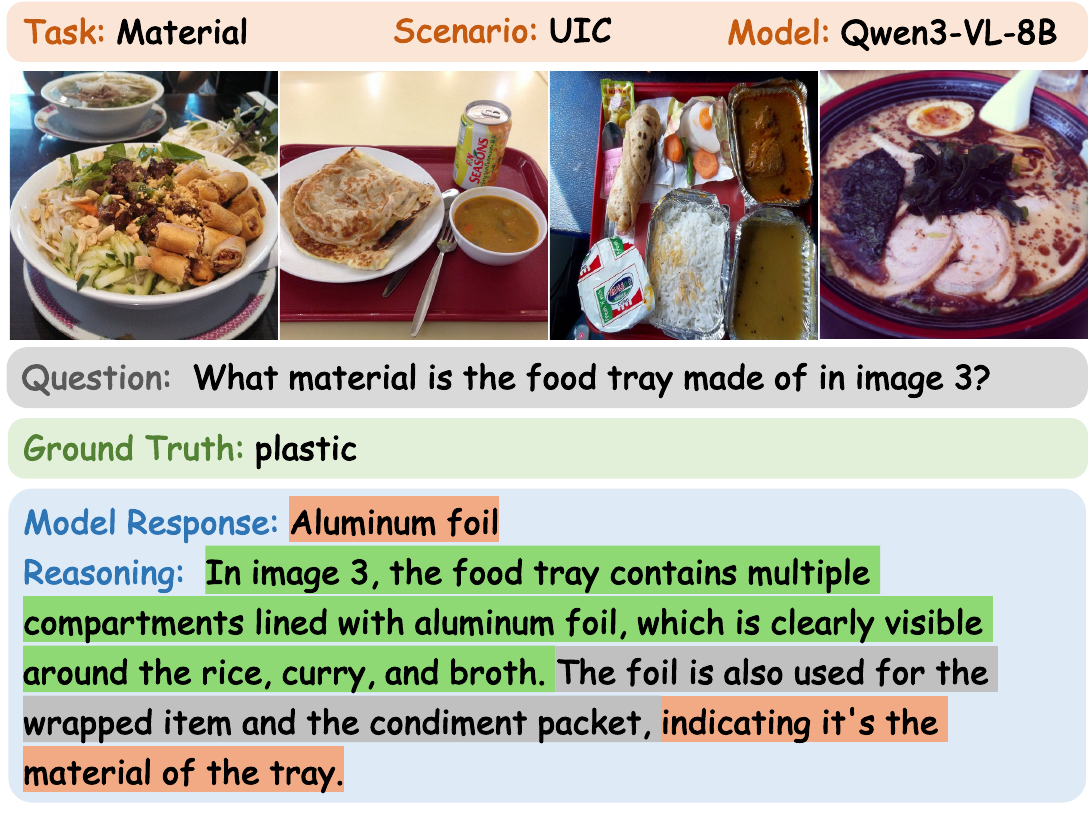}
    \end{subfigure}\hfill
    \begin{subfigure}[t]{0.46\linewidth}
        \centering
        \vspace{0pt}
        \includegraphics[width=\linewidth]{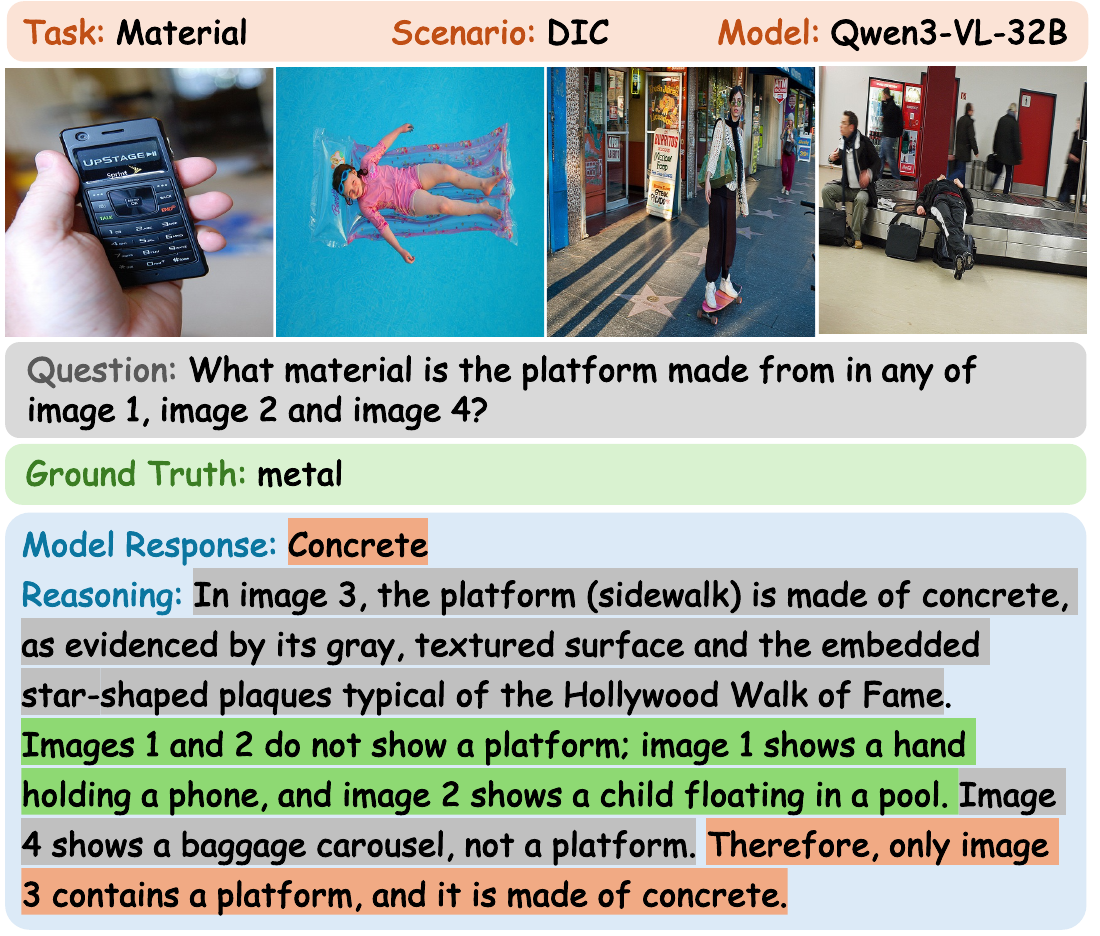}
    \end{subfigure}

    \caption{Open-Ended QA examples under UIC (left) and DIC (right).}
    \label{fig:case_study_openqa}
\end{figure*}

\begin{figure*}[!tbp]
    \centering
    \begin{subfigure}[t]{0.47\linewidth}
        \centering
        \vspace{0pt}
        \includegraphics[width=\linewidth]{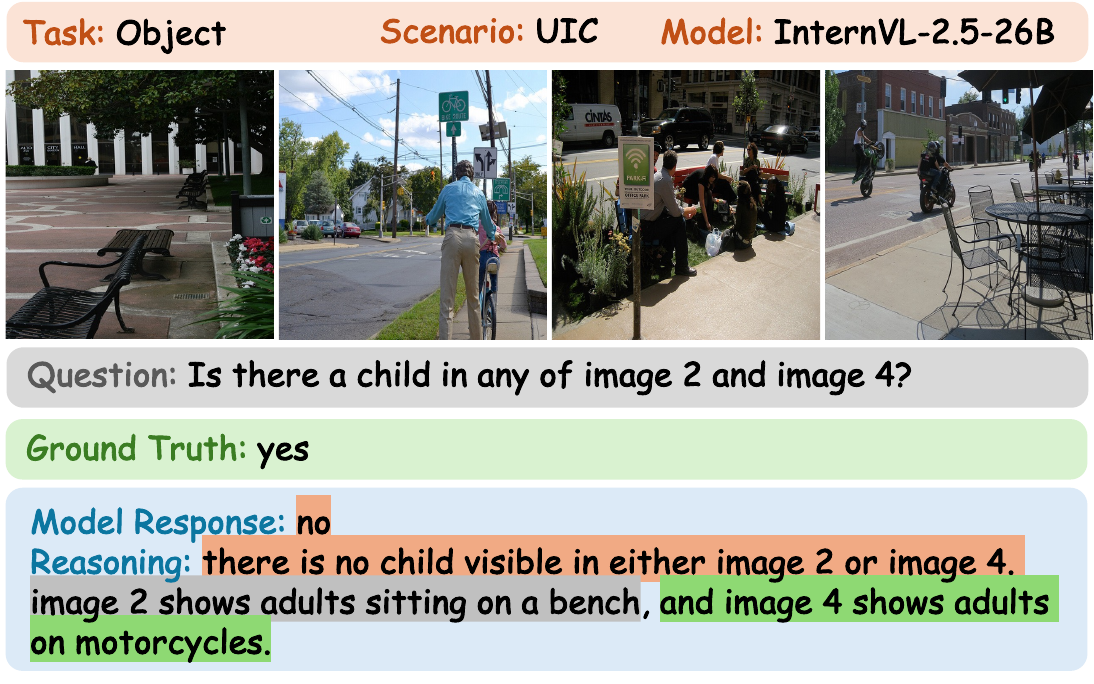}
    \end{subfigure}\hfill
    \begin{subfigure}[t]{0.47\linewidth}
        \centering
        \vspace{0pt}
        \includegraphics[width=\linewidth]{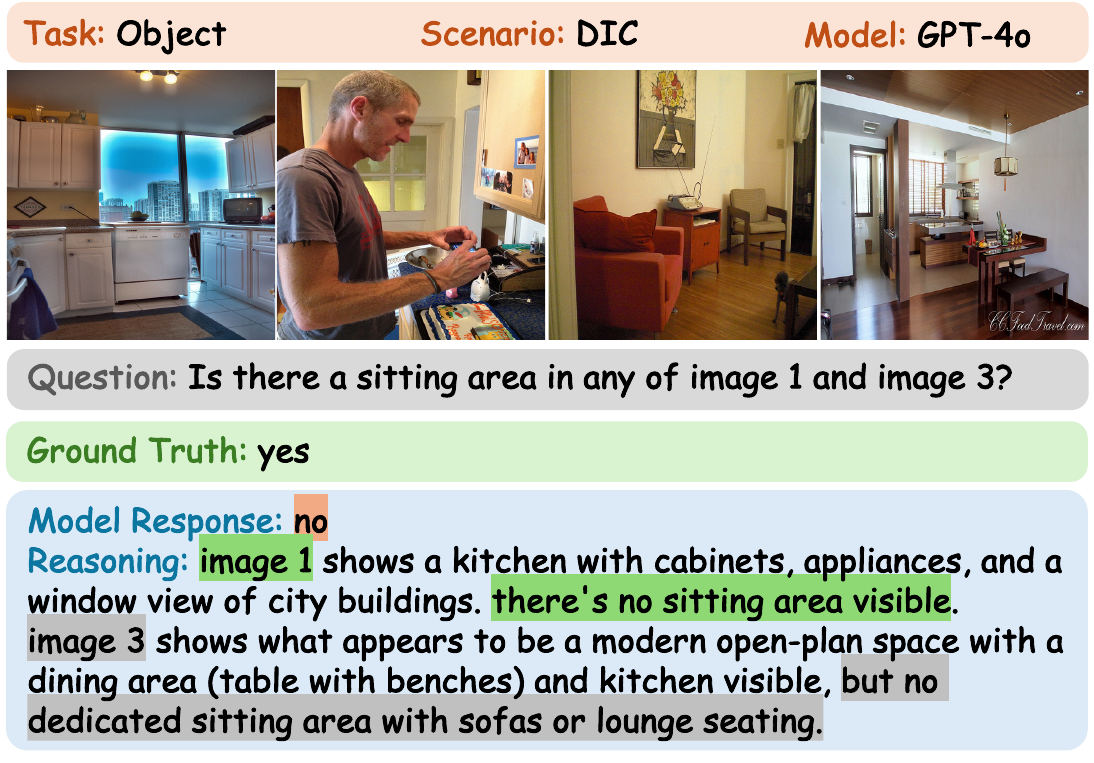}
    \end{subfigure}

    \par\vspace{2mm}
    \begin{subfigure}[t]{0.47\linewidth}
        \centering
        \vspace{0pt}
        \includegraphics[width=\linewidth]{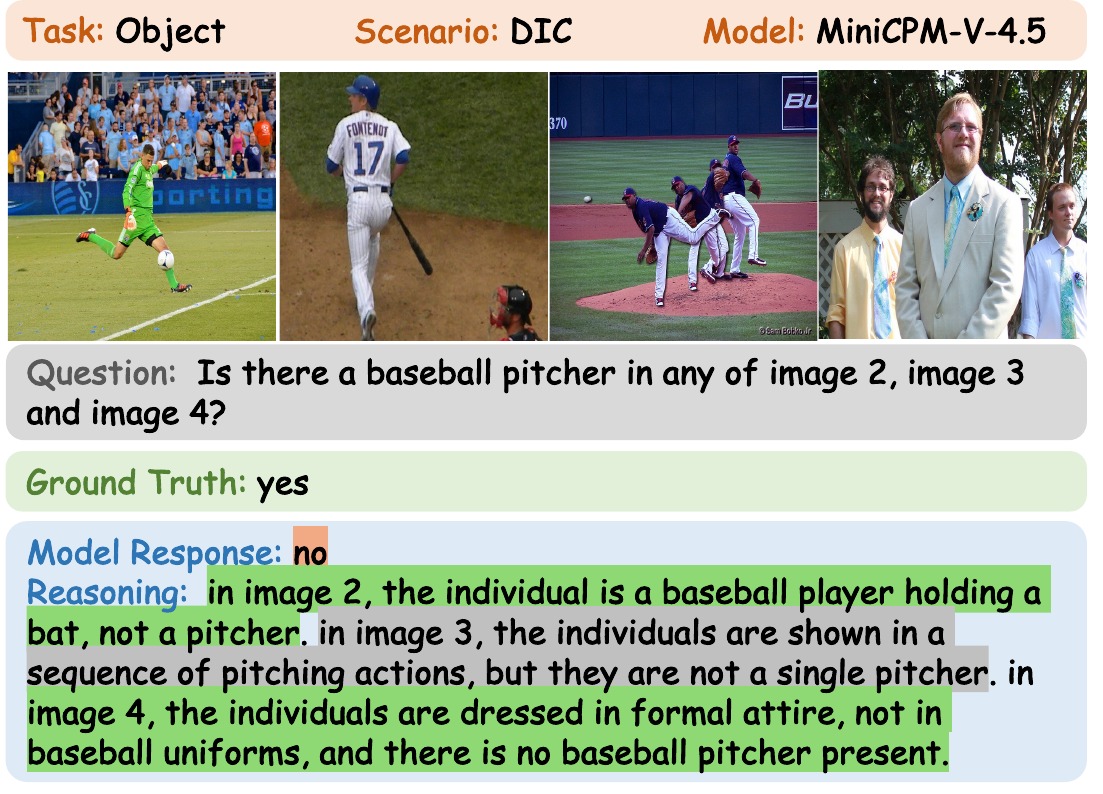}
    \end{subfigure}\hfill
    \begin{subfigure}[t]{0.47\linewidth}
        \centering
        \vspace{0pt}
        \includegraphics[width=\linewidth]{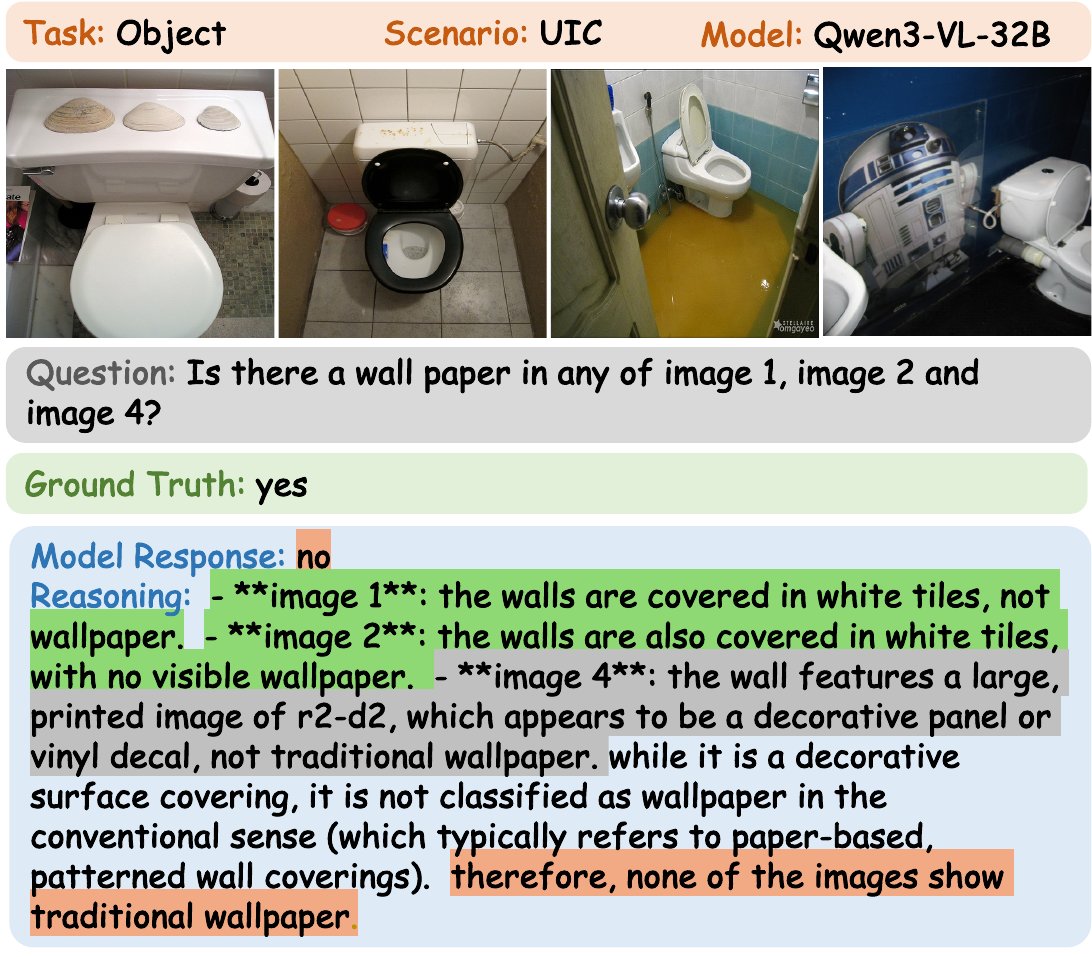}
    \end{subfigure}

    \par\vspace{2mm}
    \begin{subfigure}[t]{0.47\linewidth}
        \centering
        \vspace{0pt}
        \includegraphics[width=\linewidth]{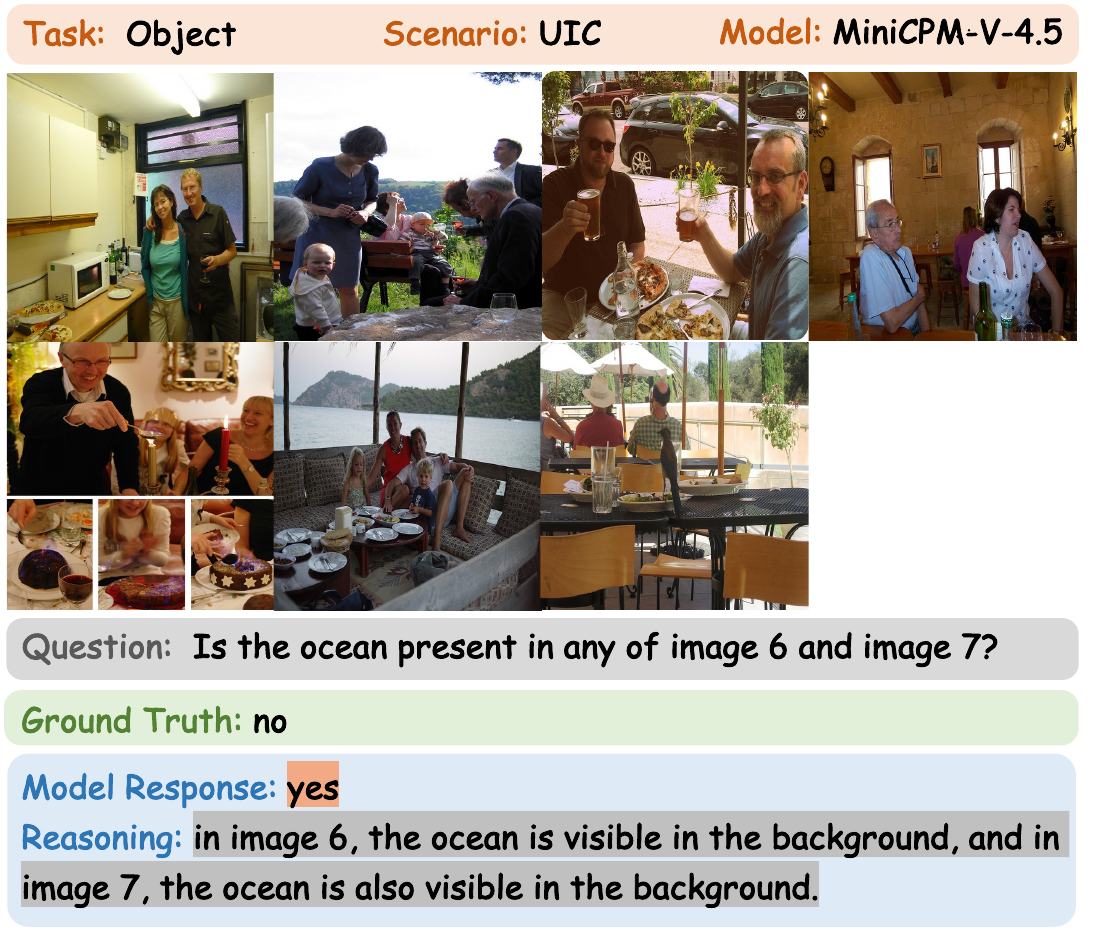}
    \end{subfigure}\hfill
    \begin{subfigure}[t]{0.47\linewidth}
        \centering
        \vspace{0pt}
        \includegraphics[width=\linewidth]{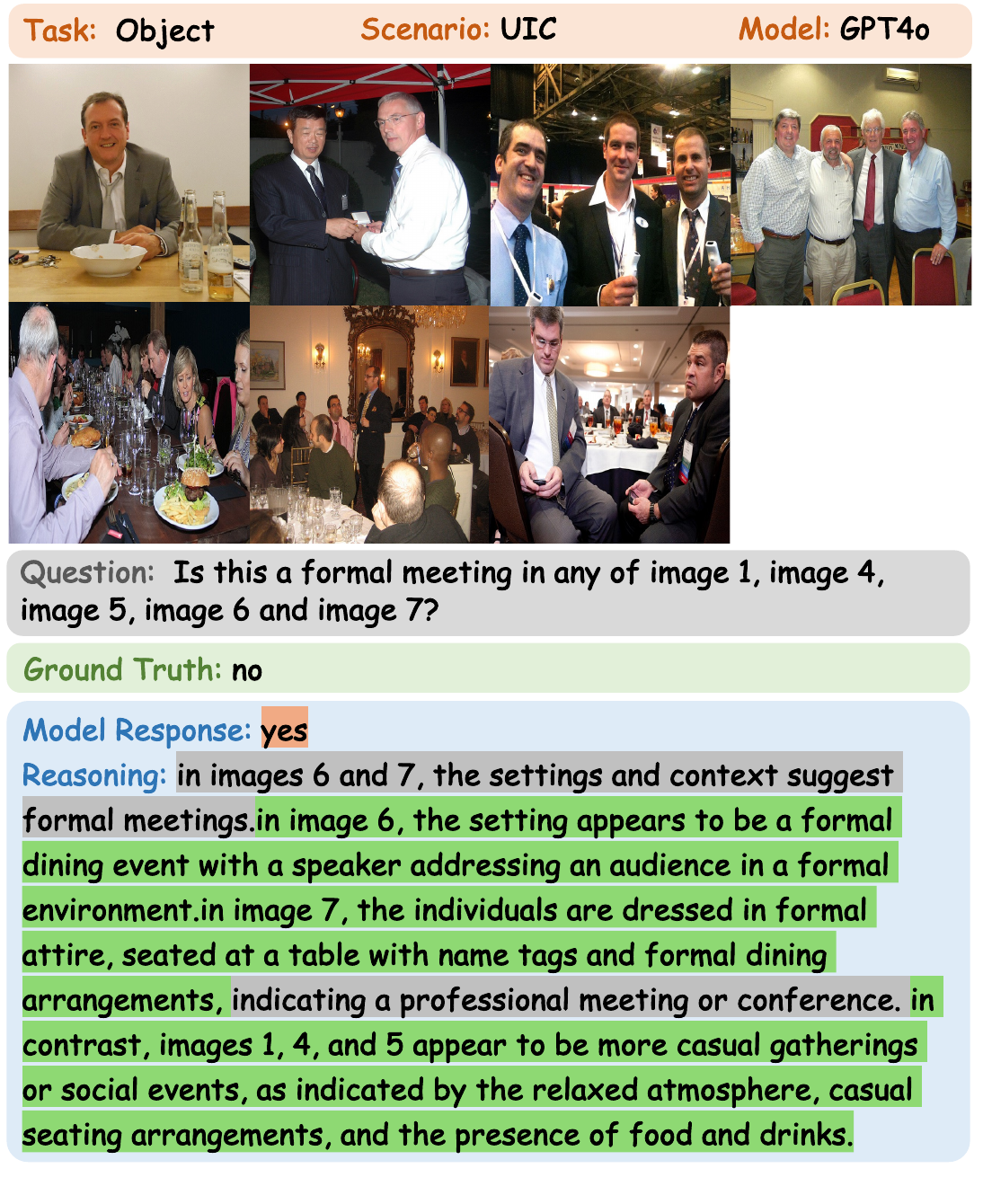}
    \end{subfigure}

    \caption{Object recognition examples under UIC and DIC ($k=4$--7).}
    \label{fig:case_study_k_scale_1}
\end{figure*}

\begin{figure*}[!tbp]
    \centering
    \begin{subfigure}[t]{0.47\linewidth}
        \centering
        \vspace{0pt}
        \includegraphics[width=\linewidth]{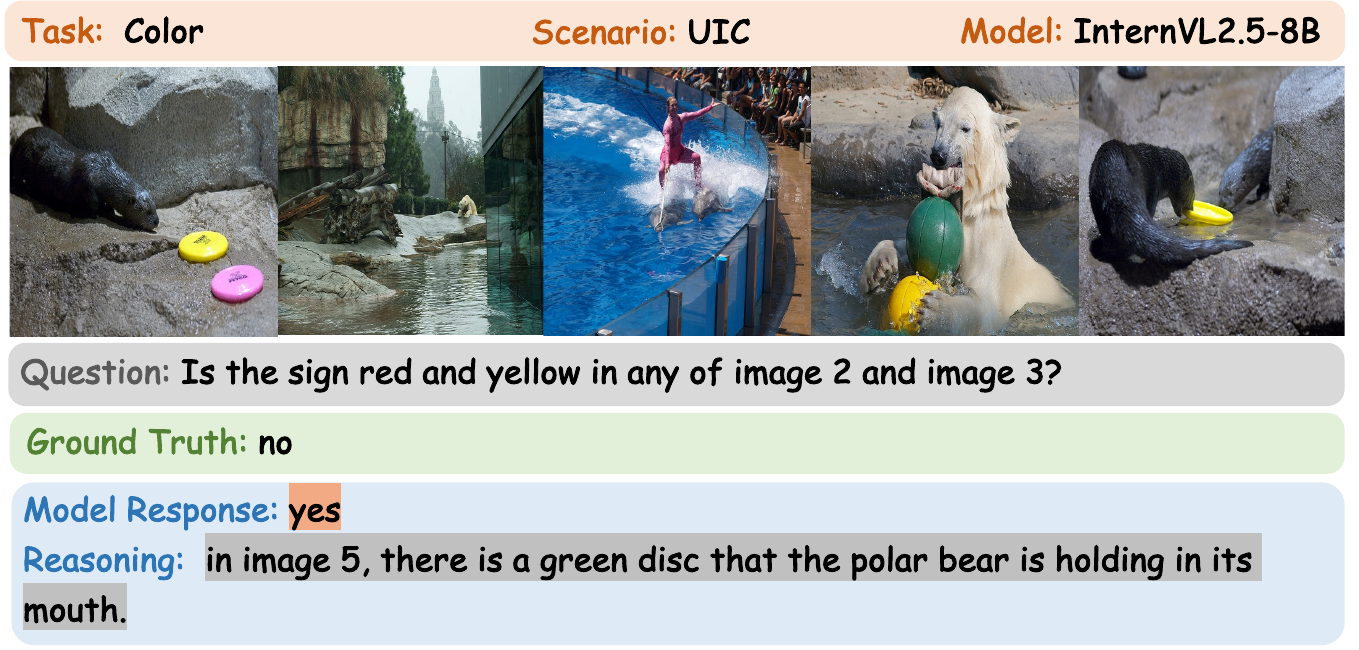}
    \end{subfigure}\hfill
    \begin{subfigure}[t]{0.47\linewidth}
        \centering
        \vspace{0pt}
        \includegraphics[width=\linewidth]{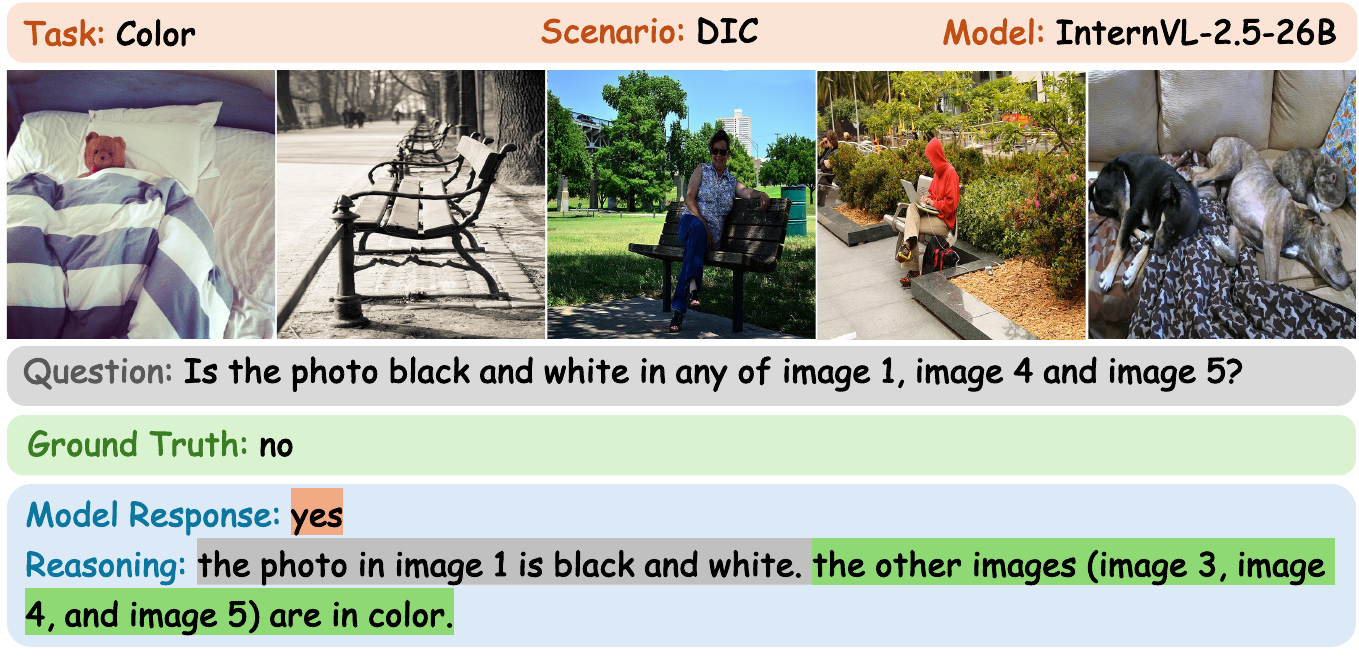}
    \end{subfigure}

    \par\vspace{2mm}
    \begin{subfigure}[t]{0.47\linewidth}
        \centering
        \vspace{0pt}
        \includegraphics[width=\linewidth]{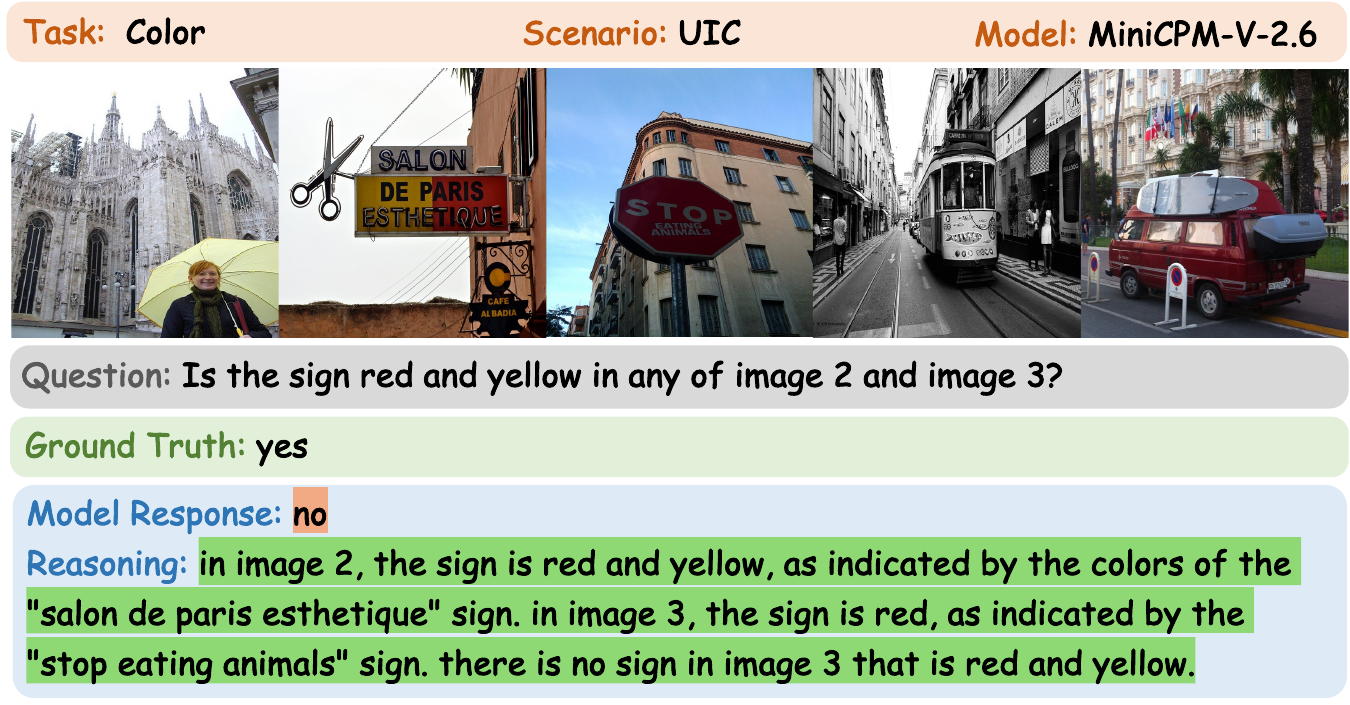}
    \end{subfigure}\hfill
    \begin{subfigure}[t]{0.47\linewidth}
        \centering
        \vspace{0pt}
        \includegraphics[width=\linewidth]{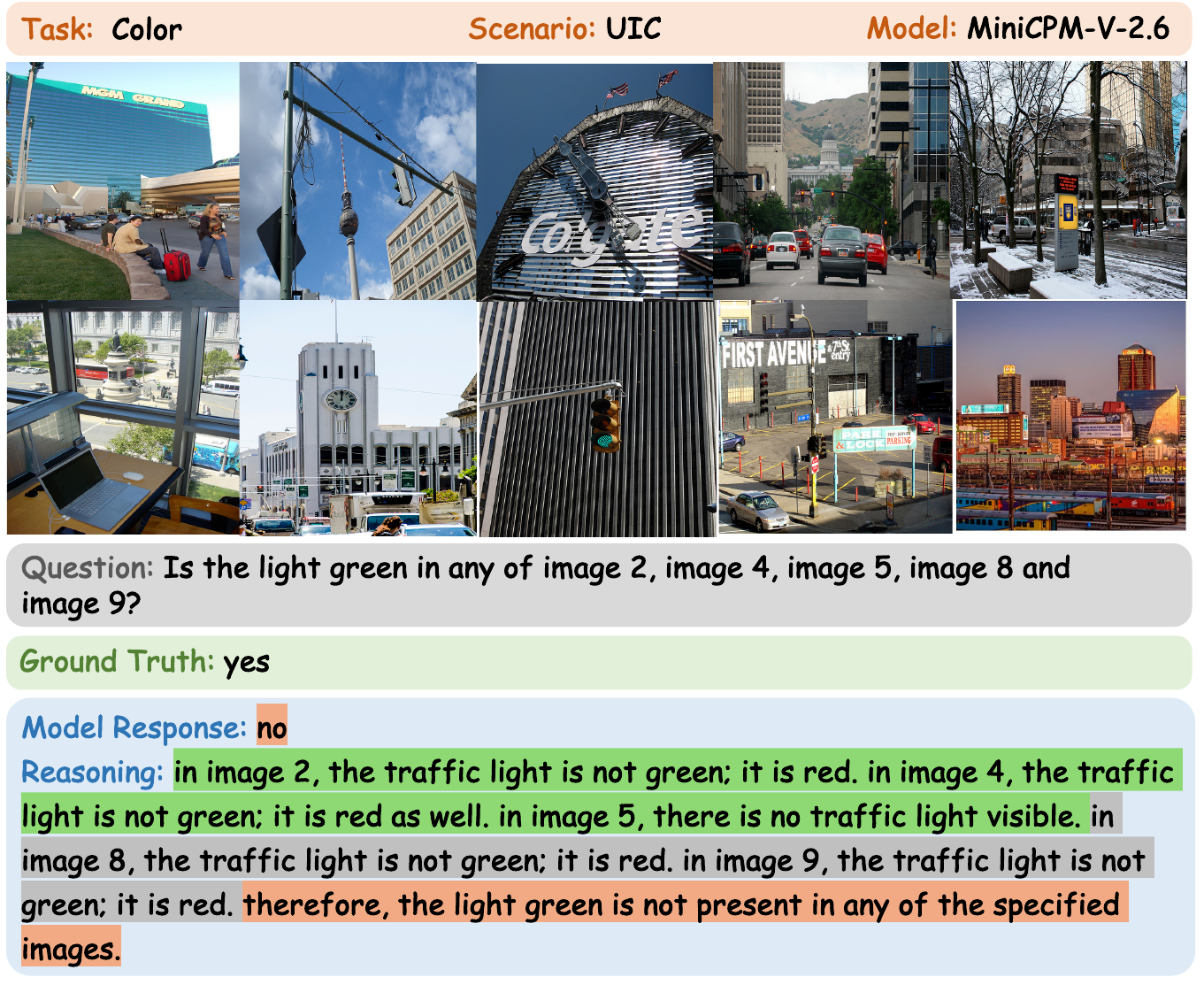}
    \end{subfigure}

    \par\vspace{2mm}
    \begin{subfigure}[t]{0.47\linewidth}
        \centering
        \vspace{0pt}
        \includegraphics[width=\linewidth]{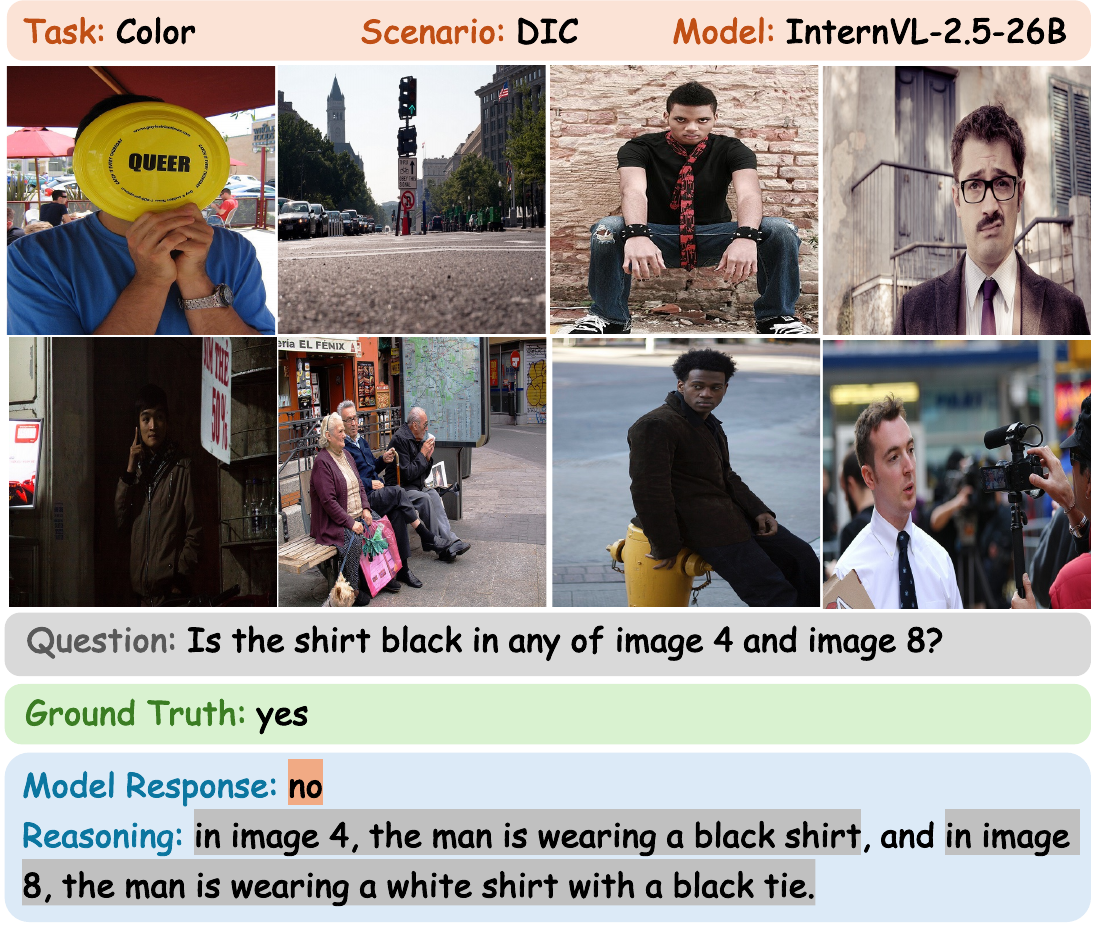}
    \end{subfigure}\hfill
    \begin{subfigure}[t]{0.47\linewidth}
        \centering
        \vspace{0pt}
        \includegraphics[width=\linewidth]{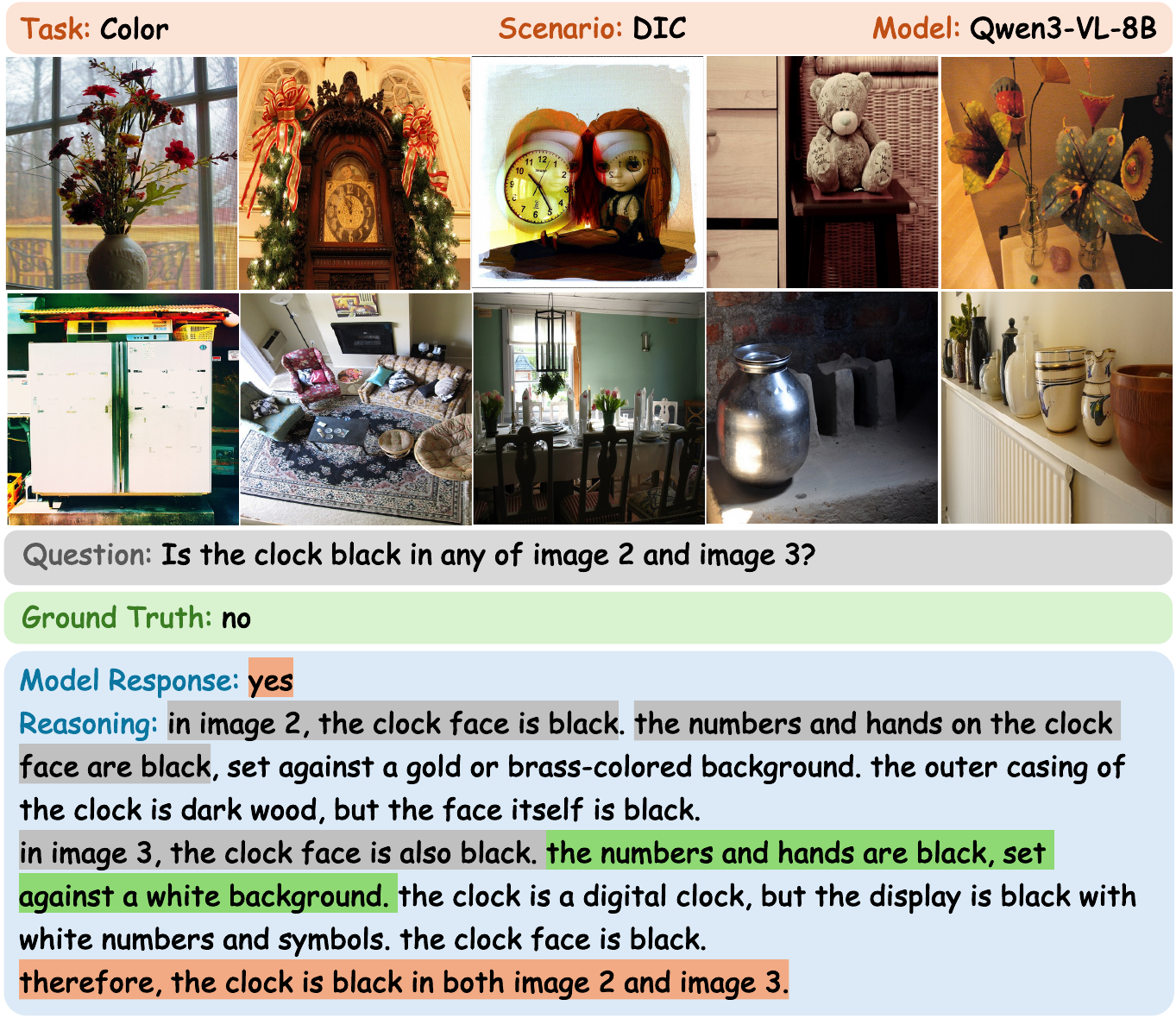}
    \end{subfigure}

    \caption{Color recognition examples under UIC and DIC ($k=5$--10).}
    \label{fig:case_study_k_scale_2}
\end{figure*}

\begin{figure*}[!tbp]
    \centering
    \begin{subfigure}[t]{0.47\linewidth}
        \centering
        \vspace{0pt}
        \includegraphics[width=\linewidth]{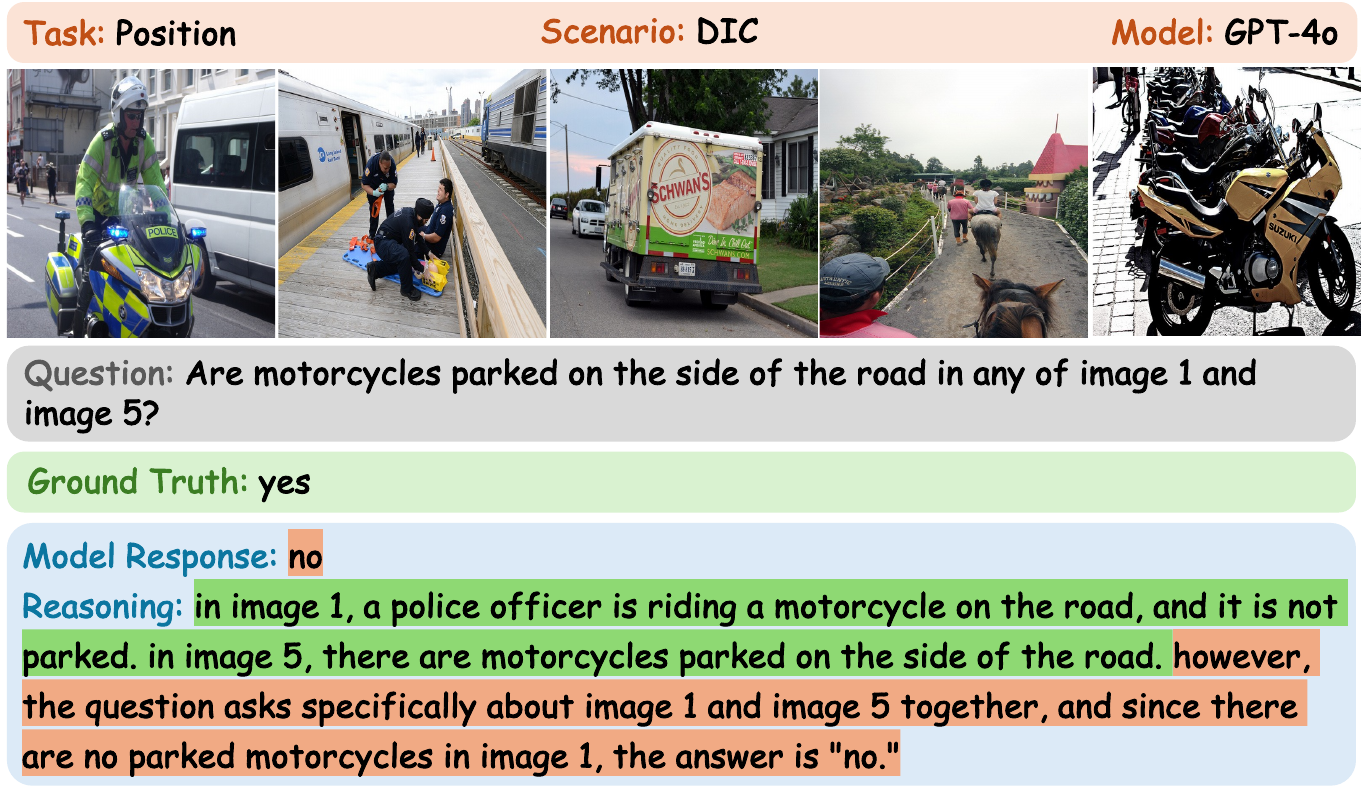}
    \end{subfigure}\hfill
    \begin{subfigure}[t]{0.47\linewidth}
        \centering
        \vspace{0pt}
        \includegraphics[width=\linewidth]{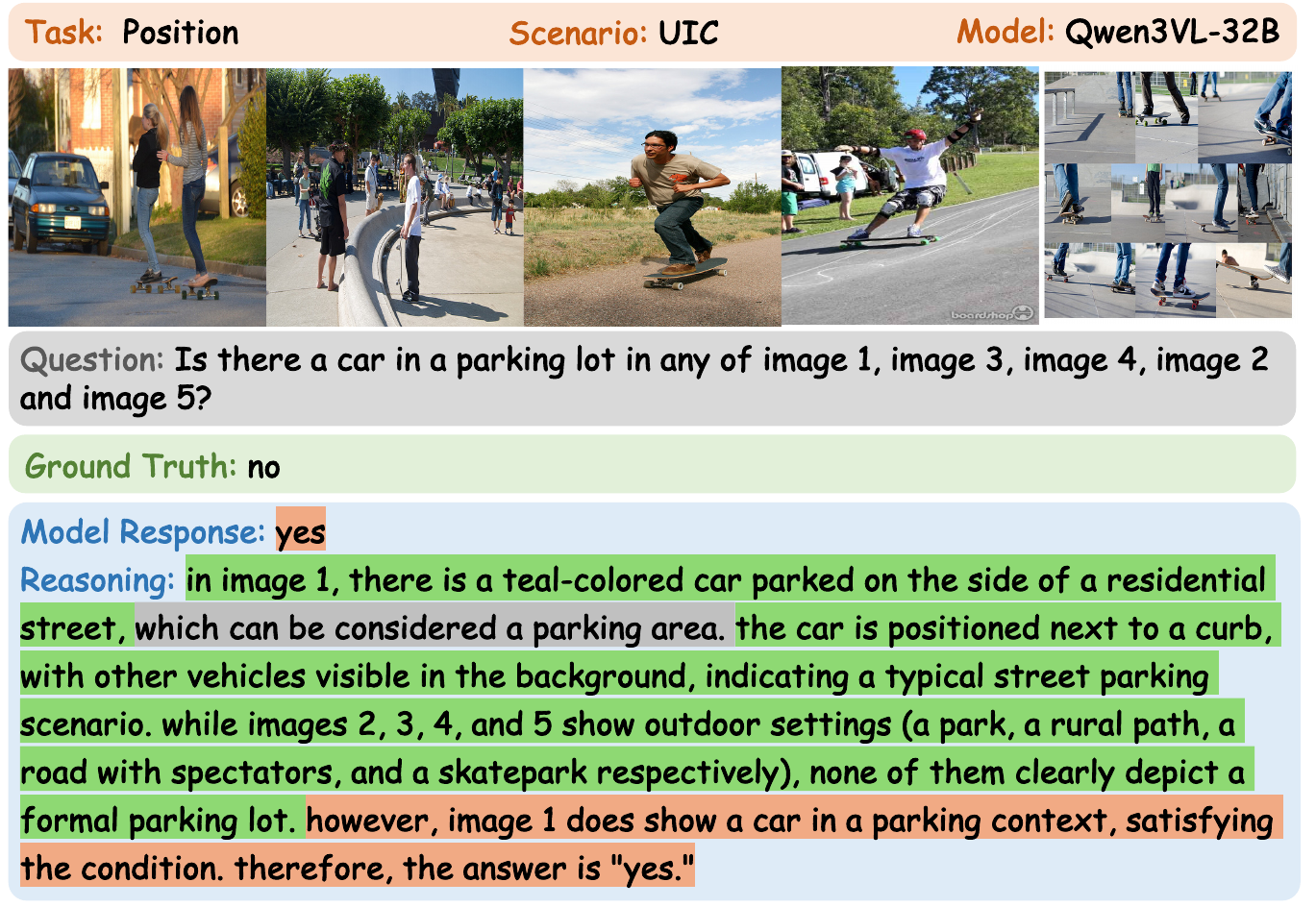}
    \end{subfigure}

    \par\vspace{2mm}
    \begin{subfigure}[t]{0.47\linewidth}
        \centering
        \vspace{0pt}
        \includegraphics[width=\linewidth]{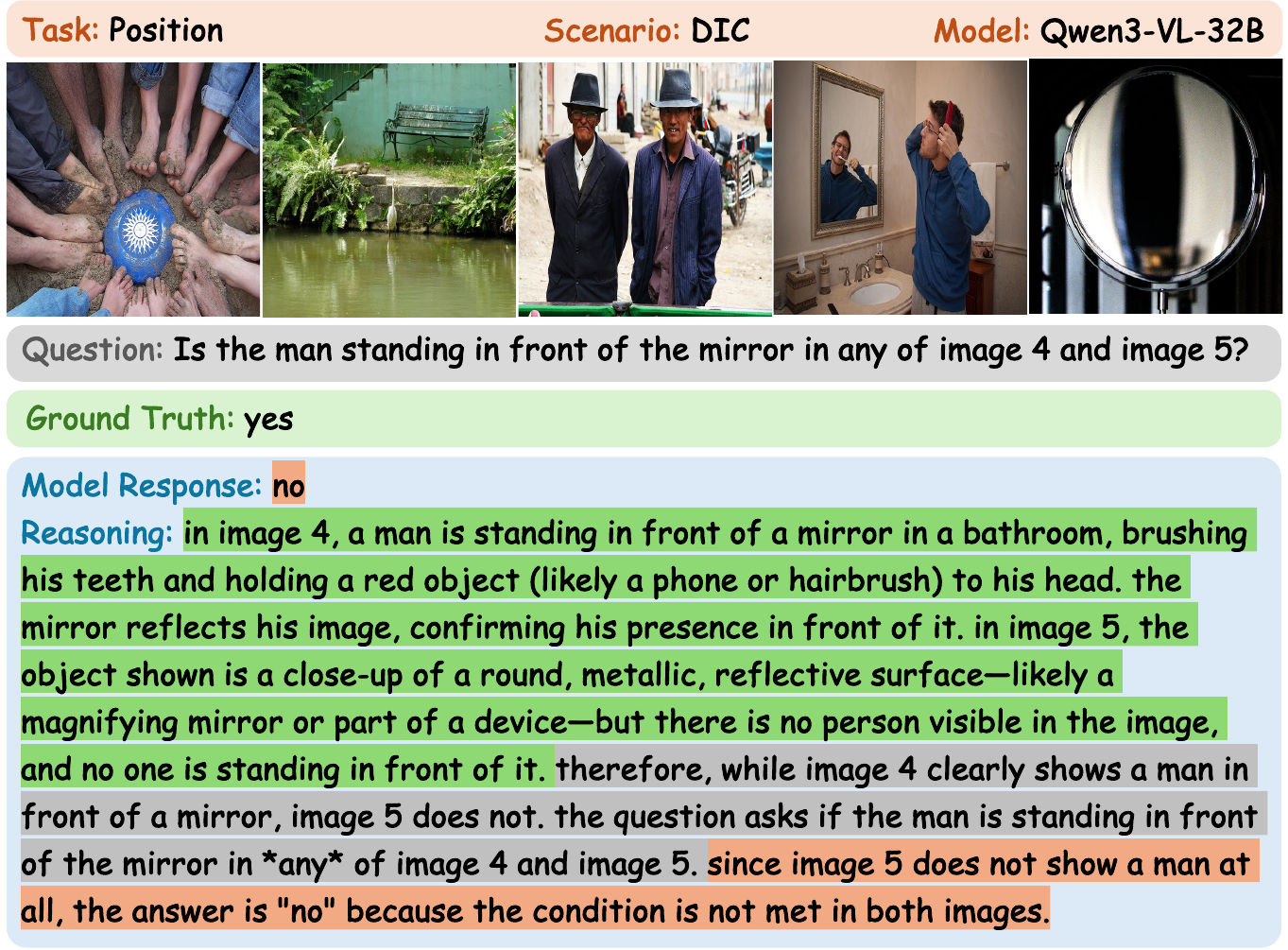}
    \end{subfigure}\hfill
    \begin{subfigure}[t]{0.47\linewidth}
        \centering
        \vspace{0pt}
        \includegraphics[width=\linewidth]{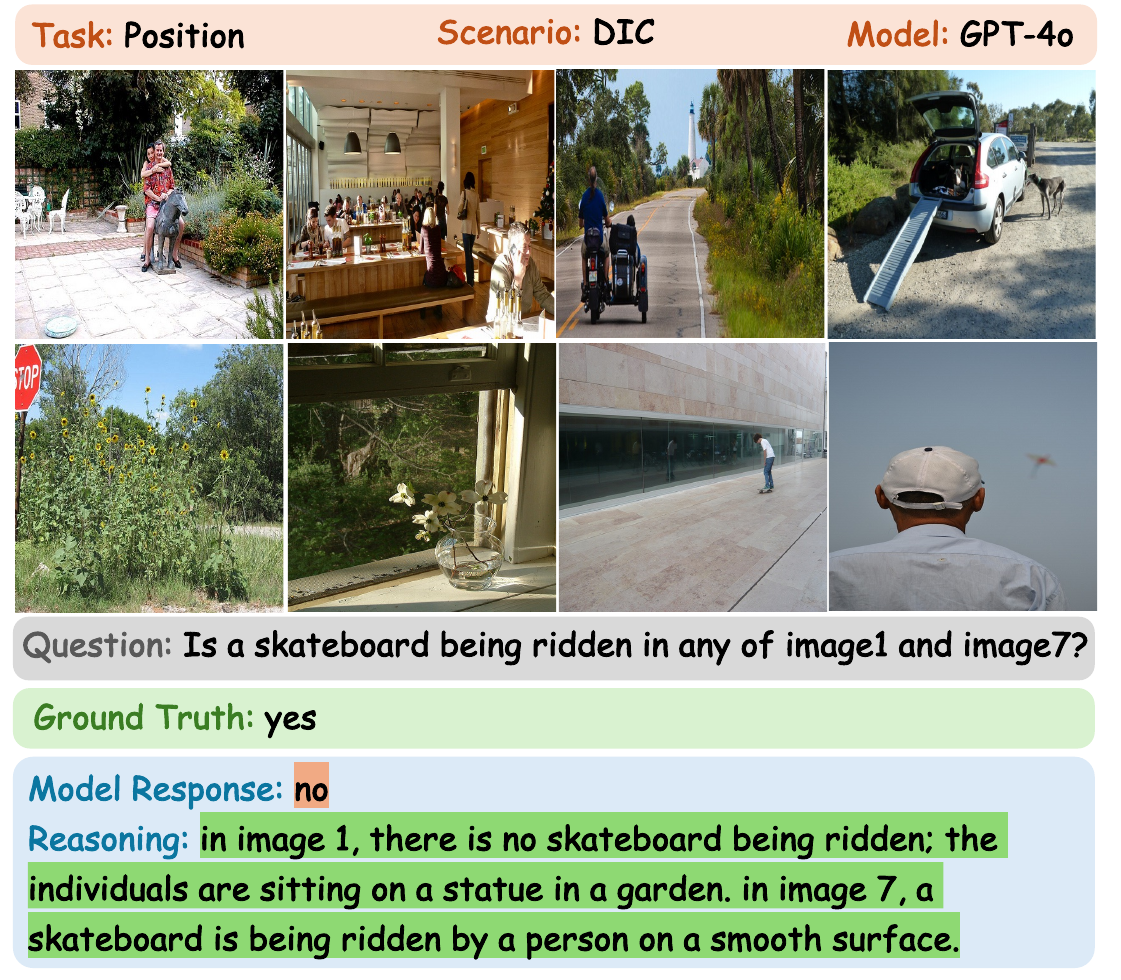}
    \end{subfigure}

    \par\vspace{2mm}
    \begin{subfigure}[t]{0.47\linewidth}
        \centering
        \vspace{0pt}
        \includegraphics[width=\linewidth]{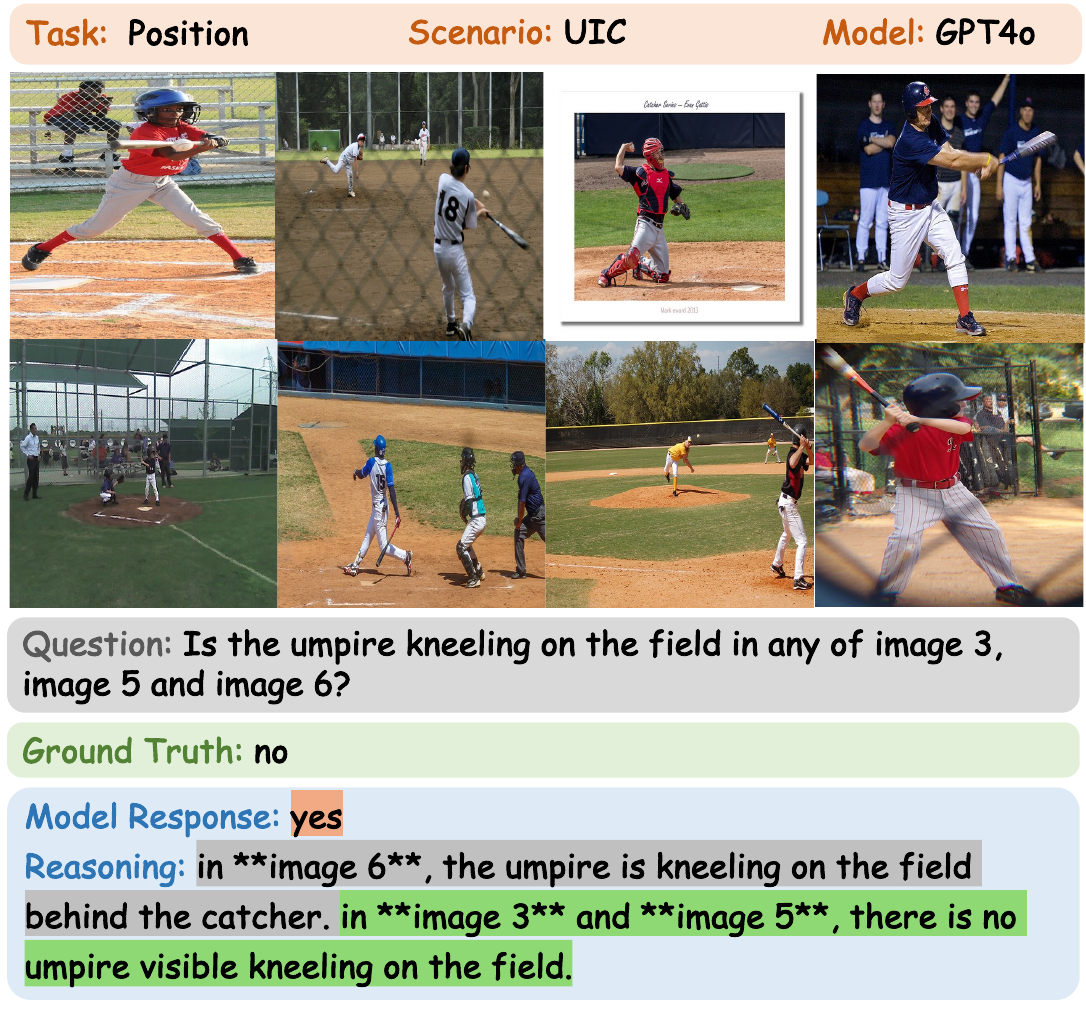}
    \end{subfigure}\hfill
    \begin{subfigure}[t]{0.47\linewidth}
        \centering
        \vspace{0pt}
        \includegraphics[width=\linewidth]{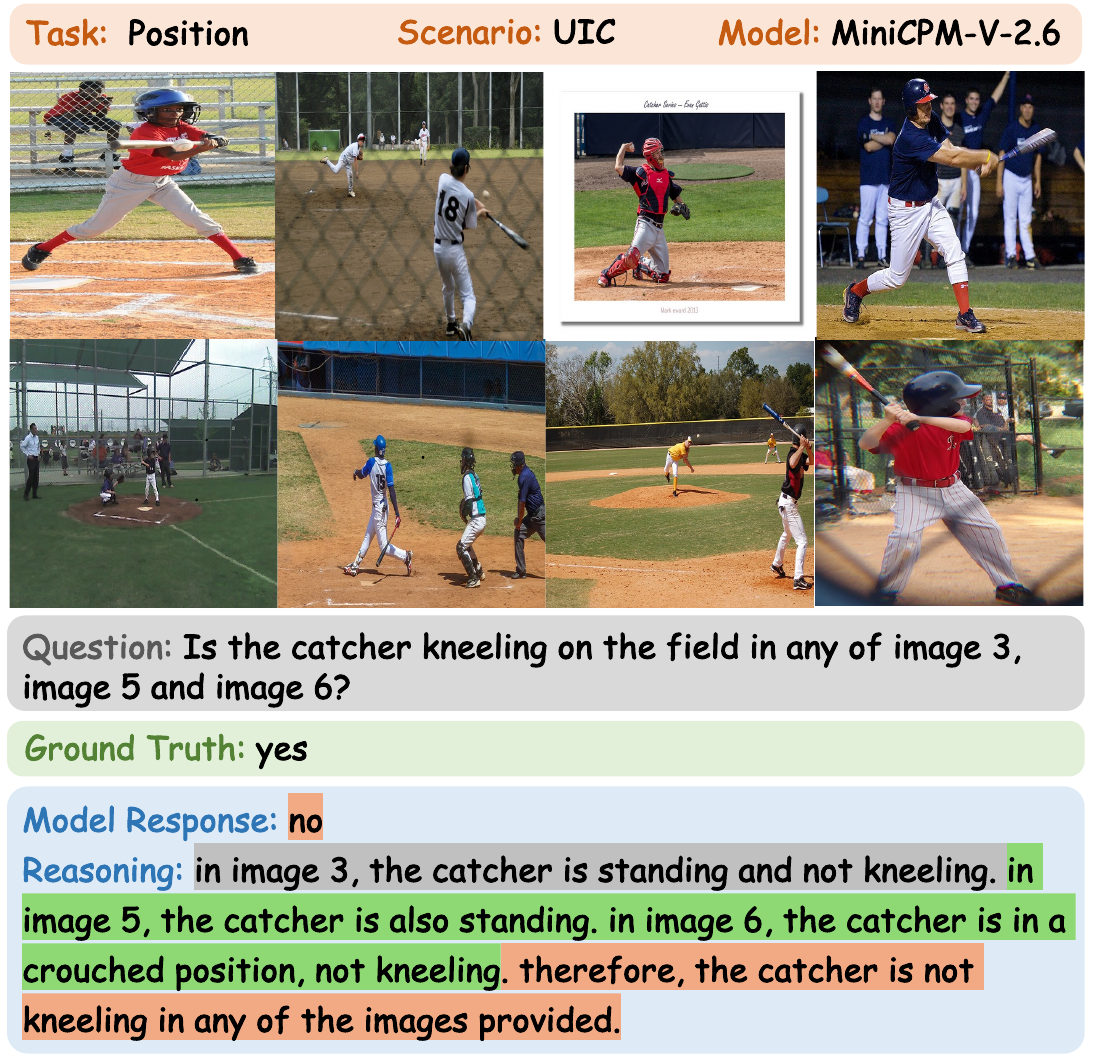}
    \end{subfigure}

    \caption{Position recognition examples under UIC and DIC ($k=5$--8).}
    \label{fig:case_study_k_scale_3}
\end{figure*}

\begin{figure*}[!tbp]
    \centering
    \begin{subfigure}[t]{0.47\linewidth}
        \centering
        \vspace{0pt}
        \includegraphics[width=\linewidth]{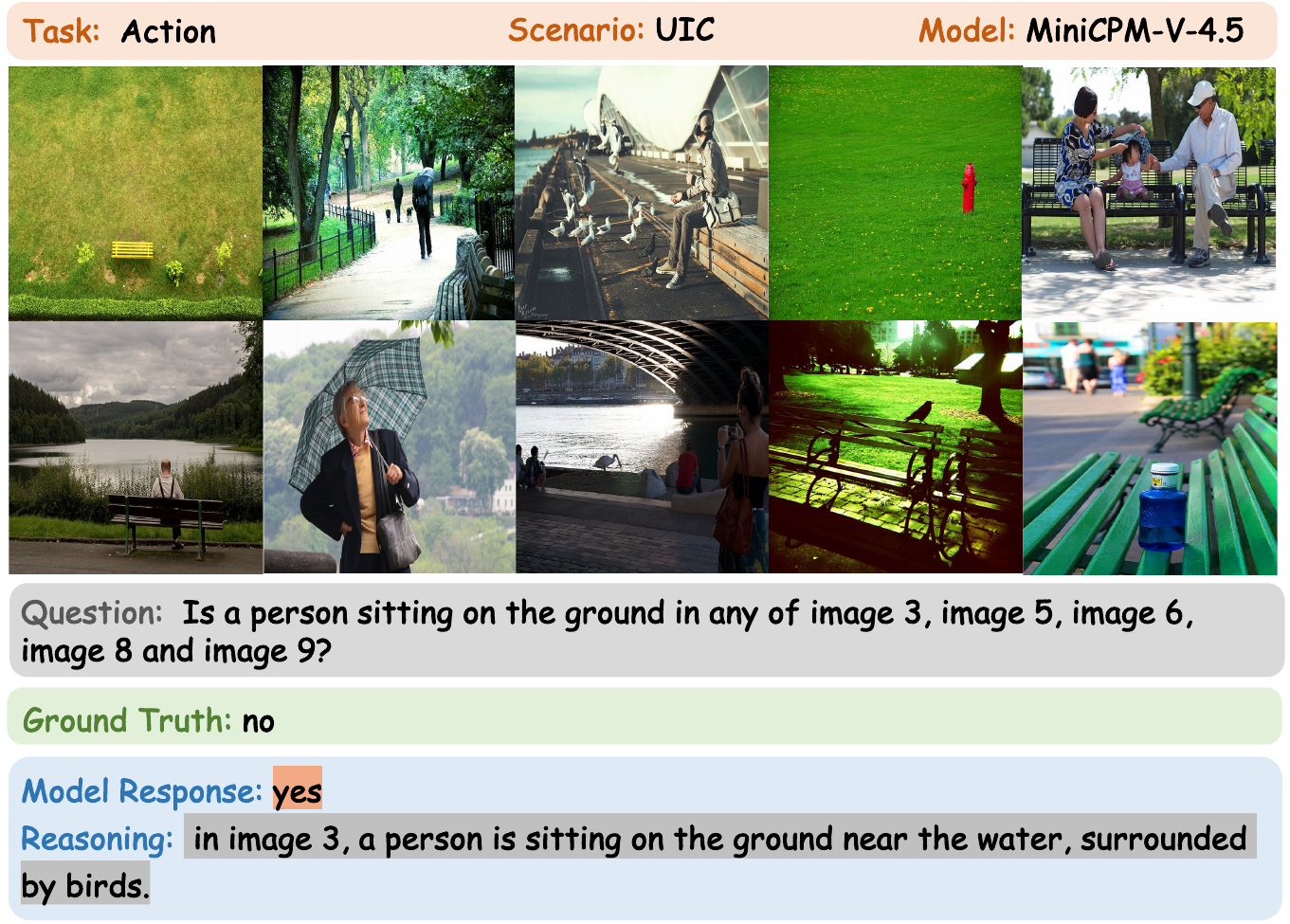}
    \end{subfigure}\hfill
    \begin{subfigure}[t]{0.47\linewidth}
        \centering
        \vspace{0pt}
        \includegraphics[width=\linewidth]{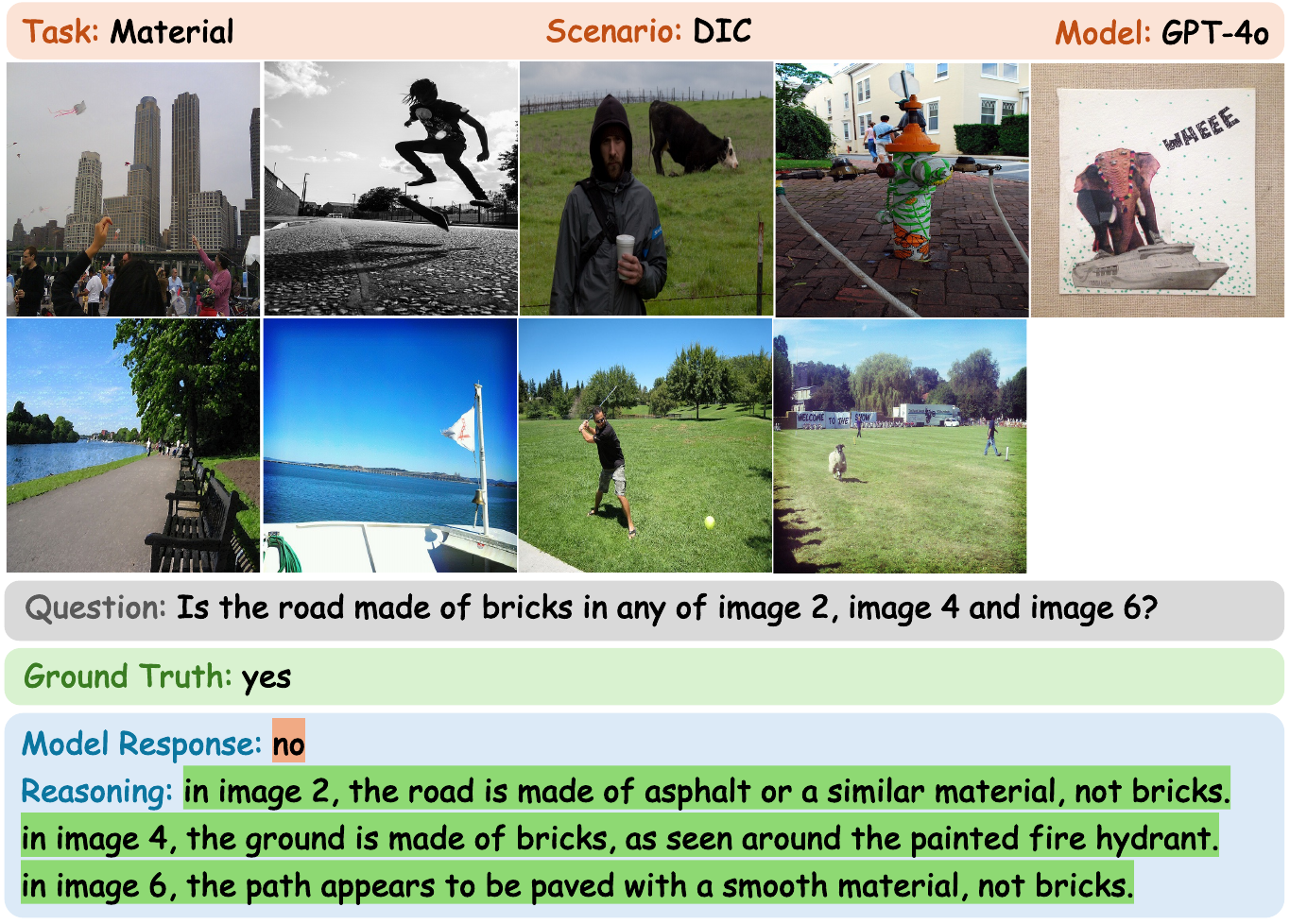}
    \end{subfigure}

    \par\vspace{2mm}
    \begin{subfigure}[t]{0.47\linewidth}
        \centering
        \vspace{0pt}
        \includegraphics[width=\linewidth]{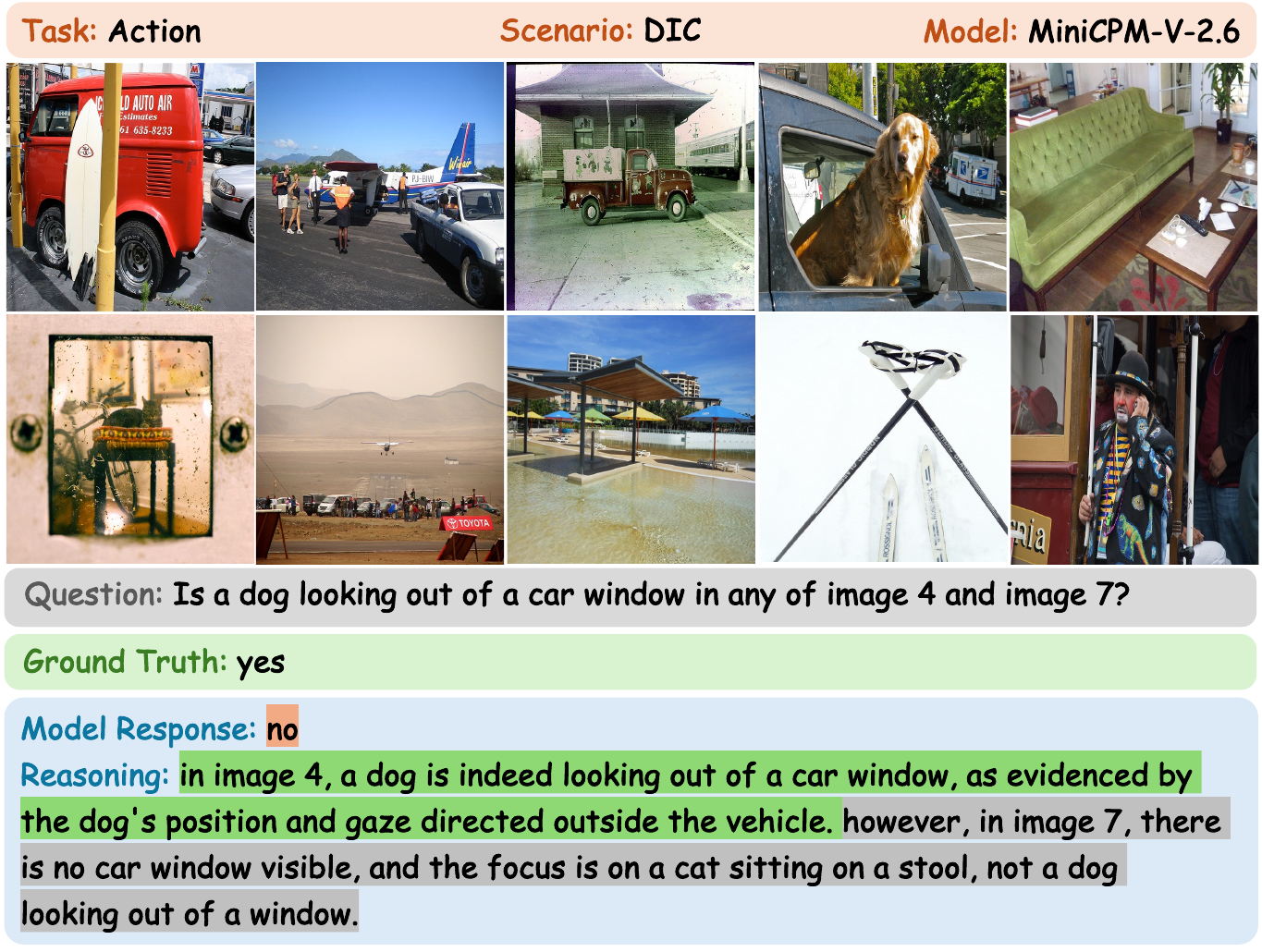}
    \end{subfigure}\hfill
    \begin{subfigure}[t]{0.47\linewidth}
        \centering
        \vspace{0pt}
        \includegraphics[width=\linewidth]{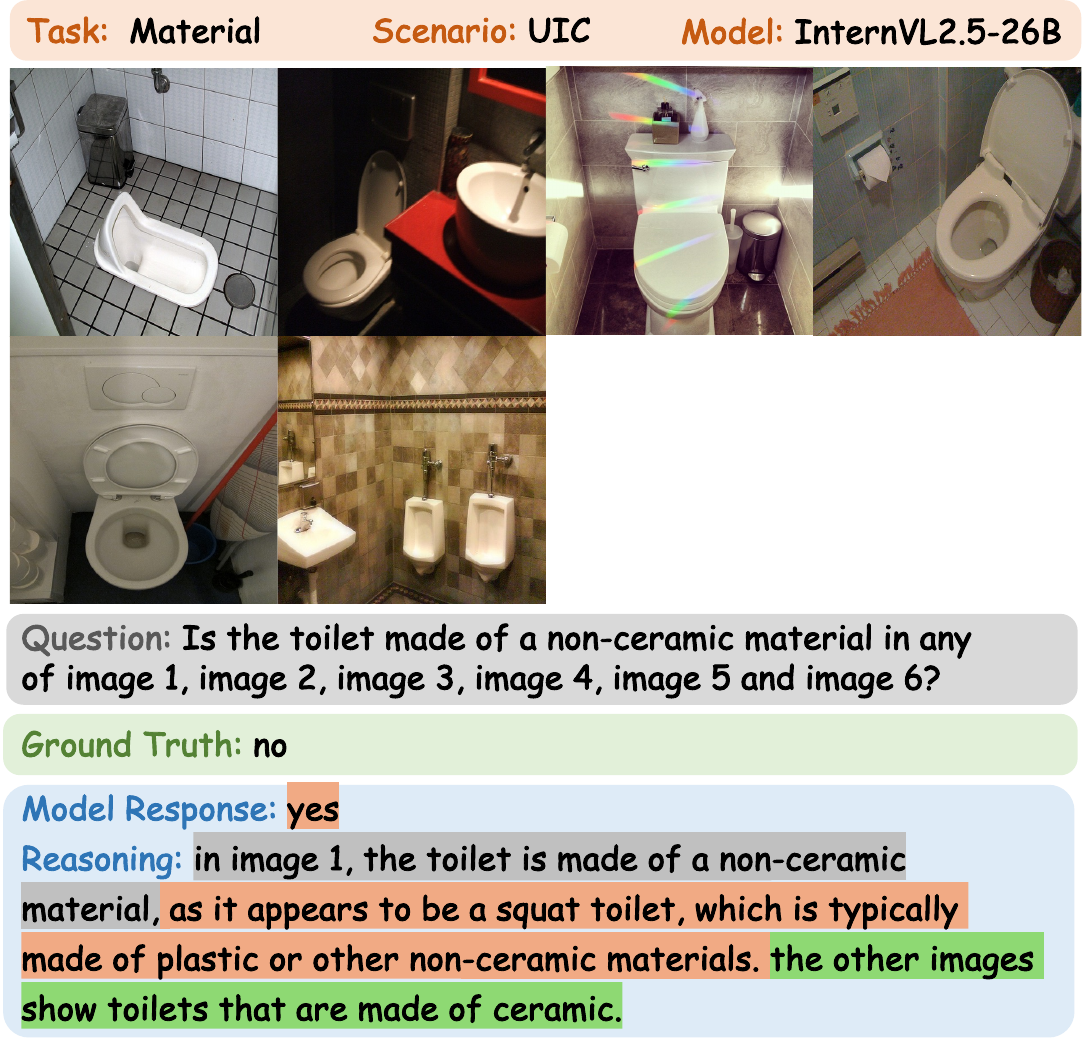}
    \end{subfigure}

    \par\vspace{2mm}
    \begin{subfigure}[t]{0.47\linewidth}
        \centering
        \vspace{0pt}
        \includegraphics[width=\linewidth]{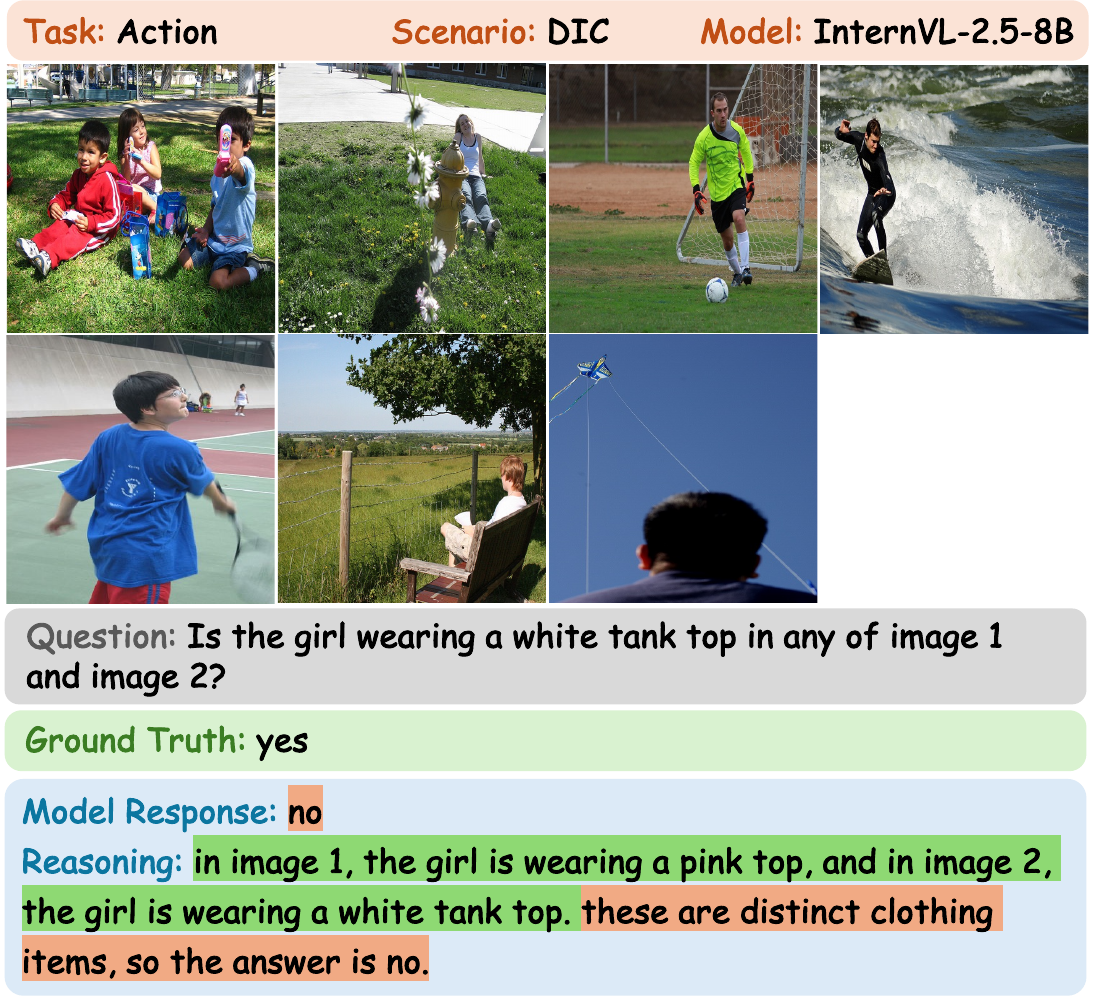}
    \end{subfigure}\hfill
    \begin{subfigure}[t]{0.47\linewidth}
        \centering
        \vspace{0pt}
        \includegraphics[width=\linewidth]{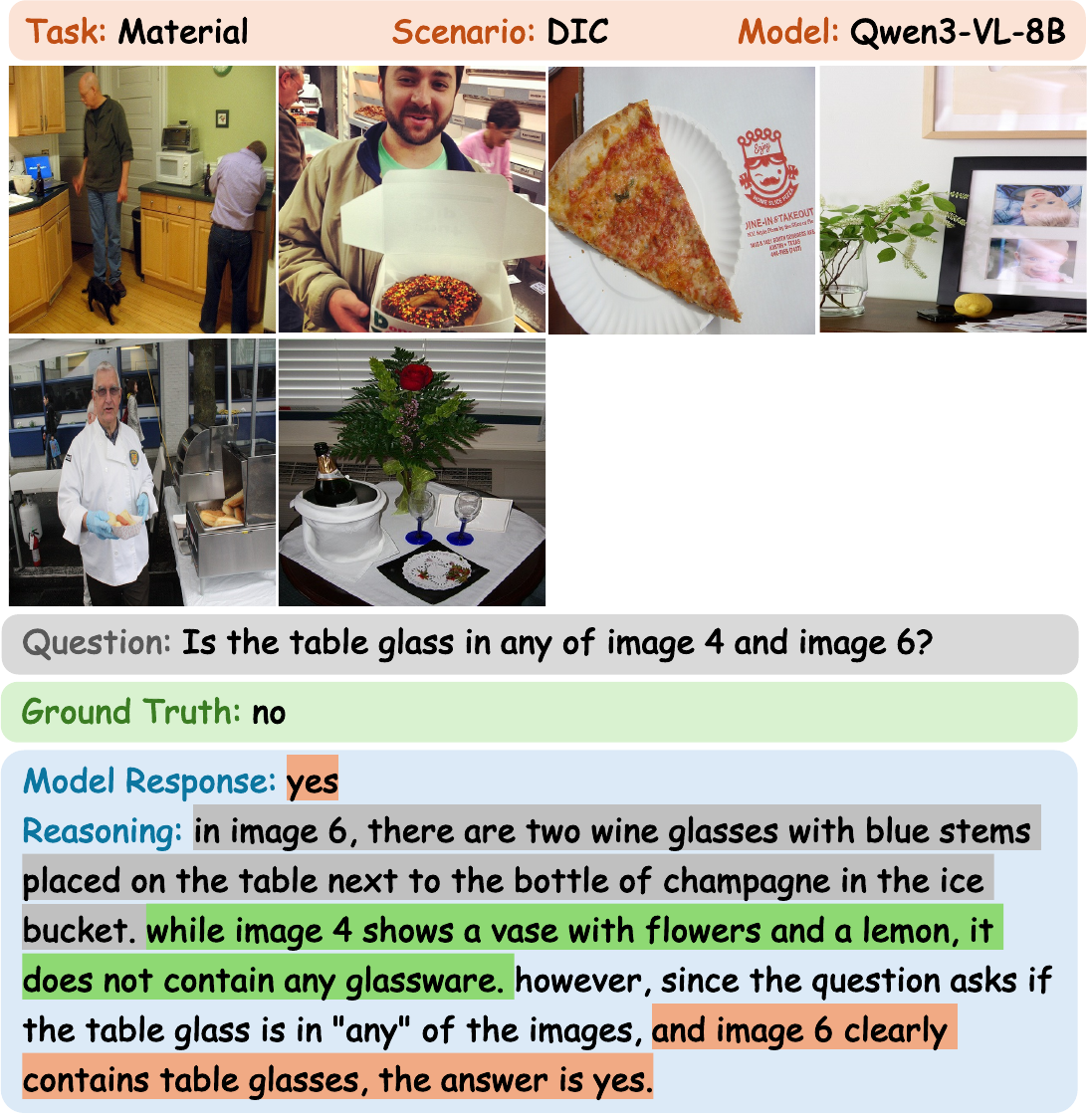}
    \end{subfigure}

    \caption{Action (left) and material (right) recognition examples ($k=6$--10).}
    \label{fig:case_study_k_scale_4}
\end{figure*}

\begin{figure*}[!tbp]
    \centering
    \begin{subfigure}[t]{0.47\linewidth}
        \centering
        \vspace{0pt}
        \includegraphics[width=\linewidth]{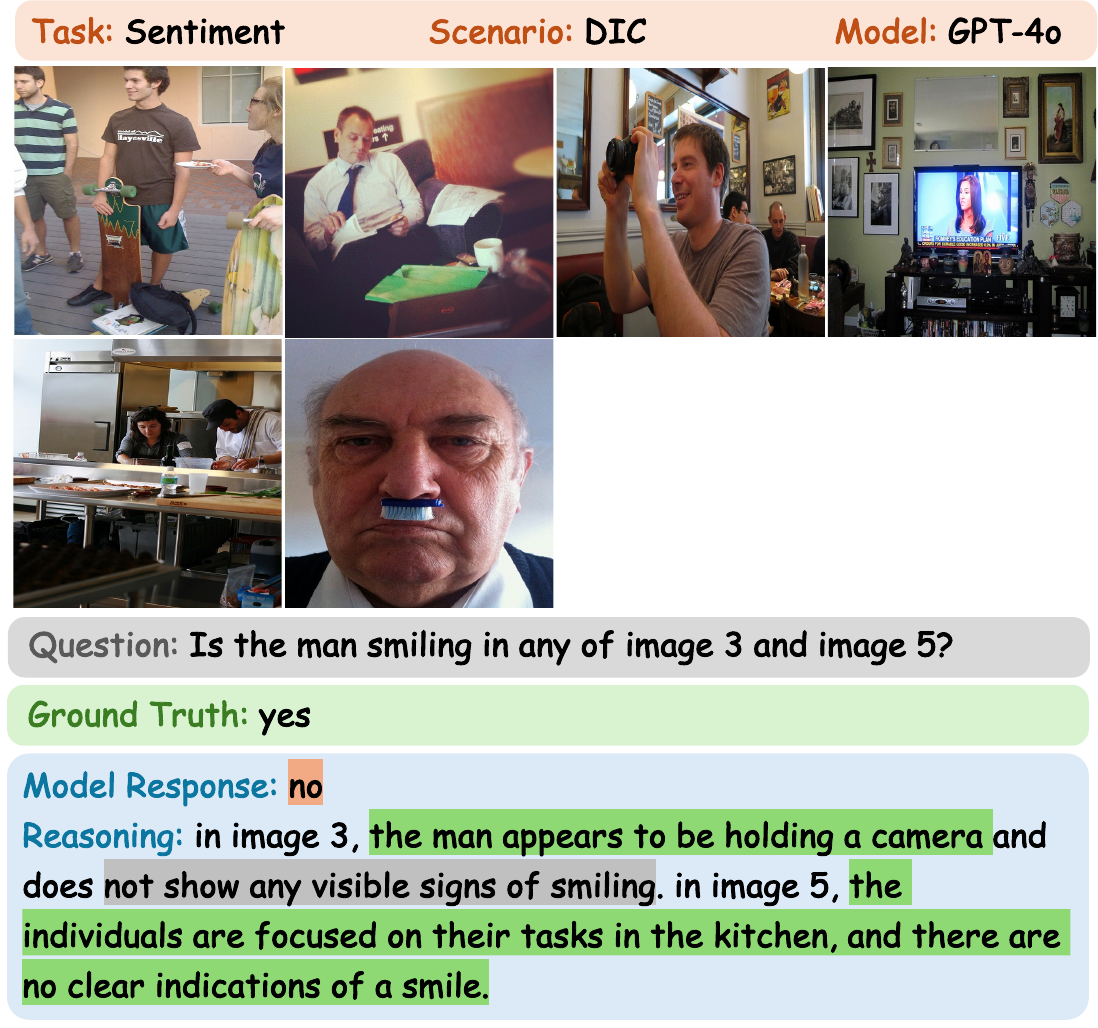}
    \end{subfigure}\hfill
    \begin{subfigure}[t]{0.47\linewidth}
        \centering
        \vspace{0pt}
        \includegraphics[width=\linewidth]{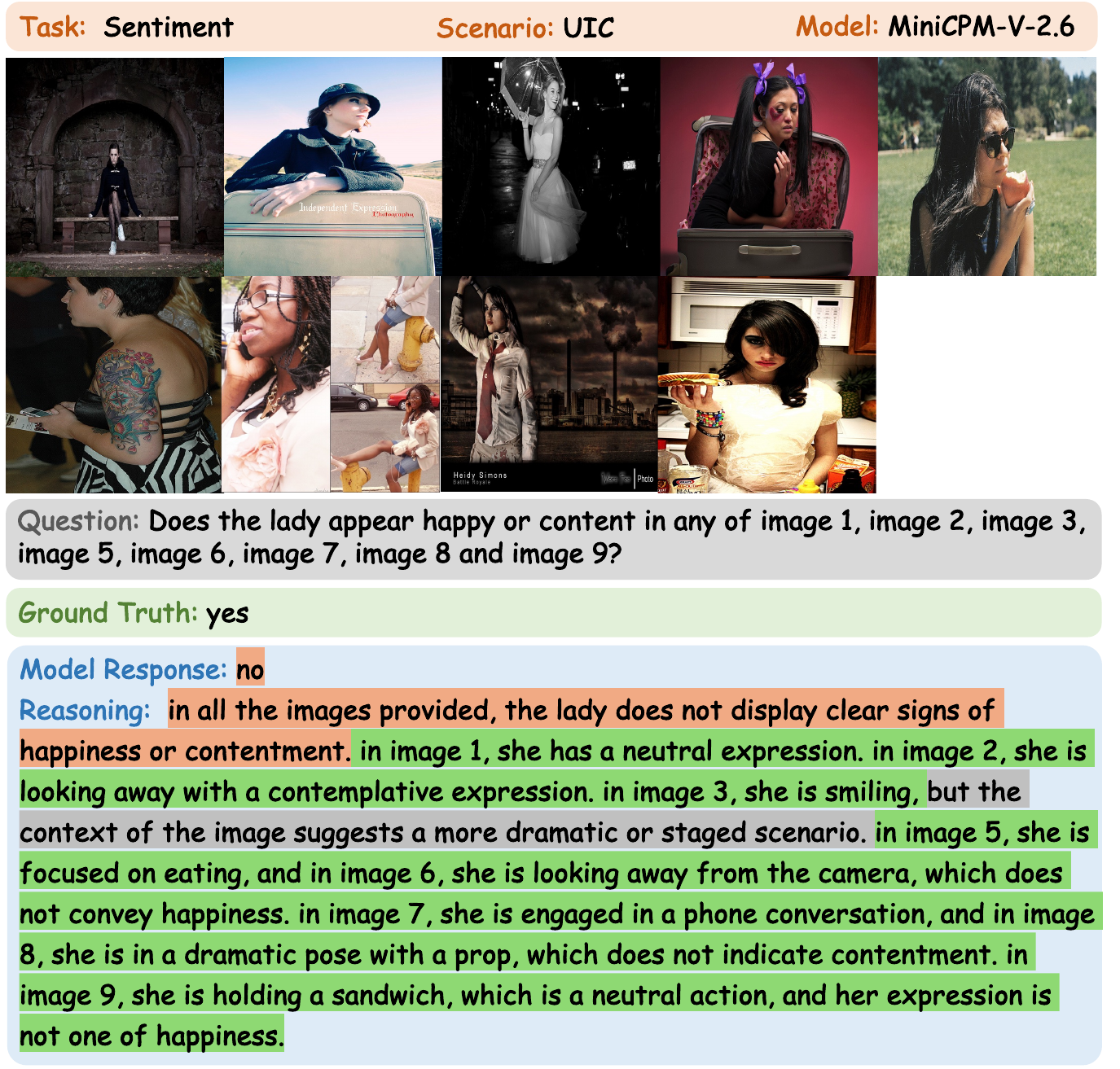}
    \end{subfigure}

    \par\vspace{2mm}
    \begin{subfigure}[t]{0.47\linewidth}
        \centering
        \vspace{0pt}
        \includegraphics[width=\linewidth]{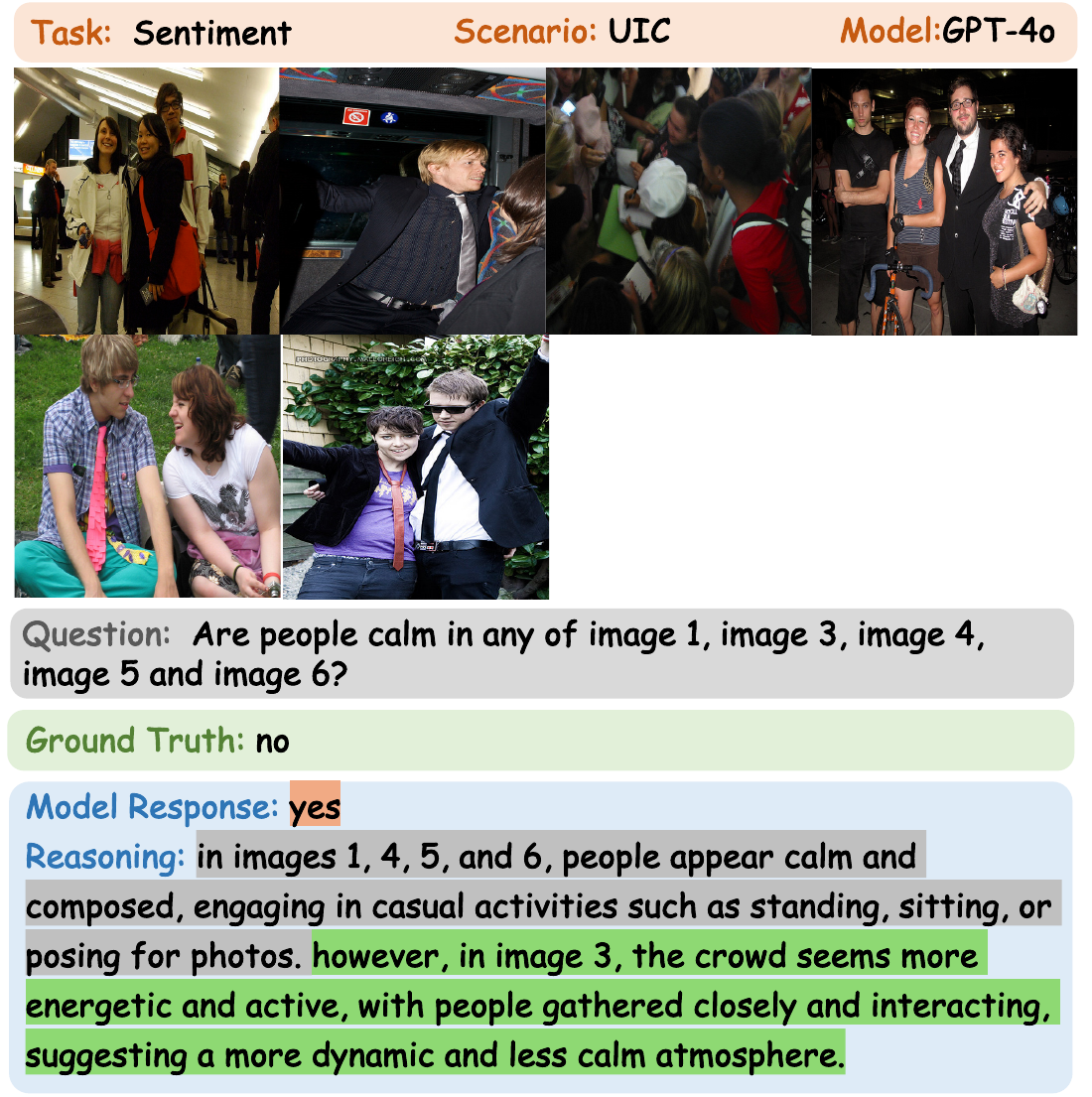}
    \end{subfigure}\hfill
    \begin{subfigure}[t]{0.47\linewidth}
        \centering
        \vspace{0pt}
        \includegraphics[width=\linewidth]{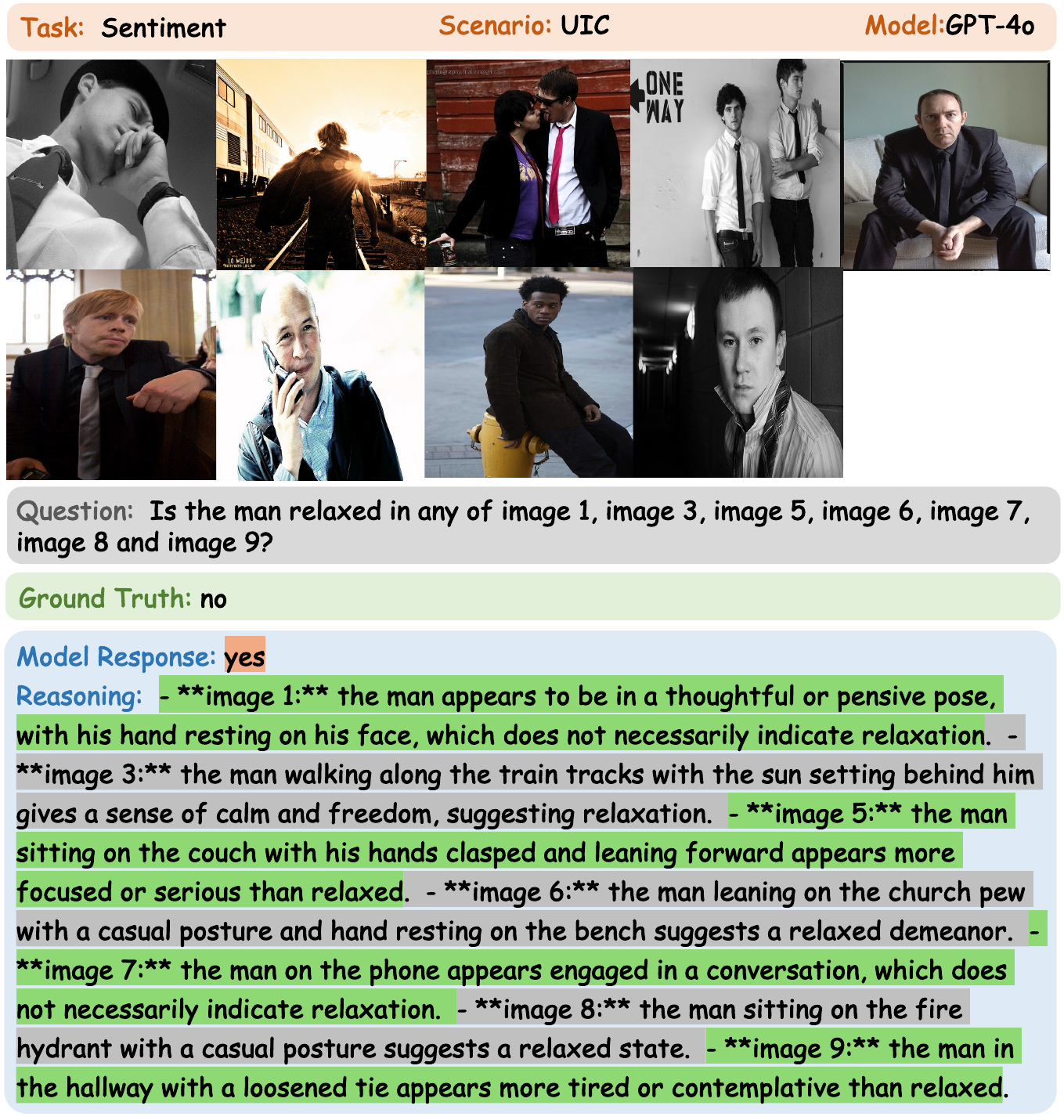}
    \end{subfigure}

    \par\vspace{2mm}
    \begin{subfigure}[t]{0.47\linewidth}
        \centering
        \vspace{0pt}
        \includegraphics[width=\linewidth]{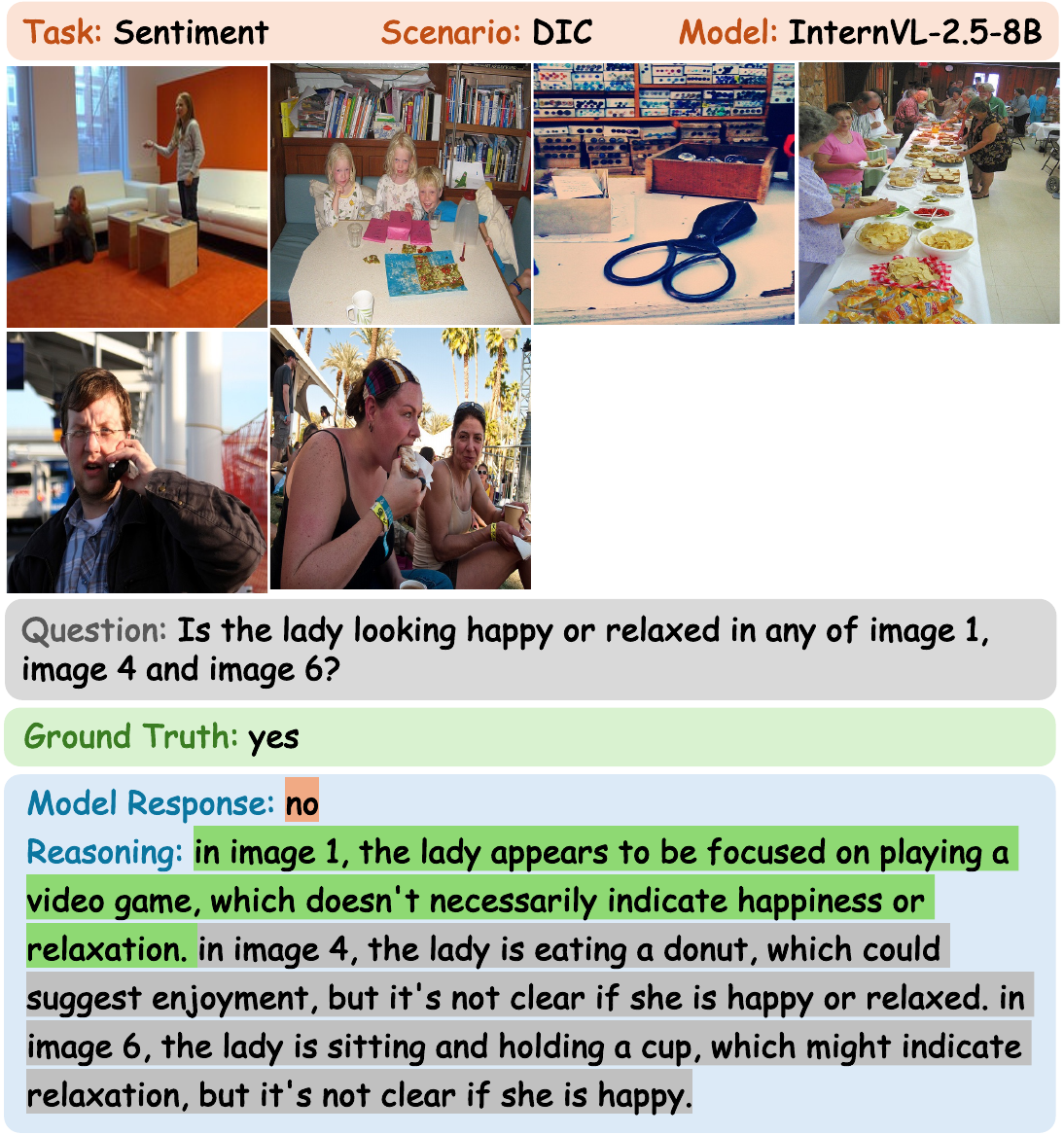}
    \end{subfigure}

    \caption{Sentiment recognition examples under UIC and DIC ($k=6$--9).}
    \label{fig:case_study_k_scale_5}
\end{figure*}

Figures~\ref{fig:case_study_id}, \ref{fig:case_study_mcq}, and~\ref{fig:case_study_openqa} show incorrect responses under Image Identification, Multiple-Choice QA, and Open-Ended QA, respectively. Figures~\ref{fig:case_study_k_scale_1}--\ref{fig:case_study_k_scale_5} show binary cases across all six recognition tasks with $k=4$--10 images, grouped by task. Together, these cases illustrate content confusion in UIC and context interference in DIC across response formats and context sizes.

In all cases, \cbox{myorange} marks factual errors, \cbox{mygray} marks conflicts with the target image, and \cbox{mygreen} marks correct target-image details.

